\documentclass{article}

\usepackage{arxiv}
\usepackage[numbers]{natbib}

\usepackage[utf8]{inputenc} % allow utf-8 input
\usepackage[T1]{fontenc}    % use 8-bit T1 fonts
\usepackage{hyperref}       % hyperlinks
\usepackage{url}            % simple URL typesetting
\usepackage{booktabs}       % professional-quality tables
\usepackage{amsfonts}       % blackboard math symbols
\usepackage{nicefrac}       % compact symbols for 1/2, etc.
\usepackage{microtype}      % microtypography
\usepackage{lipsum}
\usepackage{graphicx}
\usepackage{caption}
\makeatletter
\def\blfootnote{\gdef\@thefnmark{}\@footnotetext}
\makeatother
\theoremstyle{plain}

\theoremstyle{definition}

\theoremstyle{remark}

\title{Batch Clipping and Adaptive Layerwise Clipping for Differential Private Stochastic Gradient Descent}

\author{Toan N. Nguyen$^{1^{*}}$, \textbf{Phuong Ha Nguyen}$^{2^{*}}$, \textbf{Lam M. Nguyen}$^{3}$,Marten van Dijk$^{4,5,6}$ \\ 
   \\
$^{1}$ Department of Computer Science and Engineering, University of Connecticut, CT, USA \\
$^{2}$ eBay, CA, USA\\
$^{3}$ IBM Research, Thomas J. Watson Research Center, Yorktown Heights, NY, USA\\
$^{4}$ CWI Amsterdam, The Netherlands\\
$^{5}$ Department of Computer Science, Vrije Universiteit Amsterdam, The Netherlands \\
$^{6}$
Department of Electrical and Computer Engineering, University of Connecticut, CT, USA\\
\\
 \texttt{toan.nguyen@uconn.edu}, \texttt{phuongha.ntu@gmail.com},\\
\texttt{LamNguyen.MLTD@ibm.com}, \texttt{marten.van.dijk@cwi.nl}
}

\begin{document}
\maketitle

\blfootnote{$^{*}$ these authors contributed equally.}
% \blfootnote{$^{\dagger}$ Supported by NSF grant CNS-1413996 “MACS: A Modular
% Approach to Cloud Security.”}
% \blfootnote{}
\begin{abstract}
Each round in Differential Private Stochastic Gradient Descent (DPSGD) transmits a sum of clipped gradients obfuscated with Gaussian noise to a central server which uses this to update a global model which often represents a deep neural network. Since the clipped gradients are computed separately, which we call Individual Clipping (IC), deep neural networks like resnet-18  cannot use Batch Normalization Layers (BNL) which is a crucial component in deep neural networks  for achieving a high accuracy. To utilize BNL, we introduce Batch Clipping (BC) where, instead of clipping single gradients as in the orginal DPSGD, we average and clip batches of gradients. Moreover, the model entries of different layers have different sensitivities to the added Gaussian noise. Therefore, Adaptive Layerwise Clipping methods (ALC), where each layer has its own adaptively finetuned clipping constant, have been introduced and studied, but so far without rigorous DP proofs. In this paper, we propose {\em a new ALC and provide rigorous DP proofs for both BC and ALC}. Experiments show that our modified DPSGD with BC and ALC  for CIFAR-$10$ with resnet-$18$ converges while DPSGD with IC and ALC does not.  
\end{abstract}

% Content %
\section{Introduction}
\label{section:Introduction}

%TODO: Intro for ML and DP

% Add the citations here
% Machine Learning and Deep Learning have had a huge impact on both research and the technology industry~\cite{GoodBengCour16}. Many machine learning and deep learning models have been developed and improved for problems such as regression, classification, and recognition, with higher accuracy rates~\cite{resnet18paper}. However, these models are vulnerable to many attacks, such as membership-inference attacks~\cite{shokri2017membership}, model inversion [CITE], model poisoning[CITE], and training data poisoning[CITE], and deep-leakage attacks~\cite{ligengzhu201deepleakage}. These attacks aim to gain access to the training databases used to train the deep learning models, replicate those models, or even weaken those models during the training phase.

% In this work, we focus on protecting the privacy of the training database of deep learning models, preventing individual data in the training database from being leaked in the training phase by using Differential Privacy (DP) mechanisms. 

% Differential Privacy (DP)\citep{dwork2006calibrating, dwork2011firm,dwork2014algorithmic,dwork2006our} is a privacy-preserving data analysis method that helps us measure privacy attributes and provide privacy-preserving mechanisms that we can combine with machine learning training algorithms~\citep{mcmahan,nguyen2018sgd,nguyen2018new}. Differential privacy ensures that adding or removing a single database item does not affect the outcome of any analysis. 

Differential Private Stochastic Gradient Descent (DPSGD)~\cite{abadi2016deep} combines Stochastic Gradient Decent (SGD)~\cite{RM1951}
%,nguyen2018sgd,nguyen2018new} 
and Differential Privacy (DP)\citep{dwork2006our}
%\citep{dwork2006calibrating, dwork2011firm,dwork2014algorithmic,dwork2006our} 
to train deep neural networks privately. It has been widely studied since its introduction.
%in 2016. 

In each round of DPSGD a mini-batch of $m$ samples $\xi_{i_1},\dotsc,\xi_{i_m}$ from a larger data set $d=\{\xi_i\}_{i=1}^N$ of $N$ samples is randomly subsampled:
\begin{equation}
    \{i_1,\ldots, i_m\} \leftarrow {\tt Sample}_m(N). \label{eq:sample}
\end{equation}
The global model $w$ is updated by first computing gradients
$\nabla_w f(w;\xi_{i_j})$  for each sample $\xi_{i_j}$ given loss function $f$. Each gradient is clipped by using a clipping operation $[x]_C= x/\max\{1,\|x\|/C\}$ where $C$ denotes a fixed clipping constant. The clipped gradients are aggregated in a sum after which a noise vector $n$ drawn
from a Gaussian distribution\footnote{$\mathbf{I}$ is the identity correlation matrix of the multivariate Gaussian distribution. The factor $2$ is due to sampling exactly $m$ data points every round; by using a Poisson process to  probabilistically sample $\xi \in d$ such that in expectation the mini-batch has size $m$ removes the factor 2 and leads to better DP. This is implemented in \citep{Opacus}. In order to simplify our exposition we work with deterministic sampling and keep the factor 2.} $\mathcal{N}(0,(2C\sigma\mathbf{I})^2)$ is added. The resulting "noised and clipped" update $U$ 
is transmitted to a central server where it is averaged over the size $m$ of the used mini-batch and multiplied by a step-size $\mu$ before subtracting it from the global model:
% \begin{equation}
% \label{algo:DPSGDx}
%     w := w - \eta (\frac{1}{m} \sum_{i=1}^m [\nabla f_w(w;\xi_i)]_C + n), 
% \end{equation}
%
%\textcolor{red}{[TO DO: I think we need the formula below and say that $n$ is a vector with entries drawn from ... But we need a different symbol since we already use $n$ in the abstract.]}
%
%\textcolor{red}{Ha to Marten} Can we say that we abuse the notion $n$, i.e., $n$ can be a number or vector depending on the context we are discussing? It may make the reading smoothly. About the formula, we should use your proposed formula because I did not correctly remember the formula of the weight updating. 
%
\begin{eqnarray}
\label{algo:DPSGDx}
U &:=& n+\sum_{j=1}^m [\nabla_w f(w;\xi_{i_j})]_C 
%% Do not change to f_w !!!!!!!
\ \ \ \ \mbox{ and } \ \ \ \
    w := w -  \frac{\eta}{m} U.
\end{eqnarray}
%where $\eta$ is a step-size, $\nabla f_w(w;\xi_i)$ is the gradient of sample $\xi_i$ given loss function $f$, $n$ is a noise drawn from a Gaussian distribution\footnote{In the abstract we use $n$ for Gaussian noise which is drawn from $\mathcal{N}(0,(C\sigma\mathbf{I})^2)$ for each clipped  } $\mathcal{N}(0,(C\sigma\mathbf{I})^2)$, and where $[x]_C= x/\max\{1,\|x\|/C\}$ is a clipping operation with value $C$. 
In DPSGD, 
$\sigma$ translates to a DP guarantee and is chosen carefully by the designer to balance DP and the accuracy of the final global model.

% \textcolor{red}{HA To Marten: We have found that our proposed adaptive is not completely new. People already knows how to use published dataset U and layer gradient norms for deriving layerwise clipping constants $C_i$. Compared to the existing works, we combine all three following ideas together: decaying master constant C, published dataset U and layer gradient norms for deriving layerwise clipping constants $C_i$. The derivation of $C_i$ from the master $C$ and layer gradient norms is new. More importantly, it seems there are no papers providing the rigorous DP proof. Anyway, the new adaptive clipping method seems to be a minor contribution. The major contribution now is batch clipping because batch clipping allows us to any deep neural networks and existing training method without any modifications due to the restriction of individual clipping mode. However, we have the following problem. We still want to write that new clipping method is important because we cannot significantly change the title and abstract. The thing we want to add to title and abstract is we have to mention the batch clipping there. By combining batch clipping and adaptive clipping method, we can train deep neural network with high accuracy very fast with low privacy budget}

% In each round, SGD computes gradients of a mini-batch of samples first and then clips each gradient with a clipping constant $C$ and its norm. Finally, for each entry in the sum of clipped gradients is added. The parameter $\sigma$ is privacy guarantee chosen by the designer. 

%==== TO DO - edit text below ====

The global model is represented by a weight vector $w$ which can be written as a concatenation $(w_1 | \ldots | w_L)$ where $w_h$ corresponds to the $h$-th layer in the neural network which is being trained. DPSGD uses a fixed clipping constant $C$ for clipping  gradients 
$$
\nabla_w f(w;\xi_{i_j})
= (\nabla_{w_1} f(w;\xi_{i_j}) | \ldots | \nabla_{w_L} f(w;\xi_{i_j}) ).
$$
In expectation the norms $\|\nabla_{w_h} f(w;\xi_{i_j}) \|$ of different layers $h$  in a deep neural network model vary layer by layer. This makes different layers have different sensitivity to the added Gaussian noise. For this reason, \citep{van2018three,Xu2021,Zhang2018CoRR} proposed\footnote{For completeness, per-layer clipping was originally introduced in DP-FEDAVG \citep{mcmahan2018learning} where a clipping budget is {\em evenly} (not customized) distributed  among all layers and a DP guarantee is proven based on the moment accountant from \citep{abadi2016deep}.} Adaptive Layerwise Clipping (ALC) which tunes the added Gaussian noise with respect to the (sensitivity of the) layer. Each layer of the model has its own customized clipping constant based on estimating the expectation of the gradient norms of each layer by sampling  a given small public dataset $\mathcal{D}_{pub}$. 
%Unfortunately, there are 
We notice that  \citep{van2018three,Xu2021,Zhang2018CoRR} have not provided rigorous DP proofs 
%provided 
for ALC. In this paper we prove DP guarantees for ALC and we enhance the ALC method of \citep{Zhang2018CoRR} for improved test accuracy.

% If $\mathcal{D}$ is private, i.e., it is a small subset of the training dataset $d$, then the layer clipping constants must be computed privately; Gaussian noise is drawn and added to the layer clipping constants\footnote{\cite{van2018three}  computes layer
% clipping constants $C_i$  based on the layer gradient norms from the previous round (${\cal D}$ is the mini-batch used in the previous round).
% DPSGD-F \citep{Xu2021} calculates the layerwise clipping constants $C_i$ proportional to the ratio of gradients exceeding a threshold.}. The main shortcoming of this approach is that these noised layer clipping constants can become large because the sampled noise can be large. This hurts the convergence of the training process (see the experiments in~\citep{van2018three} for more details). 
% We also notice that \citep{van2018three,Xu2021} do not have rigorous DP proofs. We remedy this situation by using a public $D$ which leads to high accuracy and by providing a DP analysis for LC.

%\textbf{Motivation:} 
Besides the enhanced ACL, we introduce Batch Clipping (BC) where 
(\ref{algo:DPSGDx}) is replaced by
\begin{eqnarray}
\label{algo:DPSGDxBC}
U &:=& n+\left[\frac{1}{m} \sum_{j=1}^m  \nabla_w f(w;\xi_{i_j}) \right]_C 
%% Do not change to f_w !!!!!!!
\ \ \ \ \mbox{ and } \ \ \ \
    w := w -  \eta U.
\end{eqnarray}
BC allows us to first compute an average of a batch of gradients before clipping, as opposed to (\ref{algo:DPSGDx}) which averages a sum of clipped individual gradients; for this reason we call (\ref{algo:DPSGDx}) the Individual Clipping (IC) approach.
BC gives us the ability to properly train Batch Normalization Layers (BNL) \citep{ioffe2015batch} in a neural network; 
%BNL is crucial for high accuracy needed 
in  very deep neural networks the use of  BNLs is crucial for achieving high accuracy. 
The BC and ALC techniques are complimentary and can be implemented in parallel, also BC and ALC do not require any changes to be made in deep neural networks, i.e., BC and ALC apply directly without modifications to the neural network that we wish to train. This makes our training framework with BC and ALC flexible.
BC allows us to use resnet-18\citep{resnet18paper} (which uses BNLs) in our experiments and have a DP guarantee. 
%while  still maintaining the high accuracy of standard SGD without DP. 
DPSGD without BC cannot properly train BNLs, therefore, DPSGD with IC (even with ALC) for resnet-18 
%without batch normalization  
does not converge and leads to poor accuracy, while DPSGD with BC and the enhanced ALC does converge. In the context of DPSGD we are the first to introduce BC together with proving its DP guarantee; no prior related work on DPSGD with BC exists.

%Besides BC we analyse a complimentary technique for achieving high accuracy.

%== merge with above ===

%Per-layer clipping is introduced in DP-FEDAVG \citep{mcmahan2018learning} where the clipping budget is distribute equally among all model layers and proves privacy guarantee based on moment accountant. \cite{van2018three} proposed the adaptive clipping method where each layer in the neural network has its own adaptive clipping constant $C_i$. The layer clipping constants $C_i$ are computed privately before each new round, i.e., these 

%== write about BC ==

Our main contribution are:
\begin{itemize}
    \item  We propose an enhanced Adaptive Layerwise Clipping method based on~\cite{Zhang2018CoRR}. Our experiments show that DPSGD with our enhanced ALC  converges faster to 
 a higher accuracy. 
    
    % allows us to use a significantly large amount of Gaussian noise without any compromise of testing accuracy. 

    \item  We prove and characterize the DP guarantee for ALC
    %Layerwise Clipping 
    by using the $f$-DP framework~\cite{dong2019gaussian}. We explain how layerwise clipping degrades DP and we show how to set the noise parameters of DPSGD without ALC and DPSGD with ALC so that they have the same DP guarantee and their test accuracy can be fairly compared.
    
    %provide a rigorous privacy guarantee proof for all existing type of adaptive layer clipping methods based on Gaussian DP notion or $f$-DP~\cite{dong2019gaussian}.

    \item  
    We introduce and propose to use Batch Clipping during training.
    %We develop one new training mode named as batch clipping mode. Basically, this batch clipping mode 
    We also define General Batch Clipping (GBC) of which BC and IC are special cases and notice that GBC  is compatible with  \textit{any first order optimizers} in mini-batch mode.
BC allows us to implement Batch Normalization Layers which are crucial for attaining high accuracy for deep neural networks (BNLs cannot be trained properly by the original DPSGD which uses Individual Clipping). 
    
    %This type of work is never done before and significantly important. As explained in~\cite{ioffe2015batch}, Batch Normalization Layer working with a mini-batch dataset is very important to guarantee a high testing accuracy for deep neural networks. The current DPSGD works with individual clipping mode and Batch Normalization Layer is prohibited. Moreover, 
    %%%%%%% ??????? %%%%%
    %the individual clipping mode makes DPSGD only work with mini-batch type first order optimizers. We develop a rigorous DP guarantee for DPSGD with batch clipping mode. 

   \item  
   %\textcolor{red}{[TO DO -- this is commonly already understood -- We notice that it has been commonly understood that ... etc. This argument also applies to our BC and GBC approach.]} We prove and characterize the DP guarantee for BC and, more generally, GBC by using the $f$-DP framework~\cite{dong2019gaussian}. 
    For proving differential privacy guarantees it is commonly understood that the privacy argument does not depend on how the gradients in the clipped values in (\ref{algo:DPSGDx}) are computed; the to-be-clipped values may as well be computed as in (\ref{algo:DPSGDxBC}).
   % In particular 
    This shows that (the original) DPSGD with IC given by (\ref{algo:DPSGDx}) and (the new) DPSGD with BC given by (\ref{algo:DPSGDxBC}) offer the exact same DP guarantee (also if both implement ALC).

    \item Our experiments show that DPSGD modified by using  our enhanced ACL and using BC allows us to train the deep neural network resnet-$18$~\cite{resnet18paper} (which uses BNL) on CIFAR10~\cite{CIFAR10dataset} while DPSGD with ALC and IC does not converge. This shows that BC outperforms IC  in practice.

 %    and achieve the same 
 %    %prediction accuracy on testing dataset 
 %    test accuracy
 %    as a standard training method that runs classical SGD without DP (i.e., without clipping and without noise) in the same training framework. 
 %    %When the number of epochs are of 20 and 50 
 %    For 20 and 50 epochs, respectively, with
 % $\sigma=8, 10, 12, \dots, 62$, the test accuracy of our DPSGD vs standard no-DP SGD are $86\%$ vs $88\%$ and $90\%$ vs $90\%$, respectively. This result has not yet been achieved for the original DPSGD. 
    
\end{itemize}
Our main conclusion is that ACL and BC are two techniques that provide a better balance between DP and accuracy. However, our experiments are for small $\sigma$ which corresponds to weak differential privacy. We still need additional techniques beyond ALC and BC for training a deep neural network like resnet-18 with CIFAR10  in order to achieve a practical balance between test accuracy and DP guarantee.

\textbf{Outline:}
We first provide the necessary background on $f$-DP in Section \ref{sec:background}. Section \ref{sec:modifiedDPSGD} introduces BC (and GBC), layerwise clipping and our ACL, and proves DP guarantees in the $f$-DP framework.
Experiments are in Section \ref{sec:experiments}. Section~\ref{chapter:Conclusion} concludes our paper. 
%The organization of this paper is as follows. 
%We first introduce basic concept of DP, $f$-DP, DPSGD and how the norms of layers vary layer by layer in deep neural network and in this case is resnet-18 in Section~\ref{sec:background}. After that we discuss about the our proposed clipping method in details in Section~\ref{section:layerwiseclipping} with the privacy guarantee argument. We demonstrate our experiments in Section~\ref{section:Experiments} and conclude our paper in Section~\ref{sec:conclusion}. 

% In this paper, We explain the basic concepts of Privacy, Differential privacy, Deep Learning in chapters \ref{chapter:Differentialprivacy}, \ref{chapter:Deeplearning}, show how differential privacy stochastic gradient descent \cite{Abadi_2016} \cite{vandijk2023generalizing} works in chapter \ref{chapter:DPSGD} and then introduce our method in chapter \ref{section:layerwiseclipping} and finally show our experiments in section \ref{section:Experiments}.

% In this paper, we use a lot of parameters which are defined in table \ref{tbl:notations}
% \begin{table}[H]
% \centering
% \label{tbl:notations}
% \begin{tabular}{|l|l|}
% \hline
%  Parameters & Description    \\ \hline
%  $N$ & Training database size    \\ \hline
%  $K$ & Number of training iterations    \\ \hline
%  $D$ & database \\ \hline
%  $\eta$ & Learning rate \\ \hline
%  $C$ & Gradient Clipping Constant \\ \hline
%  $\sigma$ & Gaussian noise variance scaling \\ \hline
 
% \end{tabular}
% \caption{List of notations}
% \end{table}

% \section{Gaussian Differential Privacy}
% \section{Batch Clipping}
\section{\texorpdfstring{Background on 
%$(\epsilon,\delta)$ and 
$f$-DP}{}} \label{sec:background}

DP literature  first introduced $\epsilon$-DP \cite{dwork2006calibrating}, later relaxed to $(\epsilon,\delta)$-DP \citep{dwork2014algorithmic}.
In order to have a better dependency on group privacy and to improve adaptive composibility, the notion of Concentrated Differential Privacy (CDP)  \citep{CDP} was introduced.
CDP was re-interpreted and relaxed by using Renyi entropy in \citep{BS15} and its authors followed up with the notion of zero-CDP (zCDP) in \citep{zCDP}. This notion admits
simple interpretable DP guarantees for adaptive composition and group privacy. 
After the introduction of $\rho$-zCDP, Renyi DP (RDP) was introduced by \citep{RDP}.
Combining the ideas that give rise to the zCDP and RDP definitions
leads naturally to the definition of $(\rho,\omega)$-tCDP \citep{tCDP} which relaxes zCDP. 
All of these various DP measures have been superseded by $f$-DP \citep{dong2021gaussian}since (1) $f$-DP can be transformed/translated into  divergence based DP guarantees (but generally not the other way around) and can be translated into $(\epsilon,\delta)$-DP, and since (2) $f$-DP analyses the underlying core hypothesis testing problem directly and derives a {\em tight} (or exact) DP guarantee (for the adversarial model considered in the proofs of DP guarantees in literature).
%In this sense 
The $f$-DP framework  is tight and contains all the information needed to derive other known DP metrics. 
Below we summarize $f$-DP and show how to use $f$-DP to prove and formulate the DP guarantee of our modified DPSGD algorithm.

We call data sets $d=\{\xi_i\}_{i=1}^N$ and $d'=\{\xi'_i\}_{i=1}^N$ neighboring if they differ in one element;
%in : $N=|d|=|d'|$ and $|d\cap d'|=N-1$.
without loss of generality $\xi_i=\xi'_i$ for $1\leq i\leq N-1$ and $\xi_N\neq \xi'_N$ coined the {\em differentiating sample}.
In DP  a mechanism ${\cal M}$ is a process that takes either data set $d$ or data set $d'$ as input and outputs a sequence of observables which the adversary uses to distinguish which of $d$ or $d'$ has been used.  DPSGD is a mechanism which outputs a sequence of updates $U$  corresponding to each round, see (\ref{algo:DPSGDx}).
Below, we adopt the notion and explanation provided in~\cite{dong2019gaussian}
%vandijk2023generalizing}
for our short introduction of $f$-DP. We refer the reader  to \cite{dong2019gaussian}
%,vandijk2023generalizing} 
for a full description. 

\textbf{Hypothesis Testing:} From the attacker's perspective, it is natural to formulate the  problem of distinguishing two 
neighboring
data sets $d$ and $d'$ based on the output of a DP mechanism ${\cal M}$ as a hypothesis testing problem:
%
%distinguishes neighboring\footnote{$d$ and $d'$ differ in one element: $N=|d|=|d'|$ and $|d\cap d'|=N-1$.} data sets $d$ and $d'$  based on the output of a DP mechanism\footnote{DPSGD is a mechanism which outputs a sequence of updates $U$ corresponding to each round.} ${\cal M}$ as a hypothesis testing problem:
$$\mbox{versus } \begin{array}{l}
H_0: \mbox{ the underlying data set is }d \\
H_1: \mbox{ the underlying data set is }d'.
\end{array}
$$

% Here, neighboring means that either $|d\setminus d'|=1$ or $|d'\setminus d|=1$. 
% More precisely, in the context of mechanism ${\cal M}$, ${\cal M}(d)$ and ${\cal M}(d')$ take as input representations $r$ and $r'$ of data sets $d$ and $d'$ which are `neighbors.' The representations are mappings from a set of indices to data samples with the property that if $r(i)\in d\cap d'$ or $r'(i)\in d\cap d'$, then $r(i)=r'(i)$. This means that the mapping from indices to data samples in $d\cap d'$ is the same for the representation of $d$ and the representation of $d'$. In other words the mapping from indices to data samples for $d$ and $d'$ only differ for indices corresponding to the differentiating data samples in $(d\setminus d')\cup (d'\setminus d)$. In this sense the two mappings (data set representations) are neighbors.
% In our main theorem we will consider the general case $g=\max\{ |d\setminus d'|, |d'\setminus d|\}$ in order to analyse `group privacy.' 

We define the Type I and Type II errors by
$$\alpha_\phi = \mathbb{E}_{o\sim {\cal M}(d)}[\phi(o)]  \mbox{ and } \beta_\phi = 1- \mathbb{E}_{o\sim {\cal M}(d')}[\phi(o)],
$$
where $\phi$ in $[0,1]$ denotes the rejection rule which takes the output of the DP mechanism as input. We flip a coin and reject the null hypothesis with probability $\phi$. The optimal trade-off between Type I and Type II errors is given by the trade-off function
$$ T({\cal M}(d),{\cal M}(d'))(\alpha) = \inf_\phi \{ \beta_\phi \ : \ \alpha_\phi \leq \alpha \},$$ 
for $\alpha \in [0,1]$, where the infimum is taken over all measurable rejection rules $\phi$. If the two hypothesis are fully indistinguishable, then this leads to the trade-off function $1-\alpha$. We say a function $f\in [0,1]\rightarrow [0,1]$ is a trade-off function if and only if it is convex, continuous, non-increasing, at least $0$, and $f(x)\leq 1-x$ for $x\in [0,1]$. 
We define a mechanism ${\cal M}$ to be $f$-DP if $f$ is a trade-off function and for all neighboring $d$ and $d'$,
$$
 T({\cal M}(d),{\cal M}(d')) \geq f.
$$
%
% The $f$-DP framework supersedes all existing other frameworks in that a trade-off function contains all the information needed to derive known DP metrics such as $(\epsilon,\delta)$-DP and divergence based DPs. 
%
%\citep{dong2021gaussian} defines Gaussian DP as a special case of $f$-DP where $f$ is a trade-off function
%$$G_\mu(\alpha) = T({\cal N}(0,1),{\cal N}(\mu,1))(\alpha) = \Phi( \Phi^{-1}(1-\alpha) - \mu )$$
%with $\Phi$ the standard normal cumulative distribution of ${\cal N}(0,1)$. 
%
%--------------
%

\textbf{Gaussian DP:}
Gaussian DP is defined as a special case of $f$-DP where $f$ is  defined as a trade-off function
$$G_\mu(\alpha) = T({\cal N}(0,1),{\cal N}(\mu,1))(\alpha) = \Phi( \Phi^{-1}(1-\alpha) - \mu ),$$
for some $\mu\geq 0$, where $\Phi$ is the standard normal cumulative distribution of ${\cal N}(0,1)$. 
%We define a mechanism to be $\mu$-Gaussian DP if it is $G_\mu$-DP. 
%Corollary 2.13 in \citep{dong2021gaussian} shows  that a mechanism is $\mu$-Gaussian DP if and only if it is $(\epsilon, \delta(\epsilon))$-DP for all $\epsilon\geq 0$, where
%\begin{equation} \delta(\epsilon) = \Phi(-\frac{\epsilon}{\mu}+\frac{\mu}{2}) - e^{\epsilon} \Phi(-\frac{\epsilon}{\mu}-\frac{\mu}{2}).
%\label{eq:gdp}
%\end{equation}
%
Suppose that a mechanism ${\cal M}(d)$ computes some function $u(d)\in \mathbb{R}^n$ and adds Gaussian noise ${\cal N}(0,(c\sigma)^2{\bf I})$, that is,  the mechanism outputs $o\sim u(d)+{\cal N}(0,(C\sigma)^2{\bf I})$. Suppose that $c$ denotes the sensitivity of function $u(\cdot)$, that is, $$\|u(d)-u(d')\|\leq c$$ 
for neighboring $d$ and $d'$; the mechanism corresponding to one round update in DPSGD, where ${\tt Sample}_m(N)$ selects (as one of the $m$ randomly selected indices) the index $N$ of the differentiating sample, 
has {\em sensitivity} $c=2C$. After projecting the observed $o$  onto the line that connects $u(d)$ and $u(d')$ and after normalizing by dividing by $c$, we have that differentiating whether $o$ corresponds to $d$ or $d'$ is in the best case for the adversary (i.e., $\|u(d)-u(d')\|=c$) equivalent to differentiating whether a received output is from ${\cal N}(0,\sigma^2)$ or from  ${\cal N}(1,\sigma^2)$. Or, equivalently, from ${\cal N}(0,1)$ or from  ${\cal N}(1/\sigma,1)$. 
This is how the Gaussian trade-off function $G_{\sigma^{-1}}$ comes into the picture. 

\textbf{Subsampling:} Besides implementing Gaussian noise which bootstraps DP,   DPSGD also uses ${\tt Sample}_m$ for subsampling  which amplifies DP.
\citep{dong2021gaussian} defines a subsampling operator $C_{m/N}$ and  shows that if a mechanism ${\cal M}$ on data sets of size $N$ is $f$-DP, then the subsampled mechanism ${\cal M}\circ {\tt Sample}_{m}$ is $C_{m/N}(f)$-DP. We have that the mechanism corresponding to one round in DPSGD is $C_{m/N}(G_{1/\sigma})$-DP (and this is a tight analysis).

\textbf{Composition:} If DPSGD implements $T$ rounds, then the privacy leakage across rounds composes. 
\citep{dong2021gaussian} defines a commutative tensor product $\otimes$ for trade-off functions and shows this can be used to characterize adaptive composibility:
Let ${\cal M}_i$ be the mechanism corresponding to the $i$-th round with $y_i \leftarrow {\cal M}_i(\texttt{aux},d)$ where $\texttt{aux}=(y_1,\ldots, y_{i-1})$ (this captures adaptivity). If ${\cal M}_i(\texttt{aux},.)$ is $f_i$-DP for all $\texttt{aux}$, then the composed mechanism ${\cal M}$, which applies ${\cal M}_i$ in sequential order from $i=1$ to $i=T$, is 
$(f_1\otimes \ldots \otimes f_T)$-DP.
%$f^{\otimes T}$-DP.
%The tensor product is commutative.
This leads to a tight analysis of DPSGD.
We have  that DPSGD 
%(with subsampling, individual clipping and mini-batch SGD) 
as introduced in \cite{abadi2016deep} 
%with (the slightly more general) update $U$ defined in (\ref{eq:ind}) before adding Gaussian noise
%in Algorithm \ref{alg:??} [TO DO: work with $2\sigma$ in pseudo codes as in $f$-DP paper] 
is
%[TO DO -- we now use $|S_b|=ms$ instead of $s$ here]
%\footnote{TO DO: In appendix explain where DP-SGD paper goes wrong. Their DP-SGD algorithm uses noise ${\cal N}(0,C^2(2\sigma)^2 {\bf I})$ compared to ${\cal N}(0,C^2\sigma^2 {\bf I})$ in our version of the DP-SGD algorithm.}
$$ C_{m/N}(G_{1/\sigma})^{\otimes T}\mbox{-DP}.$$
Notice that if DPSGD computes a total of $E$ epochs of gradients, i.e., $EN$ gradient computations in total, then $T=(N/m)\cdot E$ (since each round computes $m$ gradients). We have
\begin{equation} C_{m/N}(G_{1/\sigma})^{\otimes (N/m)\cdot E}\mbox{-DP}.
\label{eq:DPIC}
\end{equation}

\section{Application $f$-DP toward a Modified DPSGD}
\label{sec:modifiedDPSGD}

We discuss two main  DPSGD modifications. The first is coined Batch Clipping (BC) and the second Layerwise Clipping (LC) leading to Adaptive LC (ACL).

\subsection{Batch Clipping}
\label{sec:BC}

The presented $f$-DP analysis is more general in that it holds for (\ref{algo:DPSGDx}) where $U$ is not just computed as a noised sum of clipped {\em gradients} but  computed as
\begin{equation}
U := n+\sum_{j=1}^m [g(w;\xi_{i_j})]_C
\label{eq:g}
\end{equation}
for some other fixed function $g$. 

Suppose that we partition data set $d$ of size $N$ into $N/s$ mini-sets of size $s$ each. We use this to define a new data set $d_s$ which has as elements the $N/s$ mini-sets, which we denote as $S_i$, $1\leq i\leq N/s$. Data set $d_s$ has size $N/s$ and its samples are mini-sets $S_i$ of size $s$. Each $S_i$ contains $s$ data points $\xi_j$ from $d$.
We apply DPSGD  to this new data set $d_s$ for the general (\ref{eq:g}) where we replace $m$ by $k$. This yields 
\begin{equation}
U := n+\sum_{j=1}^k [g(w;S_{i_j})]_C.
\label{eq:g1}
\end{equation}
%where mini-sets $S_{i_j}$ have size $s$. We call this General Clipping (GC).
We call this General Batch Clipping (GBC) since we clip vectors $g(w;S_{i_j})$ which are computed based on a batch (mini-set) $S_{i_j}\subseteq d$ of data points from $d$. Notice that in GBC, $g$ can implement any moment based SGD type algorithm that iteratively scans the data points in $S_{i_j}$.
%(while treating $w$ as a local variable within $g$'s evaluation and  updating the local $w$ from).

Applying the $f$-DP analysis for DPSGD for a data set $d_s$ of size $N/s$ with ${\tt Sample}_k$, see (\ref{eq:g1}), yields
$C_{k/(N/s)}(G_{1/\sigma})^{\otimes T}\mbox{-DP}$, see Section \ref{sec:background}. By setting $m=sk$, we conclude $C_{m/N}(G_{1/\sigma})^{\otimes T}\mbox{-DP}$.
Notice that if the modified DPSGD with GBC computes a total of $E$ epochs of gradients, then again $T=(N/m)\cdot E$ since each round still computes $m=sk$ gradients. For GBC with $m=sk$ we conclude the exact same DP guarantee as the one for 
%IC in 
(\ref{eq:DPIC}).

In our experiments we use the special case
\begin{equation} g(w;S_{i_j}) = \frac{1}{s} \sum_{\xi \in S_{i_1}} \nabla_w
%% Do not change to f_w !!!!!!!
f(w;\xi) \label{eq:g2}
\end{equation}
with $k=1$ 
%in (\ref{eq:g1}) 
and $s=m$ in GBC (with $m=sk$).
We refer to this as (non-general) Batch Clipping (BC) (since we clip $g(w;S_{i_1})$ which computes and averages a batch of gradients), see (\ref{algo:DPSGDxBC}). We call the original DPSGD formulation (\ref{algo:DPSGDx}) Individual Clipping (IC) since single/individual gradients are clipped; for completeness, this is the case $s=1$ with $k=m$ in GBC (with $m=sk$).

%When referring to BC in our experiments, we consider the special case $k=1$ with $s=m$, that is, (\ref{eq:g1}) consists of a single computation (\ref{eq:g2}).

\subsection{Layerwise Clipping}
\label{sec:layerwiseclipping}

%Based on the main idea in we propose a 
Layerwise Clipping (LC)~\cite{mcmahan2018learning,van2018three,Xu2021,Zhang2018CoRR} splits vectors $g=g(w;S_{i,j})$ in parts $g=(g_1 | \ldots | g_L)$ (i.e., $g$ is equal to the concatenation of parts $g_1$, $\ldots$, $g_L$) and clip each part $g_h$ separately, i.e., we define
$$
[g]_{(C_1,\ldots,C_L)}=
([g_1]_{C_1} | \ldots | [g_L]_{C_L}).
$$
The different $g_i$ correspond to the different layers in the neural network. We compute noise $n$ in (\ref{eq:g1}) as the concatenation
$$
n = (n_1 | \ldots | n_L) \ \ \ \ \mbox{ with } \ \ \ \ n_h\sim {\cal N}(0,(2C_h\sigma){\bf I}),
$$
where the different matrices ${\bf I}$ have sizes that correspond to the number of entries in the various $g_h$. 

In order to understand how the DP guarantee is affected, we rewrite (\ref{eq:g1}) as follows:
$$
U:=(U_1 | \ldots | U_L) \ \ \ \ \mbox{ where } \ \ \ \ U_h=n_h + \sum_{j=1}^k [g_h(w;S_{i_j})]_{C_h}.
$$
In other words, transmission of $U$ is equal to transmitting $U_h$, $1\leq h\leq L$. Each $U_h$ can be considered as a round update where we use clipping constant $C_h$ and noise ${\cal N}(0,(2C_h\sigma){\bf I})$. The $f$-DP analysis shows that such a round is $G_{1/\sigma}$-DP (see the explanation of Gaussian DP with sensitivity $c=2C_h$). We have $L$ such sub-rounds that make up the whole transmission of $U$. By composition, we have that this gives $G_{1/\sigma}^{\otimes L}$-DP. \citep{dong2021gaussian} shows that\footnote{We can also vary the noise from layer to layer and use $\sigma/p_h$ instead of $\sigma$. This leads to $G_{p_1/\sigma}\otimes \ldots \otimes G_{p_L/\sigma}=G_{ \sqrt{\sum_{h=1}^L p_h^2}/\sigma}$.} $G_{1/\sigma}^{\otimes L}=G_{\sqrt{L}/\sigma}$.

We conclude that in the DP guarantee (\ref{eq:DPIC}) we need to replace $\sigma$ by $\sigma/\sqrt{L}$ for the modified DPSGD with BC/IC and LC. If we want to compare this in a fair way with the modified DPSGD with BC/IC and {\em no} LC (we keep the original gradient clipping approach), then we should use 
$$ \bar{\sigma}:=\sigma/\sqrt{L}$$
as the privacy parameter $\sigma$ in the modified DPSGD (as this will result in the same (\ref{eq:DPIC}) with $\sigma$ replaced by $\bar{\sigma}=\sigma /\sqrt{L}$).

Clearly, we want to be careful about the number of layers $L$ we can handle since the privacy parameter $\sigma$ is divided by $\sqrt{L}$. We notice that we can group layers together and split vectors $g=g(w;S_{i,j})$ into a smaller number of parts giving a smaller $L$ which leads to a better DP guarantee.

\subsection{Adaptive LC}
\label{subsec:adaptiveclipping}

\begin{figure}[!ht]
    \centering
\includegraphics[width=01.0\textwidth]{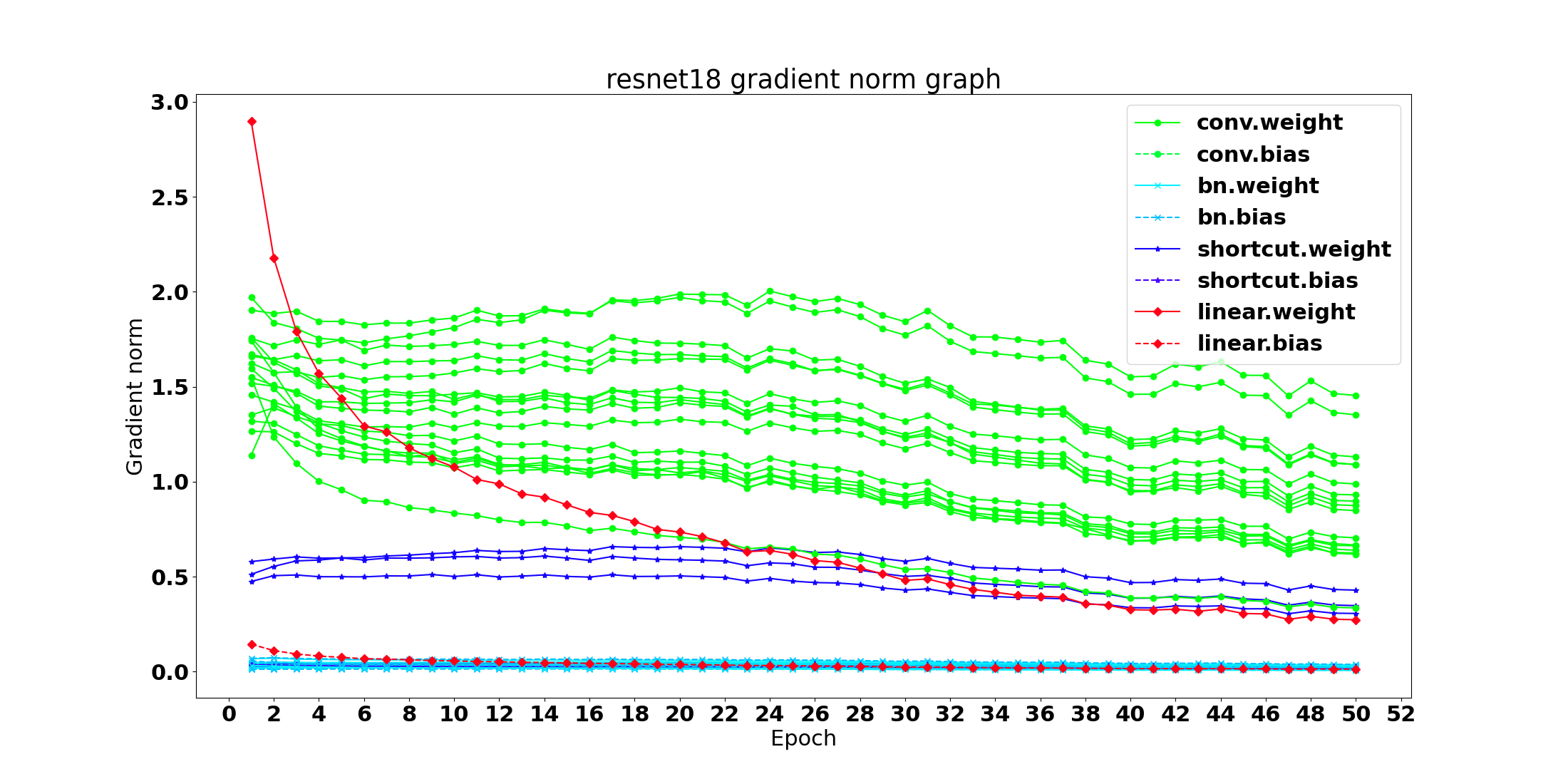}
    \caption{resnet-18 model layerwise gradient norms. We grouped the lines by the layer's type: "conv" is convolutional layer, "bn" is the batch normalization layer, "shortcut" is the shortcut connecting non-adjacent two layers, "linear" is the fully connected layer (in this case this layer corresponds to a softmax layer)}\label{fig:resnet18gradientnorm}
\end{figure}

% Tune C initially -- can make it pretty small
% C_i are decaying -- improves over Zhang
% Zhang multiplies norm with sigma --> large noise --> bad
% WE decaying C_i, hence
% less absolute noise, but the same relative noise
% sigma smaller than 60 no difference

The Adaptive LC (ALC) of~\cite{Zhang2018CoRR} uses a public dataset $\mathcal{D}_{pub}$ to estimate expectations of the layer gradient norms $\|\nabla_{w_h} f(w;\xi)\|$. These estimates are used as the layerwise clipping constants $C_h$.
%
%$C_1, \dotsc, C_L$. See 
%Figure~\ref{fig:resnet18gradientnorm},  our main observation is that the 
For standard SGD without DP, \cite{KaiHu2022} explains that even if model $w$ converges, the  gradient norms of different layers may not decrease throughout the training. We confirm this in
Figure~\ref{fig:resnet18gradientnorm}, which depicts layer gradient norms of resnet-18 trained without DP over 50 epochs. We observe that most layer gradient norms only slowly decrease from epoch to epoch.   Therefore, most of the $C_h$ in the ALC of~\cite{Zhang2018CoRR}
are only slightly adapted from epoch to epoch.
%which adapts each $C_i$ according to  the estimated expectation of the corresponding layer gradient norm,
%most of $C_1, \dotsc, C_L$ do not change much from epoch to epoch
This implies that, for such a layer $h$, the distribution ${\cal N}(0,(2C_h\sigma)^2{\bf I})$ of the Gaussian noise added to each of the weight entries in layer $h$  does not change much either. This setup does not allow the designer to optimize the clipping constants $C_1, \dotsc, C_L$ to make  convergence faster.

%This may hurt the convergence of DPSGD when $\sigma$ is large ($>60$) as shown in our experiments in Figure~\ref{fig:ZhangversusUs} in Section~\ref{subsec:ZhangUs}. 

We enhance the adaptive clipping strategy of ~\cite{Zhang2018CoRR}:
%from round to round proposed in in our modified DPSGD, 
%which consists of two parts. 
First, we determine a master clipping constant $C$ for each round. We do not impose a restriction on how $C$ is chosen, i.e., it can diminish from epoch to epoch  or $C$ can be the same constant for all rounds.  In our experiments $C$ is a constant throughout the whole training. Second, we use $C$ to derive clipping values $(C_1, C_2, ..., C_L)$ for the corresponding round. This has the property that each $C_h$ scales linearly with $C$.

Given a master clipping constant $C$ at the beginning of each round, we use a {\em public} dataset $\mathcal{D}_{pub}$ to derive clipping constants $(C_1, C_2, ..., C_L)$.
The reason for using a public dataset is that we do not need to worry about privacy leakage revealed by $(C_1, C_2, ..., C_L)$; DP analysis/proofs, where the adversary knows $(C_1, C_2, ..., C_L)$, may proceed as before. We estimate the expectation $e=(e_1, \ldots, e_L)$ of the layer gradient norms $(\|\nabla_{w_1} f(w;\xi)\|, \ldots, \|\nabla_{w_L} f(w;\xi)\|)$ over $\xi \in \mathcal{D}_{pub}$ for $w=(w_1 | \ldots | w_L)$. We compute the maximum gradient norm among all layers, i.e., $M = \max_{h=1}^{L} e_h$. Then, for each layer $h$ we define $C_h = C \cdot e_h/M$. See Section \ref{sec:Zhang}, Figure~\ref{fig:ZhangversusUs} shows that our enhanced ALC allows DPSGD with BC to converge faster to a higher accuracy compared to DPSGD with BC and  the original ALC of ~\cite{Zhang2018CoRR}.

\subsection{Sparsification}
\label{sec:sparse}

We notice that the public data set $\mathcal{D}_{pub}$ in adaptive clipping can also be used to find out whether certain weight entries in $w$ have converged sufficiently. That is, we say a weight entry has converged if, over a large number of recent rounds, it hoovers around an average with standard deviation corresponding to the added DP noise. As soon as this is the case, we may fix the weight entry to this average in all future computations/rounds (since this is a form of post processing, no additional DP leakage occurs). This reduces the number of weight entries over which we need to compute gradients and each level gets less weight entries as soon as convergence sets in. This allows us to use even smaller clipping constants per layer (depending on the number of active weight entries in the layers). We leave this optimization as an open problem.

\section{Experiments}
\label{sec:experiments}

\subsection{Setup}
\label{sec:setup}

\textbf{Data preprocessing:} 
%In this paper, 
We perform 
experiments on the CIFAR-10 dataset which consists of 50,000 training examples and 10,000 test examples, divided into 10 classes, with each example being a 32x32 image with three color channels (RGB)(\cite{CIFAR10dataset}). In our experiments, we perform data augmentation and data normalization independently. Specifically, for each training image, we crop a $32 \times 32$ region from it with padding of 4, apply a random horizontal flip to the image, and then normalize it with 
\[(mean,std) = ((0.4914, 0.4822, 0.4465), (0.2023,0.1994, 0.2010)).\]
Next, we divide the CIFAR-10 dataset $\mathcal{D}$ into two dataset $\mathcal{D}_{pub}$ and $\mathcal{D}_{train}$ where $|\mathcal{D}_{pub}| = \frac{1}{10}|\mathcal{D}|$  and $|\mathcal{D}_{train}| = \frac{9}{10}|\mathcal{D}|$.

As explained in Section \ref{subsec:adaptiveclipping}, 
%similar to \cite{Zhang2018CoRR}, 
we  estimate the expectation $(e_1,\ldots,e_L)$  of 
the layer gradient norms 
$(\|\nabla_{w_1} f(w;\xi)\|,\ldots, \|\nabla_{w_L} f(w;\xi)\|)$ over\footnote{We notice that we may not need a large sized ${\cal D}_{pub}$ for a good estimate.} $\xi \in {\cal D}_{pub}$, where $w=(w_1 | \ldots | w_L)$ is the current model. We compute the layerwise clipping constants $(C_1=C\cdot e_1/M,\dots,C_L=C\cdot e_L/M)$ with $M=\max_{h=1}^L e_h$ and master clipping constant $C$ .

%based on gradients computed from the public dataset $\mathcal{D}_{pub}$ and the current model $w$. %'s parameters. 

We sample the $m$-sized training data batches with replacement from $\mathcal{D}_{train}$ at the beginning of each epoch and  feed them to the machine learning model to train the model.

\textbf{Diminishing Learning Rate and Fixed Master Clipping Constant:} We fix the master clipping constant $C$ throughout the training process and after each epoch we update the learning rate $\eta$ as 
\begin{eqnarray*}
\eta &:=&  \eta \cdot \eta_{decay},
\end{eqnarray*}
%
% \begin{eqnarray*}
% \eta &:=&  \eta \cdot \eta_{decay}
% \ \ \ \ \mbox{ and } \ \ \ \
% C :=  C \cdot C_{decay},
% \end{eqnarray*}
where $\eta_{decay}$ is a decaying factor. After each epoch, we re-compute the layerwise clipping constants $(C_1,C_2,\dots,C_L)$ as explained above. In our experiments, we set $\eta_{decay} = 0.9$, $C = 0.095$ and $\sigma=0.01875$.

\textbf{Model Update:} resnet-18 is updated by our modified DPSGD with BC and ALC,
%as described in (\ref{algo:DPSGDxBC}), 
i.e., for a batch (mini-set) $\{\xi_{i_1},\ldots, \xi_{i_m}\}$ we compute
\begin{eqnarray*}
U &:=& n+\left[\frac{1}{m} \sum_{j=1}^m  \nabla_w f(w;\xi_{i_j}) \right]_{(C_1,\dotsc,C_L)}
\ \ \ \ \mbox{ and } \ \ \ \
    w := w -  \eta U
\end{eqnarray*}
with 
$n\sim {\cal N}(0,(2C_1\sigma)^2{\bf I})\times \ldots \times {\cal N}(0,(2C_L\sigma)^2{\bf I})$.

% In order to demonstrate our idea, we first train the light-weighted convnet model \ref{tab:convnet} with batch size $m=64$ and print out per layer gradient average norm after each epoch, which is shown in Figure \ref{fig:gradientgraph}.a. Here, we can clearly see the different between layers gradient norm magnitudes so we decide to apply our method, layerwise clipping and diminishing clipping constant $C_{master}$, instead of full gradient clipping in \cite{abadi2016deep}.

% Moreover, our method outperforms full gradient clipping method when we apply both method to train the convnet model with individual clipping under the same hyper-parameters settings: Specifically, we choose, noise multiplier $\sigma= 1.5$, constant clipping value $C =1.2$, starting dimishing clipping value $C_{master}= 1.2$, batch size $m = 64$, number of epochs $E=50$, diminishing learning rate $\eta = 0.025$ where $m$ is the mini-batch size and $C_{master}$ decreases after each epoch by $10\%$. The comparison between two methods is shown in \ref{fig:gradientgraph}.b.

% Note that batch clipping method enable us to apply batch normalization layer in the model architecture. 
% However, the convnet model does not have the batch normalization layer so we conduct all experiments with resnet18 model (\cite{resnet18paper}).

\subsection{Benchmark versus BC}
\label{subsec:benchmarkvsBC}

We compare our DPSGD with BC and ALC versus mini-batch SGD without DP, i.e., without clipping and without adding Gaussian noise. In both cases, we train resnet-18 on the CIFAR-10 dataset.

%Firstly, we compare how good is our method, batch clipping and diminishing master clipping value $C_{master}$, with the no DP configuration where we use mini-batch SGD to train the resnet-18 model on CIFAR-10 dataset without clipping the gradients and adding the Gaussian noise. 

\begin{figure}[ht]
    \centering
\includegraphics[width=0.75\textwidth]{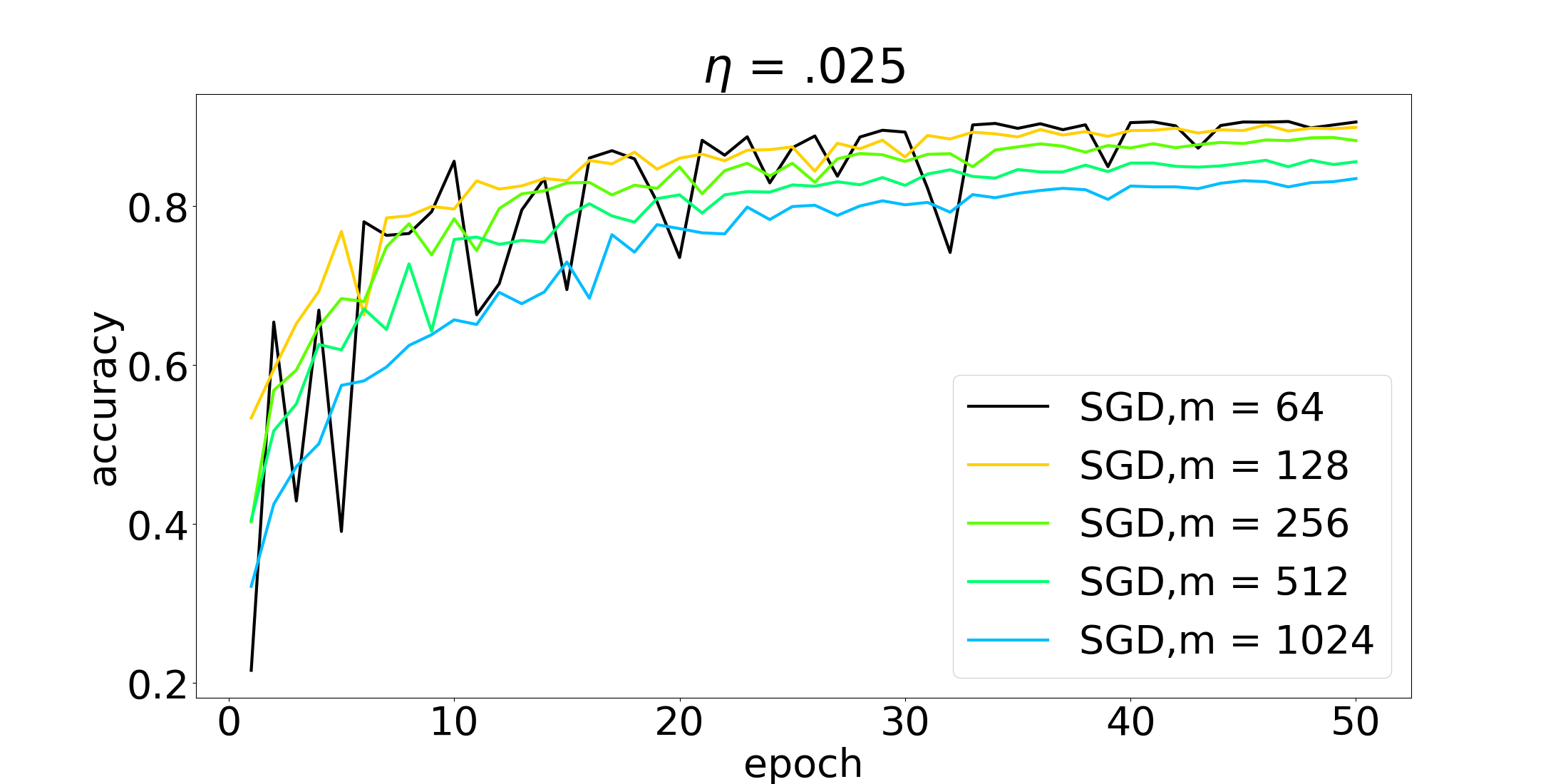}
    \caption{
    Test accuracies of mini-batch SGD without DP for different mini-batch sizes $m=(64,128,256,512,1024)$}   \label{fig:BCaccuracyA}
\end{figure}

\begin{figure}[ht]
    \centering
\includegraphics[width=0.75\textwidth]{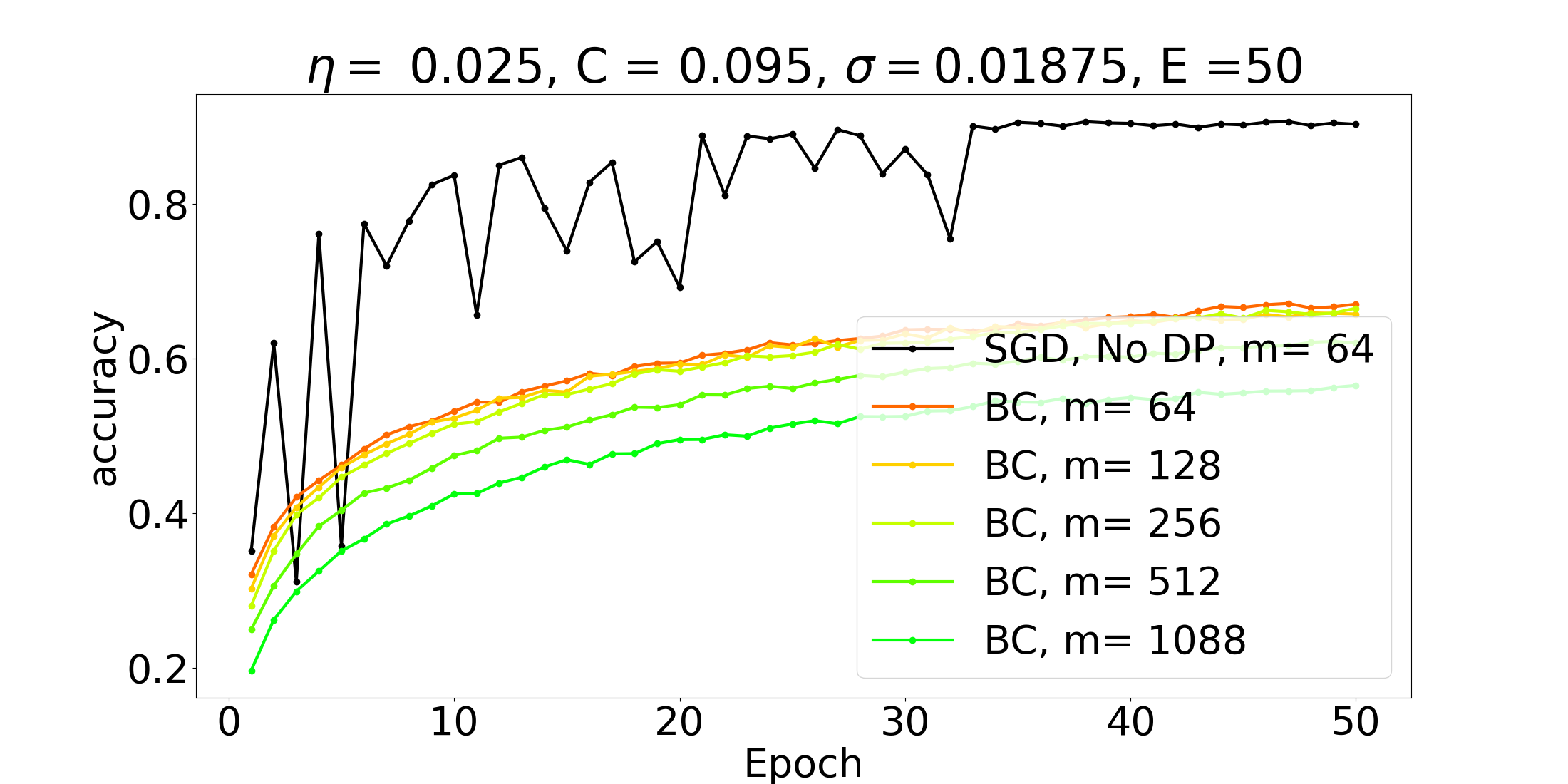}
    \caption{
    Test accuracies of DPSGD+BC+ALC vs mini-batch SGD without DP for different mini-batch sizes $m=(64,128,256,512,1024)$}   \label{fig:BCaccuracyB}
\end{figure}

For mini-batch SGD with diminishing learning rate without DP we experiment with
%train
%resnet-18 
%model with normal training (no clipping, no added noise) with 
mini-batch sizes $m = (64,128,256,512,1024)$ and initial learning rate $\eta = 0.025$.  The best test accuracy
%and achieved the best testing accuracy 
at epochs 20 and 50 is realized by mini-batch size $m=64$ and achieves $88.31\%$ and $90.24\%$ (see Figure~\ref{fig:BCaccuracyA}).
%, respectively.
%, respectively, is : $88.31\%$ and $90.24\%$ with batch size $m = 64$. 
%Next, we also run the 
The results of our DPSGD with BC and ALC and diminishing learning rate, with fixed master clipping constant $C=0.0095$, with standard deviation $\sigma=0.01875$, and with $m = (64,128,256,512,1024)$ and initial learning rate $\eta = 0.025$ are presented in Figure~\ref{fig:BCaccuracyB}. We  achieve $60\%$ and $67\%$ at epochs 20 and 50  for $m=64$.

%\begin{figure}[!ht]
%    \centering
%    \subfigure[Mini-batch SGD without DP]{\includegraphics[width=0.49\textwidth]{Figs/BC_IC_acc/resnet18_subsampling_SGD_lr_0.025_C_0.05_sigma_2_0.png}}
%    \subfigure[DPSGD+BC+ALC vs mini-batch SGD without DP]{\includegraphics[width=0.49\textwidth]{Figs/BC_IC_acc/BCversusBenchmark.png}}
%    \caption{
%    %Batch clipping 
%    Test accuracies for different mini-batch sizes $m=(64,128,256,512,1024)$}
%    \label{fig:BCaccuracy}
%\end{figure}

\subsection{IC versus BC}
We compare our DPSGD with BC and ALC versus DPSGD with IC and the same ALC. The latter is the original DPSGD \cite{abadi2016deep} combined with ALC.
%\cite{van2018three,Xu2021,Zhang2018CoRR}.
%Beside that, we also wonder how good is our batch clipping method compared with the individual clipping method which in widely used in many literature in (\cite{abadi2016deep},\cite{van2018three},\cite{Xu2021},\cite{Zhang2018CoRR}). 
Based on the result from Section \ref{subsec:benchmarkvsBC}, we choose 
%the mini batch size 
$m=64$ which gives  the best test accuracy for fixed master clipping constant $C=0.0095$ and initial learning rate $\eta=0.025$. We report the test accuracies after $E=50$ epochs for 
$\sigma = 0.01875$ in Figure \ref{fig:ZhangversusUs}. We observe that BC  converges while IC does not.
%
%Here, we choose learning rate $\eta = 0.025$, learning decaying $\eta_{decay}=0.9$, diminishing master clipping value $C_{master} = 0.005$, number of epochs $E=20$ and we vary the noise multiplier value $\sigma = 2,4,6,\dots, 60$. According to Figure \ref{fig:BCICcomparison}.a, 
%we can see that batch clipping outperformed individual clipping method by having better accuracy. 
% We zoom in on $\sigma=60$ in Figure \ref{fig:BCICcomparison}.b
% %, we choose $\sigma = 60$, then 
% where we achieve $85.97\%$ test accuracy for BC and $60.80\%$ test accuracy for IC.
% \begin{figure}[!ht]
%     \centering
%     \subfigure[Test accuracy for $\sigma=(2,4,6,\ldots, 60)$]{\includegraphics[width=0.49\textwidth]{Figs/Varying_sigma/resnet_BC_IC_Varying_Sigma.png}}
%     \subfigure[Test accuracy from epoch to epoch]{\includegraphics[width=0.49\textwidth]{Figs/BC_IC_acc/resnet18_BC_IC_accuracy.png}}
%     \caption{Comparison of Batch Clipping and Individual Clipping test accuracies}
%     \label{fig:BCICcomparison}
% \end{figure}

\subsection{Zhang's ALC versus ours}
\label{sec:Zhang}

As discussed in Section~\ref{subsec:adaptiveclipping}, the layerwise clipping constants $C_1,\dotsc,C_L$ in~\cite{Zhang2018CoRR} may not change from epoch to epoch because the layer gradient norms of resnet18 do not significantly change throughout the training process as depicted in Figure~\ref{fig:resnet18gradientnorm}. 
We explained that this may hurt the convergence if $C_1,\dotsc,C_L$ are not optimized.
%of training algorithm if the noise variance $\sigma$ is sufficient large. 
Based on this observation, our  proposed enhanced ALC implements a master clipping constant $C$ which allows us to optimize layer clipping constants $C_1, \dotsc, C_L$. 
%We show that 
By tuning the initial/fixed master clipping constant 
%manipulating 
$C$, DPSGD with BC and our enhanced ALC offers a better performance compared to
DPSGD with BC and the ALC method of ~\cite{Zhang2018CoRR}.

%We observe that our enhanced ALC
%allows us to choose a very small initial master clipping constant $C=0.005$. This in turn leads to a wide range of possibles choices for 
%$\sigma$. We notice  that, the Gaussian noise added  to the gradient of layer $h$ is drawn from $N(0,(2C_h\sigma)^2{\bf I})$. Experiments seem to confirm that 
%we can keep 
%
%increase $\sigma$ at least to the point where $C^2\sigma^2 = 1$ without hurting the model's gradient too much.

\begin{figure}[ht]
    \centering
\includegraphics[width=0.75\textwidth]{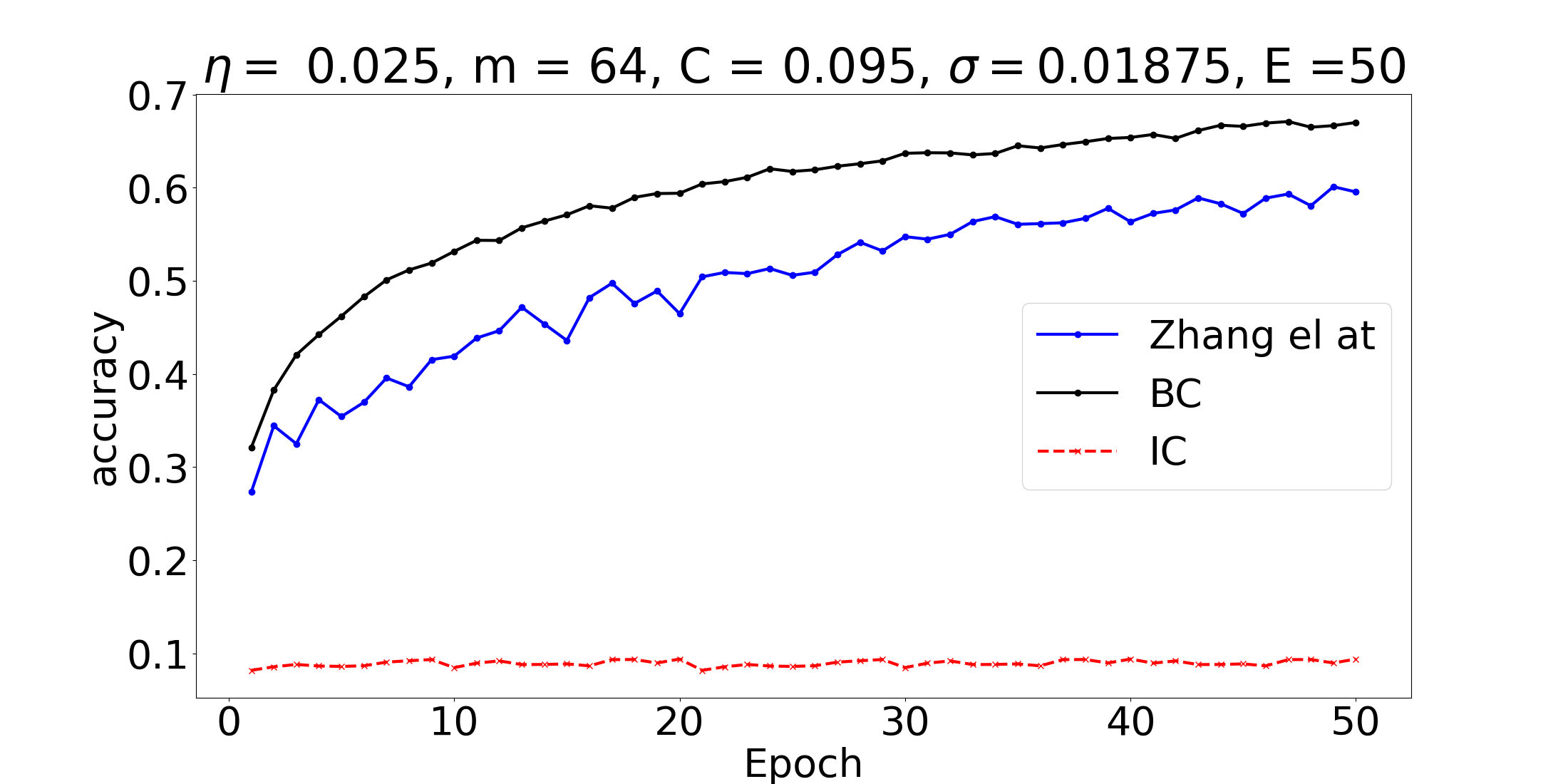}
    \caption{Comparison of the test accuracies between our enhanced ALC and the ALC method of \cite{Zhang2018CoRR}, both with BC. Moreover, we run our enhanced ALC in IC mode 
    %(with $\bar{\sigma}=\sigma/\sqrt{L}$ for $L=62$ layers in resnet-18, see Section \ref{sec:layerwiseclipping}) 
    so that we can also compare BC versus IC. ("Zhang et al." denotes DPSGD + BC + ALC of \cite{Zhang2018CoRR}; "BC" denotes DPSGD + BC + our enhanced ALC; "IC" denotes DPSGD + IC + our enhanced ALC)}
    \label{fig:ZhangversusUs}
\end{figure}

We run DPSGD with BC and the our enhanced ALC method  with initial learning rate  $\eta = 0.025$, fixed master clipping constant $C = 0.0095$, and $m=64$
%\eta_{decay} = 0.9$ and $C_{decay} = 0.9$ 
over 50 epochs for resnet-18 and CIFAR-10.
We run DPSGD with BC and the ALC of \cite{Zhang2018CoRR} with the same learning rate  $\eta = 0.025$ and $m=64$ over 50 epochs for resnet-18 and CIFAR-10.
%We work with very large $\sigma$ because our ALC and Zhang's one give good testing accuracy when $\sigma$ are smaller than 60. 
The results are shown in Figure~\ref{fig:ZhangversusUs}.

% Therefore, we investigate how large the noise multiplier $\sigma$ we can achieve until we witness the degradation in test accuracy while comparing with the method in \cite{Zhang2018CoRR}:
% {\color{red} TODO: Should we add the comparison image for small sigma (2 to 64)?}

\label{subsec:ZhangUs}

\subsection{Different noises}
\begin{figure}[!ht]
    \centering
\includegraphics[width=0.75\textwidth]{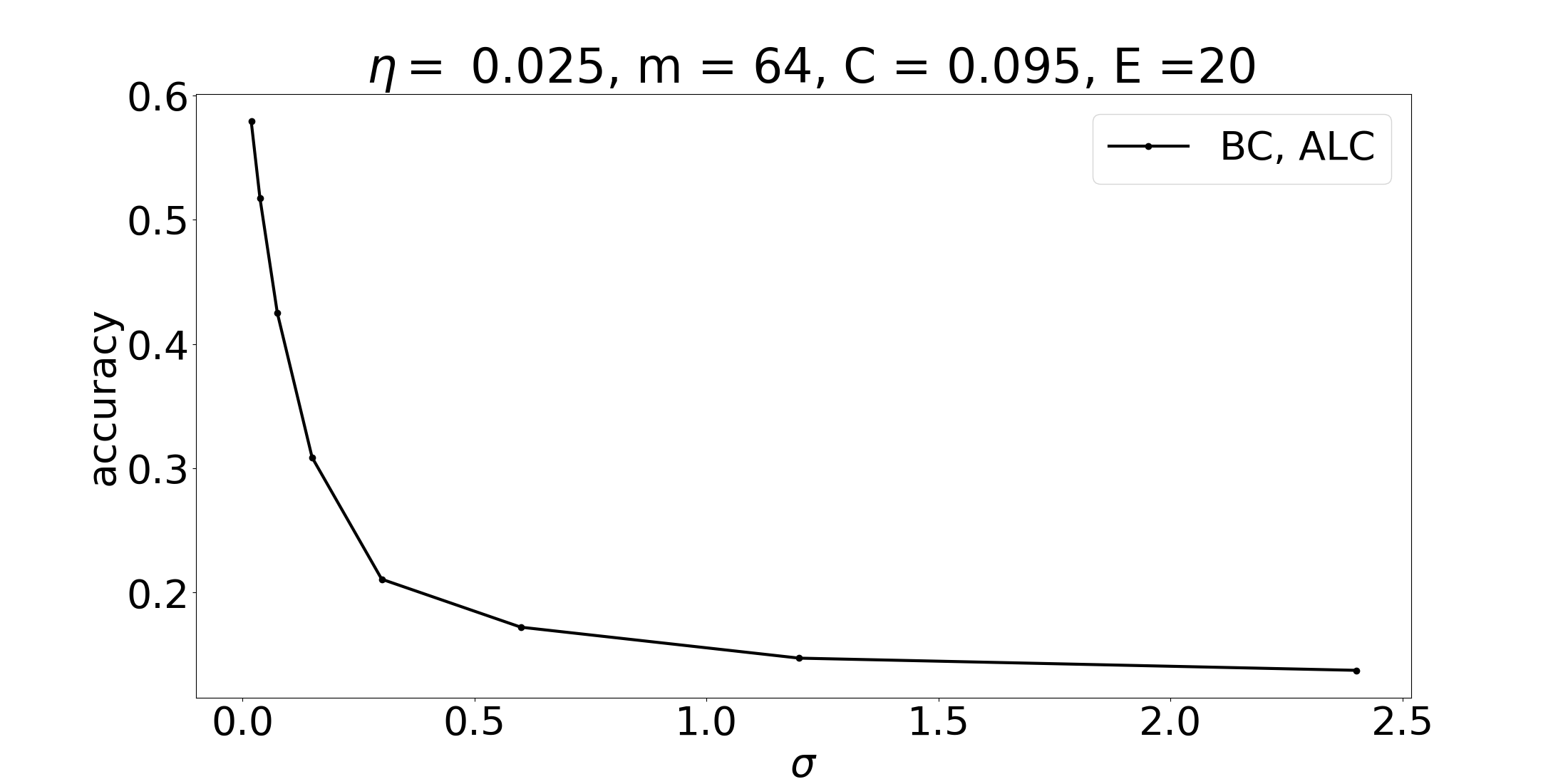}
    \caption{The test accuracy of the modified DPSGD with BC and enhanced ALC for different $\sigma$ over 10 epochs.}
    \label{fig:ourDPSGDwithDiffNoises}
\end{figure}

We study the relationship between the test accuracy and $\sigma$ 
for DPSGD with BC and the enhanced ALC
with $\eta=0.025$, $m=64$, and $C=0.095$ over 10 epochs. The result is depicted in Figure~\ref{fig:ourDPSGDwithDiffNoises} and shows that for convergence resnet-18 needs a relatively small $\sigma$ and as a consequence can only achieve a weak DP guarantee. 
This shows that a proper balance between test accuracy and DP guarantee for very deep neural networks with complex datasets is still an open problem. Our BC and ALC techniques help towards achieving a practical balance, but more complimentary tricks and methods are still needed.
%One solution may be to work with much larger data sets (large $N$) which yields more privacy amplification due to subsampling.
%This means that we need more privacy amplification by using subsampling, that is, we need to work with large data sets (large $N$). 
%However, 

\section{Conclusion}
\label{chapter:Conclusion}

%In this work, 
We have proposed a new adaptive layerwise clipping method as well as a new batch clipping method for DPSGD. Our experiments show that DPSGD with BC and new ALC  can achieve faster convergence and  higher accuracy compared to DPSGD with IC. We have provided rigorous DP proofs for ALC and BC.

Our experiments are for small $\sigma$ which leads to weak differential privacy. We still need additional techniques beyond (optimizing) ALC and BC for training a deep neural network like resnet-18 with CIFAR10  in order to achieve a practical balance between test accuracy and DP guarantee. 

%We believe that our new methods significantly reduce  the gap between DPSGD and mini-batch SGD for deep neural networks. 


% \clearpage
\bibliography{bibliography}
\bibliographystyle{plainnat}

\clearpage
 \newpage
 \appendix
 \onecolumn

\section{Supplementary Material} \label{appendix}
As shown in the main paper, the training of resnet-18 for DP-SGD with BC and ALC converges when the added Gaussian noise is small enough.
%in Batch Clipping mode and ALC. 
We suspect that this is related to the size of the networks and the complexity of the training dataset. In Section \ref{app:convnet}, we work with shallow networks and see if the training of shallow networks on CIFAR10 can converge for larger Gaussian noise. In Section \ref{app:MNIST} we train a simple network on the simpler dataset MNIST and investigate whether the Gaussian noise can even be larger.  (We remind the reader that the larger the added Gaussian noise, the better the DP guarantee.) In Section \ref{app:BCandBNL}, we give evidence that DPSGD with batch clipping preserves the merits of using batch normalization layers in convolutional neural networks. To complete the work, Section~\ref{subsec:ALCvsFGC} compares ALC with Full Gradient Clipping (FGC) showing that ALC outperforms FGC.

\subsection{Lightweight Network on a Complex Dataset: convnet with CIFAR10}
\label{app:convnet}

We conduct the same experiments as for resnet-18 on the CIFAR10 dataset with a lightweight network (convnet) which consists of 5 layers. The first 4 layers are the combination of a convolutional layer, a batch normalization layer and an average pooling layer followed by CONV-BN-POOLING order. The last layer is a softmax layer. The convent model architecture is defined in Table \ref{tab:convnet}.

\begin{table}[ht]
\centering
\scalebox{0.8}{
\begin{tabular}{|c|c|c|c|c|c|c|}
\hline
 Operation Layer & \#Filters  & Kernel size & Stride & Padding & Output size & Activation function \\ \hline 
 \parbox[c]{3cm}{\vspace{1mm} \centering $Conv2D$ \vspace{1mm}}& $32$ & $3 \times 3$ & $1 \times 1$ & $1 \times 1$ & \multirow[c]{3}{*}{$16 \times 16 \times 3$} & ReLu\\ 
 $BatchnNorm2d$&   & $32 \times 32$ &  &  &  & \\ 
 $AvgPool2d$&   & $2 \times 2$ & $2 \times 2$ &  &  & \\ \hline 
\parbox[c]{3cm}{\vspace{1mm} \centering $Conv2D$\vspace{1mm}}& $64$ & $3 \times 3$ & $1 \times 1$ & $1 \times 1$ & \multirow[c]{3}{*}{$8 \times 8 \times 32$} & ReLu\\ 
 $BatchnNorm2d$&   & $64 \times 64$ &  &  &  & \\ 
 $AvgPool2d$&   & $2 \times 2$ & $2 \times 2$ &  &  & \\ \hline 
 \parbox[c]{3cm}{\vspace{1mm} \centering $Conv2D$\vspace{1mm} }& $64$ & $3 \times 3$ & $1 \times 1$ & $1 \times 1$ & \multirow[c]{3}{*}{$4 \times 4 \times 64$} & ReLu\\  
  $BatchnNorm2d$&   & $64 \times 64$ &  &  &  & \\ 
 $AvgPool2d$&  & $2 \times 2$ & $2 \times 2$ &  &  & \\ \hline 
 \parbox[c]{3cm}{\vspace{1mm} \centering $Conv2D$ \vspace{1mm}}& $128$ & $3 \times 3$ & $1 \times 1$ & $1 \times 1$ & \multirow[c]{3}{*}{$1 \times 1 \times 128$} & ReLu\\ 
  $BatchnNorm2d$&   & $128 \times 128$ &  &  &  & \\ 
 $AdaptiveAvgPool2d$&   & $1 \times 1$ & $1 \times 1$ &    &  & \\ \hline 
 \parbox[c]{3cm}{\vspace{1mm} \centering $FC2$\vspace{1mm} }& $-$ & $-$ & $-$ & $-$ & $10$ & $softmax$\\ \hline 
\end{tabular}
}
\caption{convnet model architecture with batch normalization layers}
\label{tab:convnet}
\end{table}

\begin{figure}[ht]
    \centering
    \includegraphics[width=0.6\textwidth]{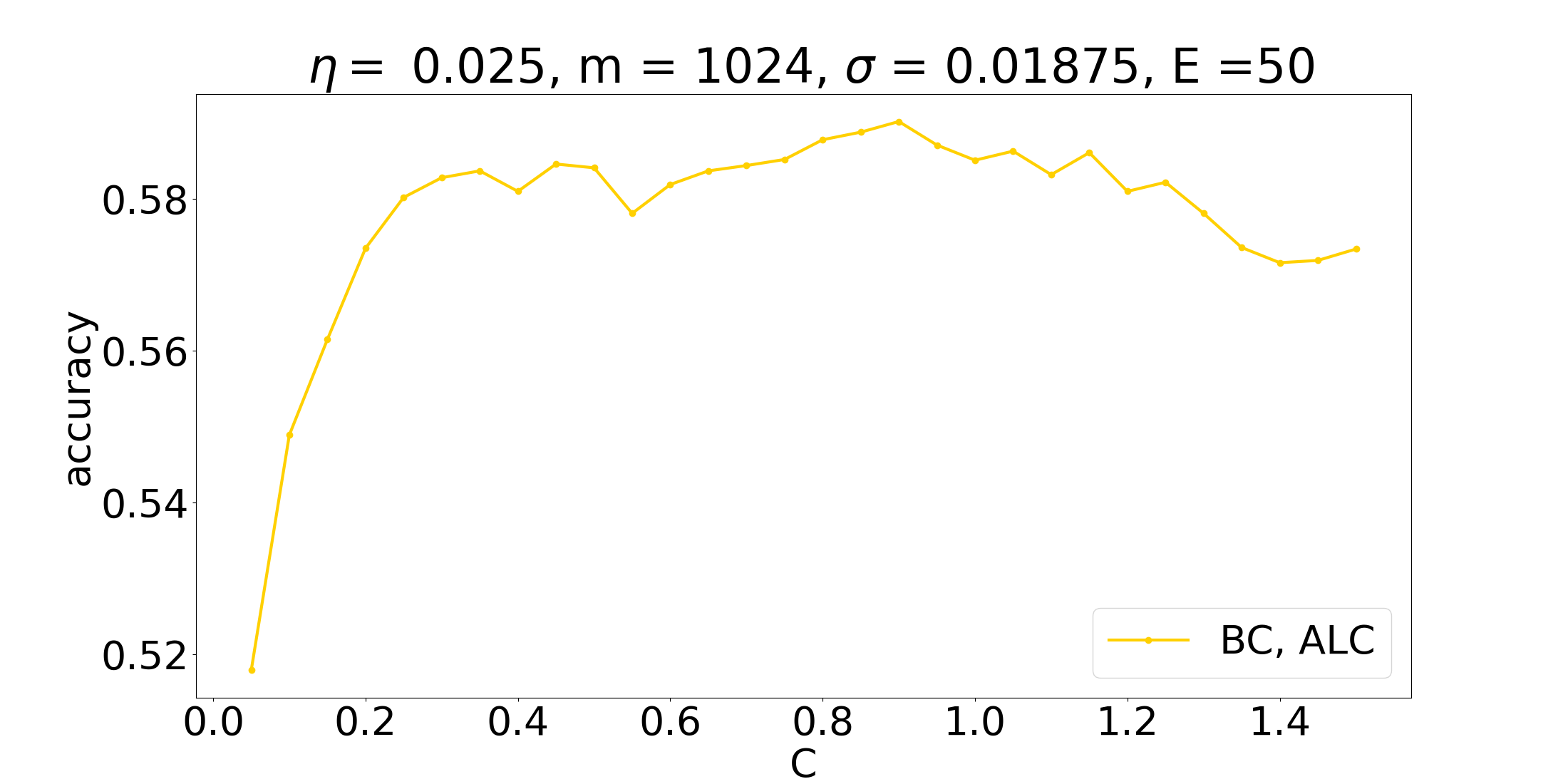}
    \caption{Test accuracy of DP-SGD with BC and ALC for convnet with  CIFAR10 with respect to various  master clipping values $C$ and fixed $\sigma = 0.01875$.}
    \label{fig:convnet_small_sigma_vary_C}
\end{figure}
In our first experiment we choose the noise multiplier $\sigma = 0.01875$, batch size $m=1024$, diminishing learning rate $\eta = 0.025$ with decay value $\eta_{decay} = 0.9$. We vary the master clipping value $C$ as shown in Figure \ref{fig:convnet_small_sigma_vary_C}. For $C = 0.9$, we achieve $59.02\%$ test accuracy after 50 epochs, which is  less than the $67\%$  test accuracy achieved by the resnet-18 model. We observe that the test accuracy  increases significantly for $0< C \leq 0.2$, is stable with some fluctuations for $0.2 < C < 0.9$, and decreases slightly for $C> 0.9$. We conclude that, for noise multiplier $\sigma=0.01875$, there is a range of $C$ where we see stable performance in terms of test accuracy.

%LET'S NOT ADD THIS OPEN PROBLEM ...
%\footnote{Rather than the experiment's trial and error process in order to find such a range of $C$, we leave it as an open problem to find a more efficient method.}

%we do not figure out how to find such a range of $C$ effectively rather than doing the trial and error process. We consider solving this problem as future works.

\begin{figure}[ht]
    \centering
    \includegraphics[width=0.75\textwidth]{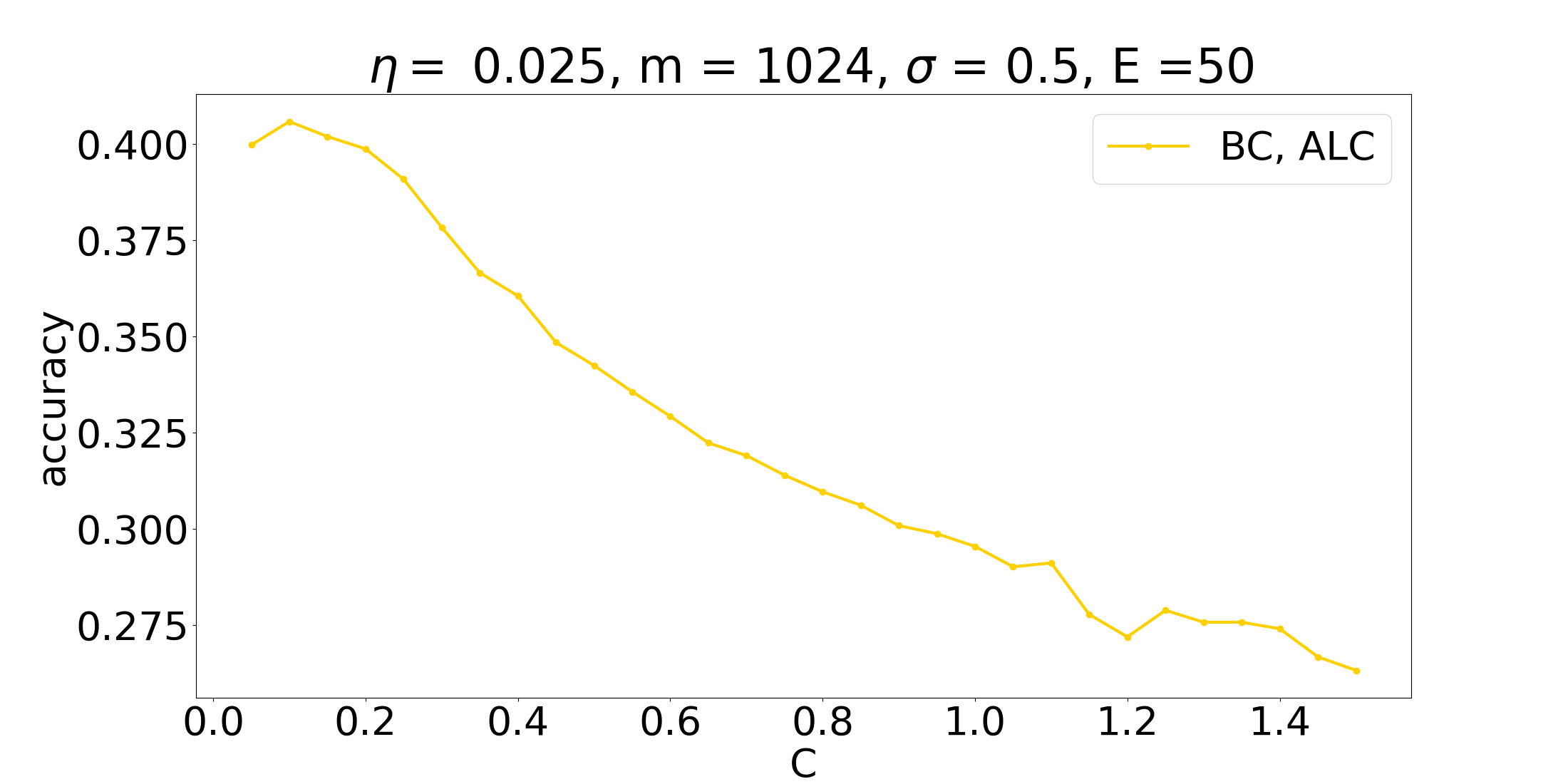}
    \caption{Test accuracy of DP-SGD with BC and ALC for convnet with  CIFAR10 with respect to various  master clipping values $C$, fixed $\sigma = 0.5$ and $m=1024$
    %CIFAR10 testing accuracy with convnet model with different master clipping value $C_{master}$ for $\sigma = 0.5$
    }
    \label{fig:convnet_big_sigma_vary_C}
\end{figure}

%\textcolor{red}{Figures 6 and 7 and 9 say "ours." Use the wording of the main body. "BC, ALC" instead of "ours, sigma=0..."}

In our second experiment we want to push the lightweight convnet model to the limit where we choose a relatively large $\sigma = 0.5$ with all other hyper-parameters remaining the same. As shown in figure \ref{fig:convnet_big_sigma_vary_C}, we only achieve $40.58\%$ test accuracy for $C = 0.1$ and we observe that the test accuracy decreases for $C> 0.1$.

\begin{figure}[ht]
    \centering
    \includegraphics[width=0.75\textwidth]{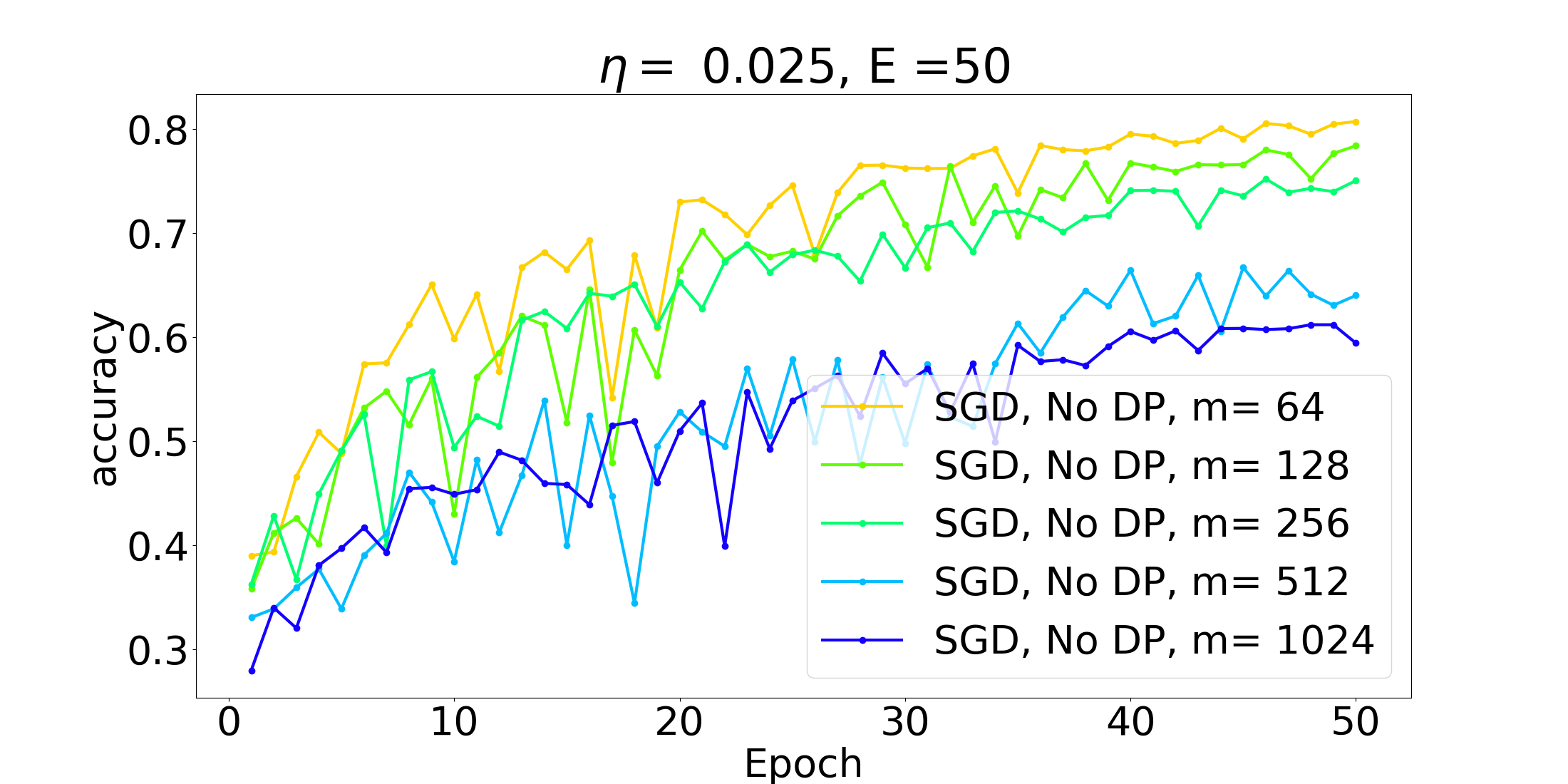}
    \caption{
     Test accuracies of mini-batch SGD without DP for different mini-batch sizes $m=(64,128,256,512,1024)$
    % Test accuracy benchmark of convnet model for CIFAR10 dataset
    }
    \label{fig:convnet_benchmark}
\end{figure}

See Figure \ref{fig:convnet_benchmark}, our third experiment studies  mini-batch SGD without DP for convnet with CIFAR10 as our benchmark. We see  that smaller batch sizes yield better accuracy. Specifically, we achieve $80.55\%$ test accuracy for batch size $m= 64$ and $61.19\%$ for batch size $m = 1024$.

\begin{figure}[ht]
    \centering
    \subfigure[Varying $C$]{\includegraphics[width=0.49\textwidth]{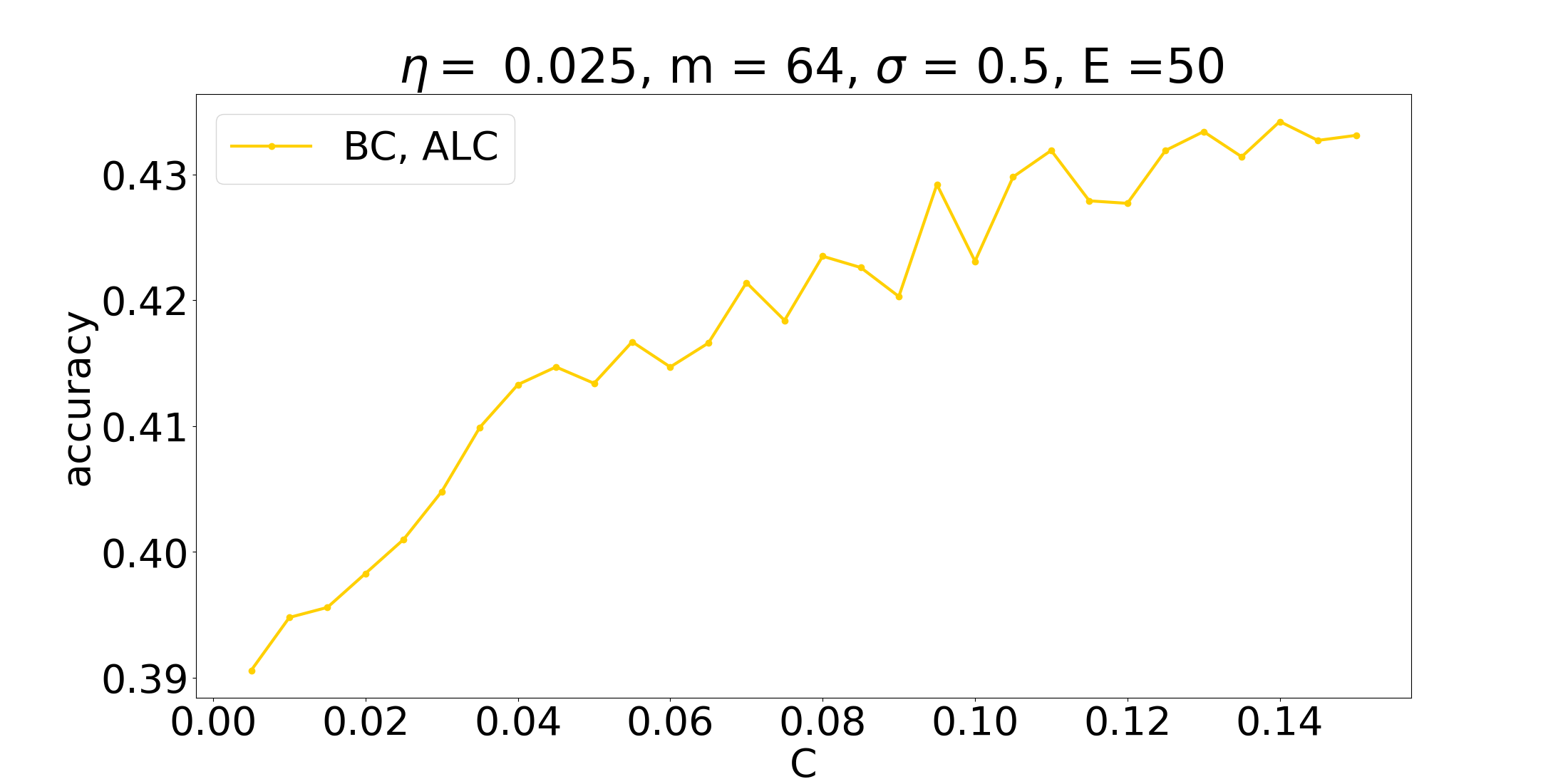}}
    \subfigure[Varying $\sigma$]{\includegraphics[width=0.49\textwidth]{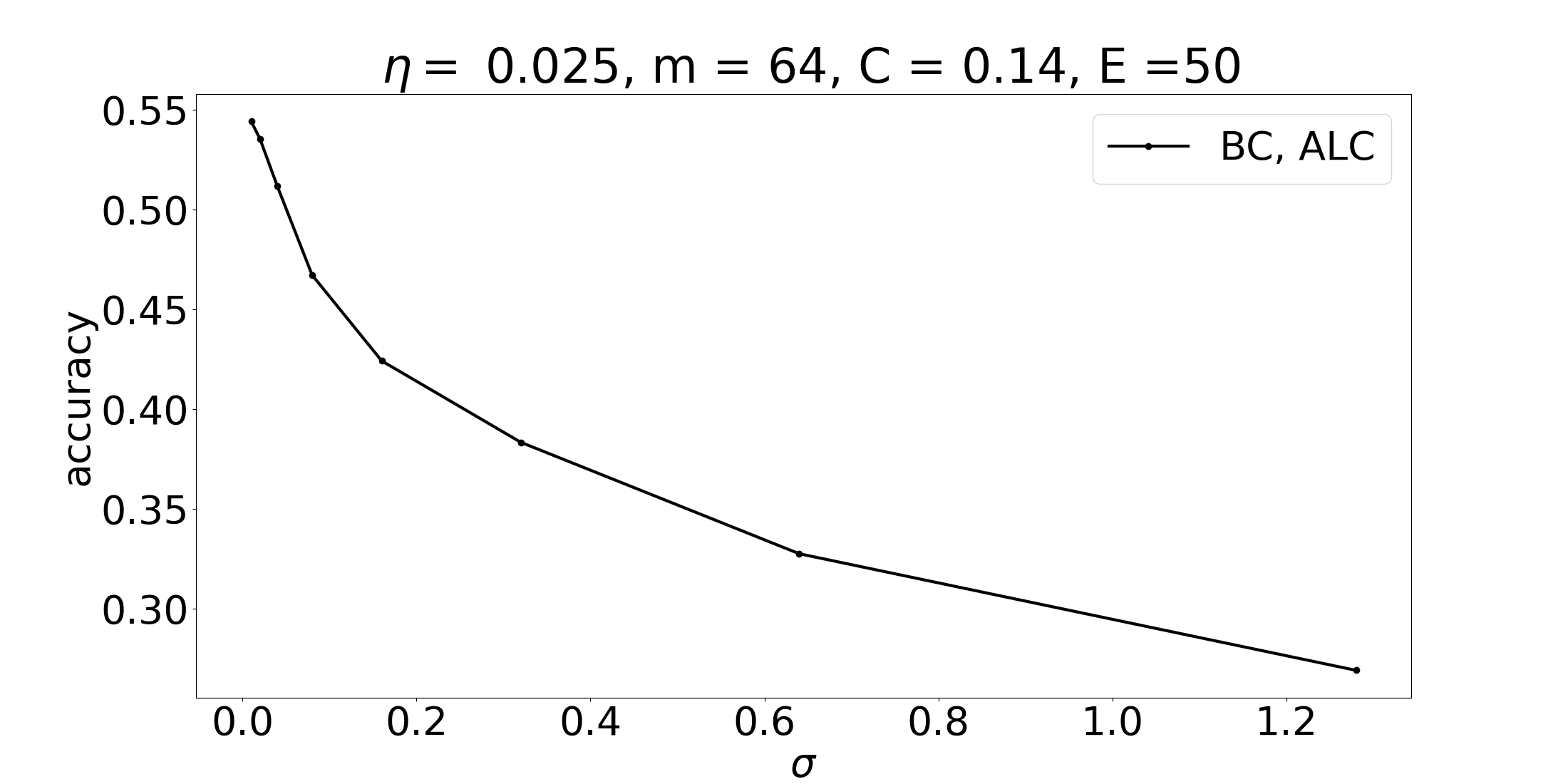}}
    \caption{Test accuracy of DP-SGD with BC and ALC for convnet with  CIFAR10 with respect to various  master clipping values $C$ and noise multiplier  $\sigma$ with batch size $m=64$
    % Varying $C$ for $\sigma = 0.5$ and $m=64$
    }
    \label{fig:convnet_vary_C_batchsize_64}
\end{figure}

This leads us to the fourth experiment, where, by choosing a smaller batch size,  we try to increase the test accuracy of DP-SGD with BC and ALC for convnet and CIFAR10 with $\sigma = 0.5$. For example, in Fig \ref{fig:convnet_vary_C_batchsize_64}.a we choose $m = 64$ and vary the master clipping constant $C$ to find the value which gives best test accuracy:  We achieve $43.42\%$ test accuracy for $C = 0.14$. Moreover, we also vary the noise multiplier value $\sigma$ with fixed master clipping value $C= 0.14$ to see whether our model can sustain larger noise. The Figure \ref{fig:convnet_vary_C_batchsize_64}.a shows that the testing accuracy drops $\approx 15\%$ when we increase $\sigma$ from $0.01$ to $0.2$ and decreases $\approx 10\%$ more from $\sigma=0.2$ to $\sigma = 0.6$. Therefore, we choose $\sigma = 0.5$ for the next experiments, where we can achieve $\approx 35\%$ testing accuracy and have better DP guarantee than our resnet-18 experiment.

\begin{figure}[ht]
    \centering
    \includegraphics[width=0.8\textwidth]{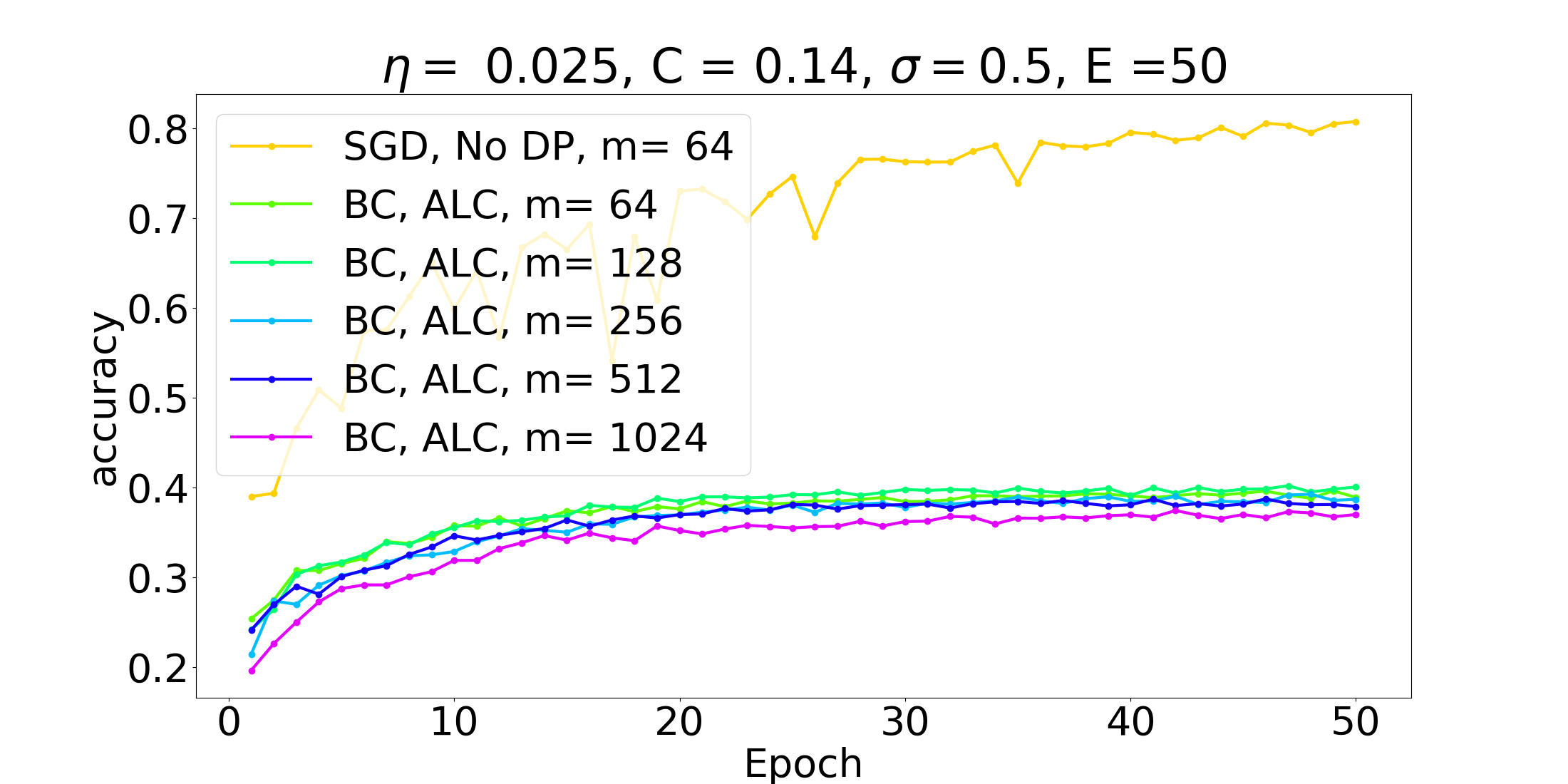}
    \caption{
     Test accuracies of DPSGD+BC+ALC vs mini-batch SGD without DP for different mini-batch sizes $m=(64,128,256,512,1024)$
    %Test accuracy for $C = 0.14$, $\sigma = 0.5$ and $m=128,256,512,1024$
    }
    \label{fig:convnet_BCversusbenchmark}
\end{figure}

Our final  experiment uses DP-SGD with BC and ALC to train convnet with CIFAR10 for different batch sizes $m=(64,128,256,512,1024)$; 
%\textcolor{red}{Why not include $m=64$ of the previous experiment which motivates $C=0.14$? Should we change the red 64 into 128 in the previous text? => There was a mistake on the settings file I create so we ran from batch size 128 onward => just retrain the model for batchsize 64 an reuploaded it} 
We use master clipping constant $C = 0.14$, initial learning rate $\eta = 0.025$ with decay rate $\eta_{decay}=0.9$, noise multiplier $\sigma = 0.5$, and number of epochs $E=50$. We compare  with mini-batch SGD without DP with $m=64$ in Figure \ref{fig:convnet_BCversusbenchmark}. Although the accuracy is hurt badly by the Gaussian noise,
%${\cal N}(0,(2C\sigma)^2{\bf I})$, 
the lightweight  convnet model is still able to converge to $\approx 40\%$ test accuracy while the test accuracy for resnet-18  starts to fall below $40\%$ for $\sigma \geq 0.2$ and only achieves $\approx 20\%$ test accuracy for $\sigma = 0.5$.

%Through this process, we are able to prove 
Our experiments show that the deep  resnet-18 network is more sensitive to the added Gaussian noise than the lightweight convnet network. This observation opens a new research direction where we want to simplify the neural network model as much as possible for a given dataset type (and corresponding learning task) while maintaining test accuracy and allowing a  large enough $\sigma$ for a reasonable DP guarantee. We expect (given our experiments) to be able to train  simpler network models with a larger noise multiplier $\sigma$ and this yields better privacy. The network simplification should not be too much in that the test accuracy of a trained model with DP noise should still be "good enough."
%while not hurting the model's performance too much.

%%%%%%%%%%%%%%%%%%%%%%%%%%%%%%%%%%%%%%%%%%%%%%%%%%%%%%
%{\color{red}Toan to Marten: I tried to run IC under these three settings: 
%\begin{itemize}
%    \item  $\sigma = 0.5$ + IC + diminishing stepsize + nor\_convnet (Batchnorm): diverged
%    \item  $\sigma = 0.5$ + IC + constant stepsize + convnet (no Batchnorm): converged to $20\%$
%    \item $\sigma = 0.5$ + IC + constant stepsize + nor\_convnet (Batchnorm):diverged
%\end{itemize}
%
%Here, we may not want to include IC because it appears that IC mode only converge when we remove the batch normalization layer. This may cause conflict to our comparison in the mainbody (where we run IC and BC with resnet-18 which has BatchNorm layer).
%
%Another note that we may not want to explain why shallow network works better with larger noise. Simply because we do not have the proof for that and we may just want to state it as an observation. Ha and I are thinking about a simple implementation to proof our intuition where shallow networks is less sensitive to DP noise. (Maybe for another paper I guess?) 
%}

\subsection{Lightweight Network on a Simple Dataset: BN-LeNet-5 with MNIST}
\label{app:MNIST}

We investigate how well our method performs on a simpler dataset compared to CIFAR-10. For this reason we conduct the same experiments of Section \ref{app:convnet} on the MNIST dataset.
%as we do on CIFAR-10 dataset.

MNIST consists of 60,000 training examples and 10,000 testing examples of handwritten digits \cite{Lenet5model}. Each example is a 28x28 gray-level image. For training,
%this dataset, 
we use the modified version of LeNet-5  \cite{Lenet5model}, where we add a batch normalization layer after each convolutional layer. The details of the modified LeNet-5 architecture (BN-LeNet-5) are described in Table \ref{tab:BN-LeNet-5}. For each training image, we crop a $32 \times 32$ region from it with padding of 4, apply a random horizontal flip to the image, and then normalize it with 
\[(mean, std) = (0.1307, 0.3081).\]

\begin{table}[ht]
\centering
\scalebox{0.8}{
\begin{tabular}{|c|c|c|c|c|c|c|}
\hline
 Operation Layer & \#Filters  & Kernel size & Stride & Padding & Output size & Activation function \\ \hline 
 \parbox[c]{3cm}{\vspace{1mm} \centering $Conv2D$ \vspace{1mm}}& $6$ & $5 \times 5$ & $1 \times 1$ & $0$ & $28 \times 28 \times 6$ & $tanh$\\ 
 $BatchnNorm2d$&   & $6 \times 6$ &  &  &  & \\ 
 \hline 
 $AvgPool2d$&   & $2 \times 2$ & $2 \times 2$ &  & $14 \times 14 \times 6$ & \\ \hline 
\parbox[c]{3cm}{\vspace{1mm} \centering $Conv2D$\vspace{1mm}}& $16$ & $5 \times 5$ & $1 \times 1$ & $0$ & $10 \times 10 \times 16$ & $tanh$\\ 
$BatchnNorm2d$&   & $16 \times 16$ &  &  &  & \\ \hline 
 $AvgPool2d$&   & $2 \times 2$ & $2 \times 2$ & &  $10 \times 10 \times 16$ &\\ \hline 
 \parbox[c]{3cm}{\vspace{1mm} \centering $Conv2D$\vspace{1mm} }& $120$ & $5 \times 5$ & $1 \times 1$ & $0$ &  $5 \times 5 \times 120$ & $tanh$\\ 
 $BatchnNorm2d$&   & $120 \times 120$ &  &  &  & \\ \hline 
 \parbox[c]{3cm}{\vspace{1mm} \centering $FC1$\vspace{1mm} }& $-$ & $-$ & $-$ & $-$ & $84$ & $tanh$\\ \hline 
 \parbox[c]{3cm}{\vspace{1mm} \centering $FC2$\vspace{1mm} }& $-$ & $-$ & $-$ & $-$ & $10$ & $softmax$\\ \hline 
\end{tabular}
}
\caption{BN-LeNet-5 model architecture}
\label{tab:BN-LeNet-5}
\end{table}

\begin{figure}[ht]
    \centering
    \includegraphics[width=0.8\textwidth]{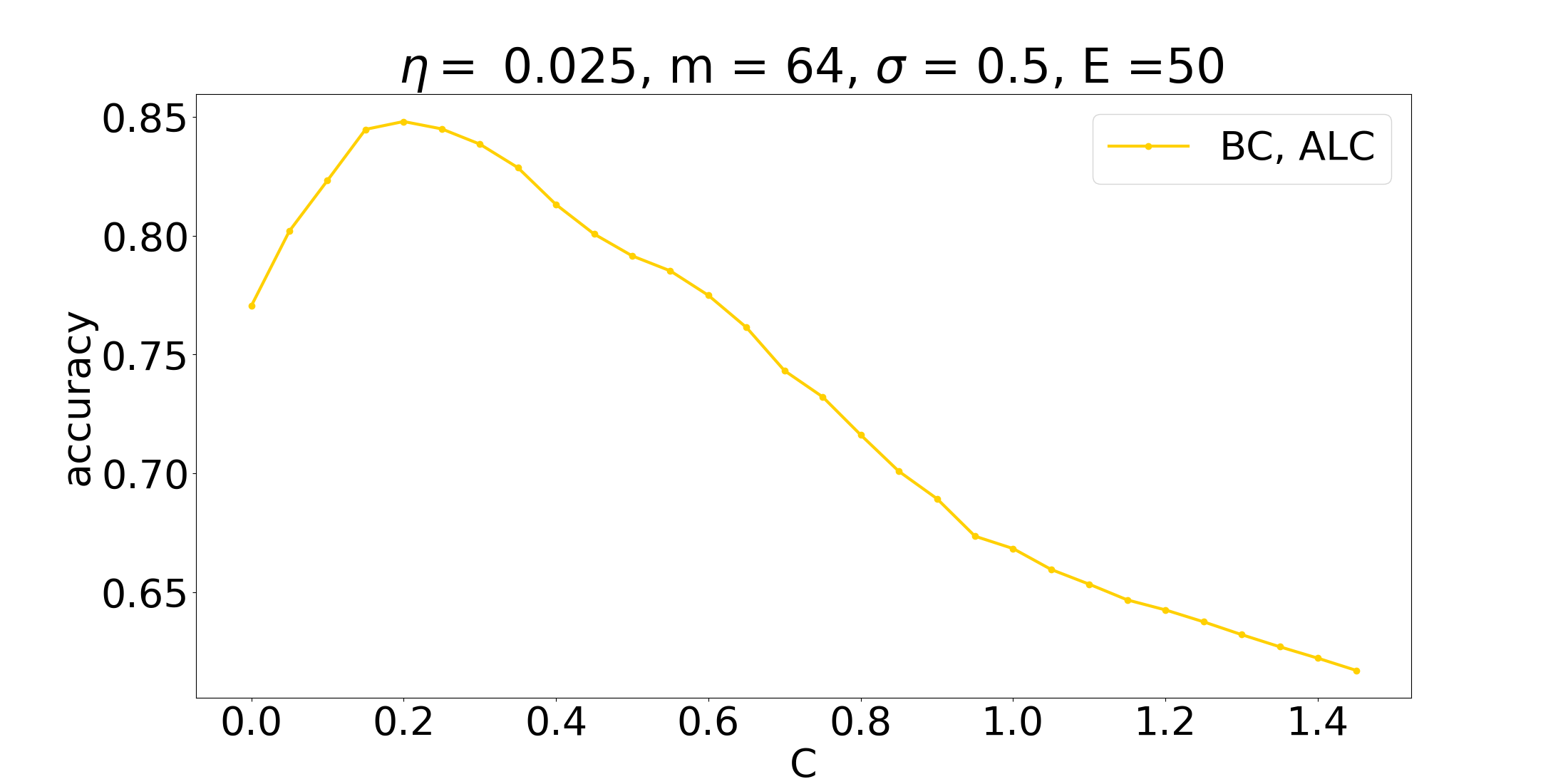}
    \caption{
     Test accuracy of DP-SGD with BC and ALC for BN-LeNet-5 with  MNIST with respect to various  master clipping values $C$, fixed $\sigma = 0.5$ and $m=64$
    %Test accuracy for $C = 0.14$, $\sigma = 0.5$ and $m=128,256,512,1024$
    }
    \label{fig:LeNet5varyC}
\end{figure}

As before, we fix the noise multiplier $\sigma = 0.5$ and   search for a good master clipping constant $C$. We use DP-SGD with BC and ALC to train the BN-LeNet-5 model with batch size $m=64$, diminishing step size $\eta = 0.025$ with decaying value $\eta_{decay} = 0.9$ in 50 epochs. See Figure \ref{fig:LeNet5varyC}, we achieve the best test accuracy $84.80\%$ for $C = 0.2$.  
%and this $C_{master}$ value gives the best accuracy because the testing accuracy decreases for $C<0.2$ and $C > 0.2$.

%%% C = 0.2, max_acc = 84.80

\begin{figure}[ht]
    \centering
    \subfigure[Varying $C$]{\includegraphics[width=0.49\textwidth]{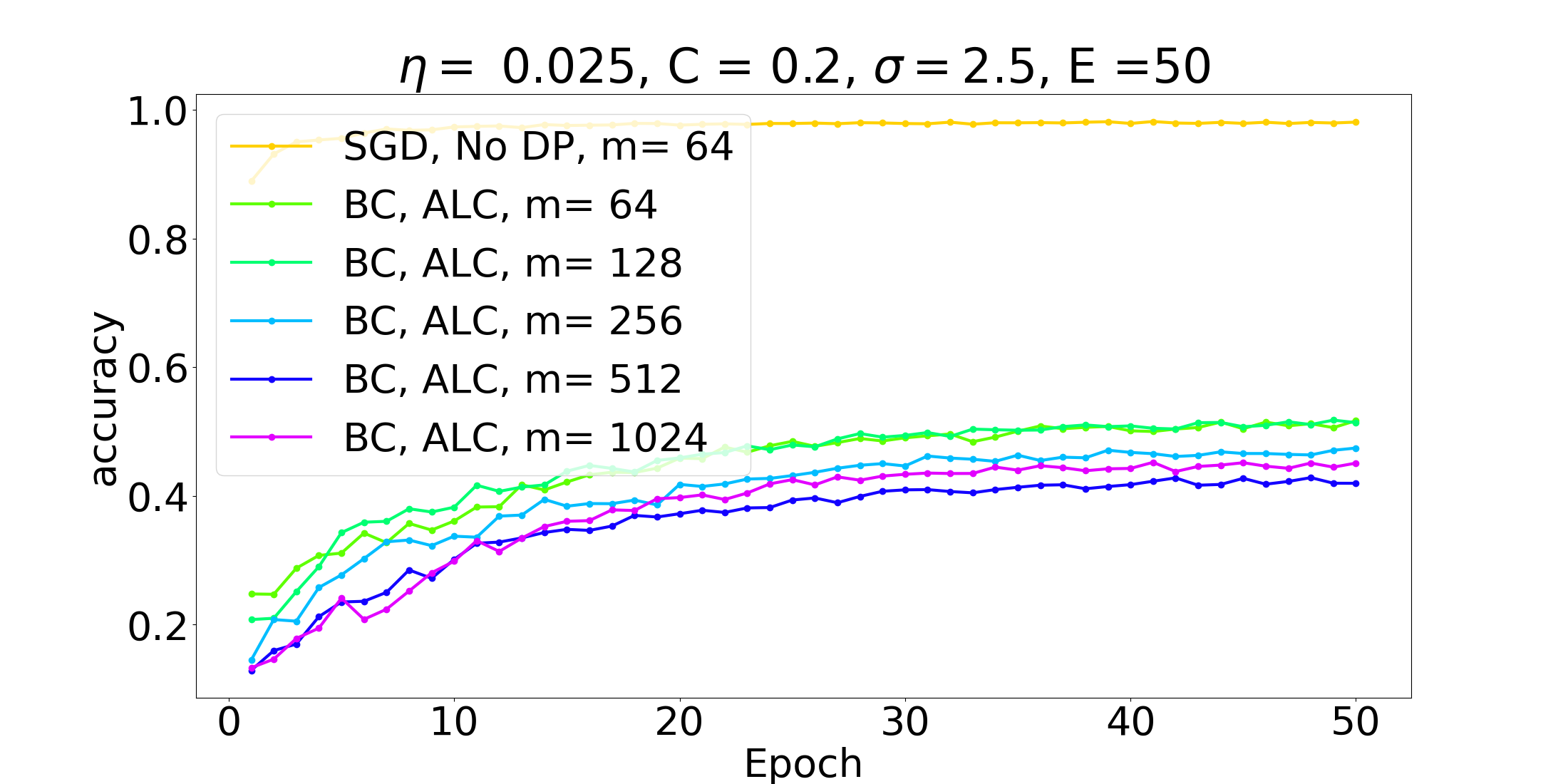}}
    \subfigure[Varying $\sigma$]{
    \includegraphics[width=0.49\textwidth]{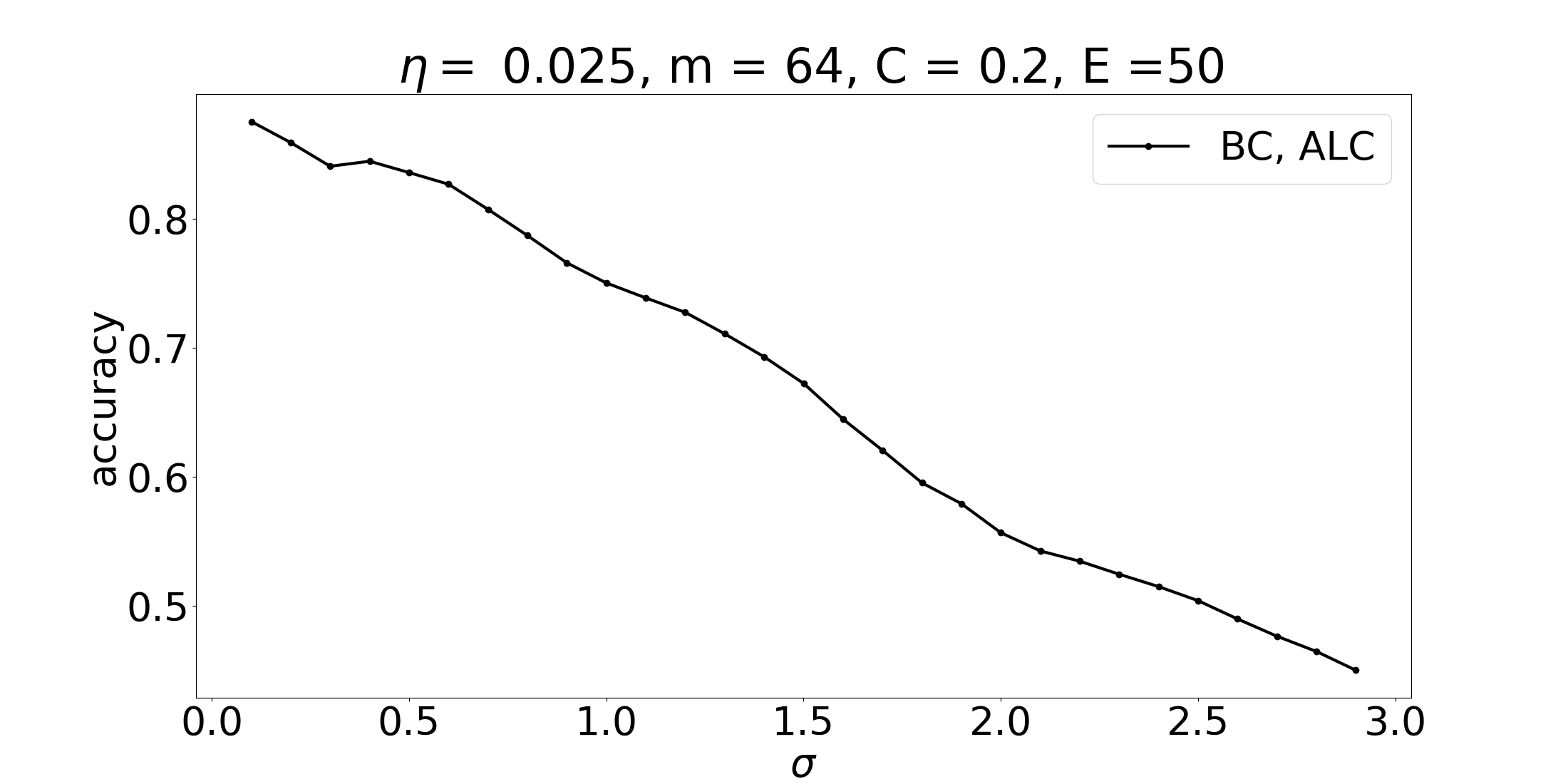}}
    \caption{
     Test accuracy of DP-SGD with BC and ALC for BN-LeNet-5 with  MNIST with respect to various  noise multiplier values $\sigma$ and master clipping values $C = 0.2$ with batch size $m=64$
    %Test accuracy for $C = 0.14$, $\sigma = 0.5$ and $m=128,256,512,1024$
    }
    \label{fig:LeNet5varySigma}
\end{figure}

Given $C = 0.2$, we push the BN-LeNet-5 model to the limit by choosing a relatively large noise multiplier $\sigma$ for which  the test accuracy does not drop below $50\%$. This allows us to see the effect of having a simpler dataset by comparing to the experiments in Section \ref{app:convnet}.
%to see how far $\sigma$ can reach for the MNIST dataset and BN-LeNet-5 model. 
We achieve $50.38\%$ test accuracy for  $\sigma = 2.5$ as shown in Figure \ref{fig:LeNet5varySigma}.b.

Let $C = 0.2$ and $\sigma = 2.5$, we train the BN-Lenet-5 model with various batch size $m = (64,128,256,512,1024)$. As shown in Figure \ref{fig:LeNet5varySigma}.a, the test accuracy decreases from $50.38\%$ to $45.11\%$ when we increase the batch size from $m=64$ to $m=1024$. 

Our main conclusion is that 
%More importantly, 
the BN-LeNet-5 model still converges for the large noise multiplier $\sigma =2.5$ when training on the MNIST dataset.
% \begin{figure}[ht]
%     \centering
%     \includegraphics[width=0.8\textwidth]{Figs/appendix_fig/nor_Lenet_subsampling_SGD_BC_lr_0.025_C_0.2_2.png}
%     \caption{
%      Test accuracies of DPSGD+BC+ALC vs mini-batch SGD without DP for different mini-batch sizes $m=(64,128,256,512,1024)$
%     %Test accuracy for $C = 0.14$, $\sigma = 0.5$ and $m=128,256,512,1024$
%     }
%     \label{fig:LeNet5varybatchsize}
% \end{figure}
Therefore, the simpler dataset allows us to use more Gaussian noise for differential privacy and this 
%a higher Gaussian Differential Privacy noise multiplier
%which 
yields an improved Differential Privacy guarantee.

%\textcolor{red}{Toan to Marten: We have experiment result which shows adding batch normalization layer yields better accuracy, should we include this in the appendix?}

%\textcolor{red}{
%Not sure ... I think BNL have been introduced to improve accuracy and everyone knows this -- so, why would this not also be the case if clipping and noise is added? Of course you have more evidence because of your experiment(s). But it may deviate from the main message of the paper?
%}
%\textcolor{red}{
%Ha suggests adding it because it shows that batch clipping + BNL is meaningful
%}
%Okay, this also makes sense. We can add just a very short A.3 (and also add a line to the intro of appendix A)

%\textcolor{red}{
%Toan to Marten: Just talked to Ha, We will add two more figures which compare the performance of the model between with and without BNL in no DP scenario. We can infer whether we can maintain the effect of the batch normalization layer in the batch clipping environment. Currently, I am re run the benchmark experiment for convnet without BNL.
%}
\subsection{Batch clipping and Batch Normalization Layer}
\label{app:BCandBNL}
\begin{figure}[ht]
    \centering
    \subfigure[convnet with batch normalization layer]{\includegraphics[width=0.49\textwidth]{Figs/appendix_fig/nor_convnet_benchmark.png}}
    \subfigure[convnet without batch normalization layer]{
    \includegraphics[width=0.49\textwidth]{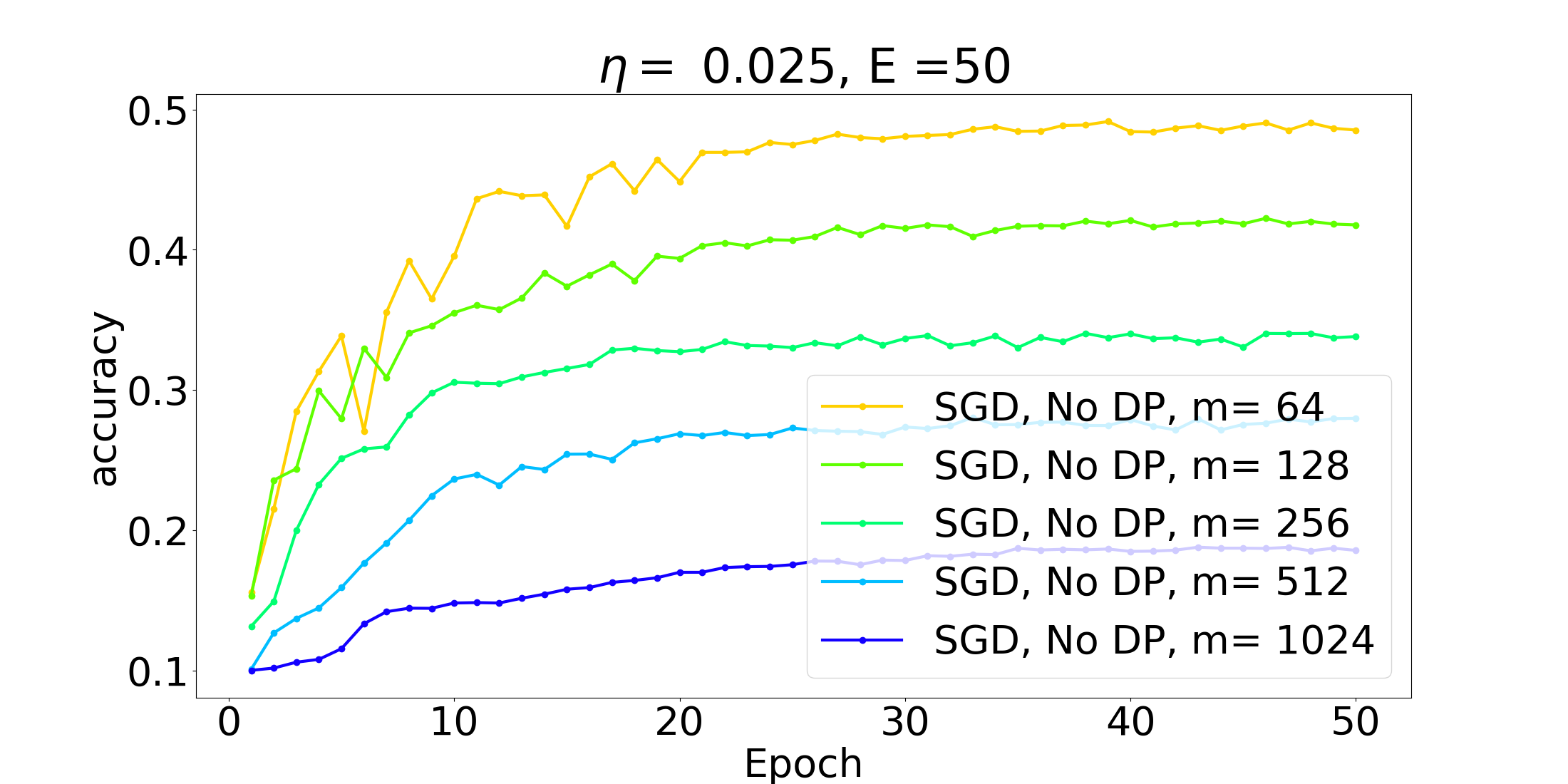}}
    \caption{Comparison of the test accuracy of training CIFAR10 with and without BNLs for convnet using SGD without DP.
    % of convnet  for CIFAR-10 between having batch normalization layers or not having batch normalization layers with normal training.
    %Test accuracy for $C = 0.14$, $\sigma = 0.5$ and $m=128,256,512,1024$
    }
    \label{fig:BNbenchmark}
\end{figure}

The concept of a Batch Normalization Layer (BNL) has been introduced in \cite{ioffe2015batch} to improve the training speed and testing accuracy for convolutional neural networks. Figure \ref{fig:BNbenchmark} shows that indeed for normal training with SGD without DP batch normalization layers allow a high test accuracy.
In this section we  investigate how using BNLs helps attaining a higher test accuracy when using DP-SGD 
%differentially private training case where we perform the batch clipping 
with BC and ALC.
%and add the noise to the gradient at each iteration. 

\begin{figure}[ht]
    \centering
    \subfigure[convnet with batch normalization layer]{\includegraphics[width=0.49\textwidth]{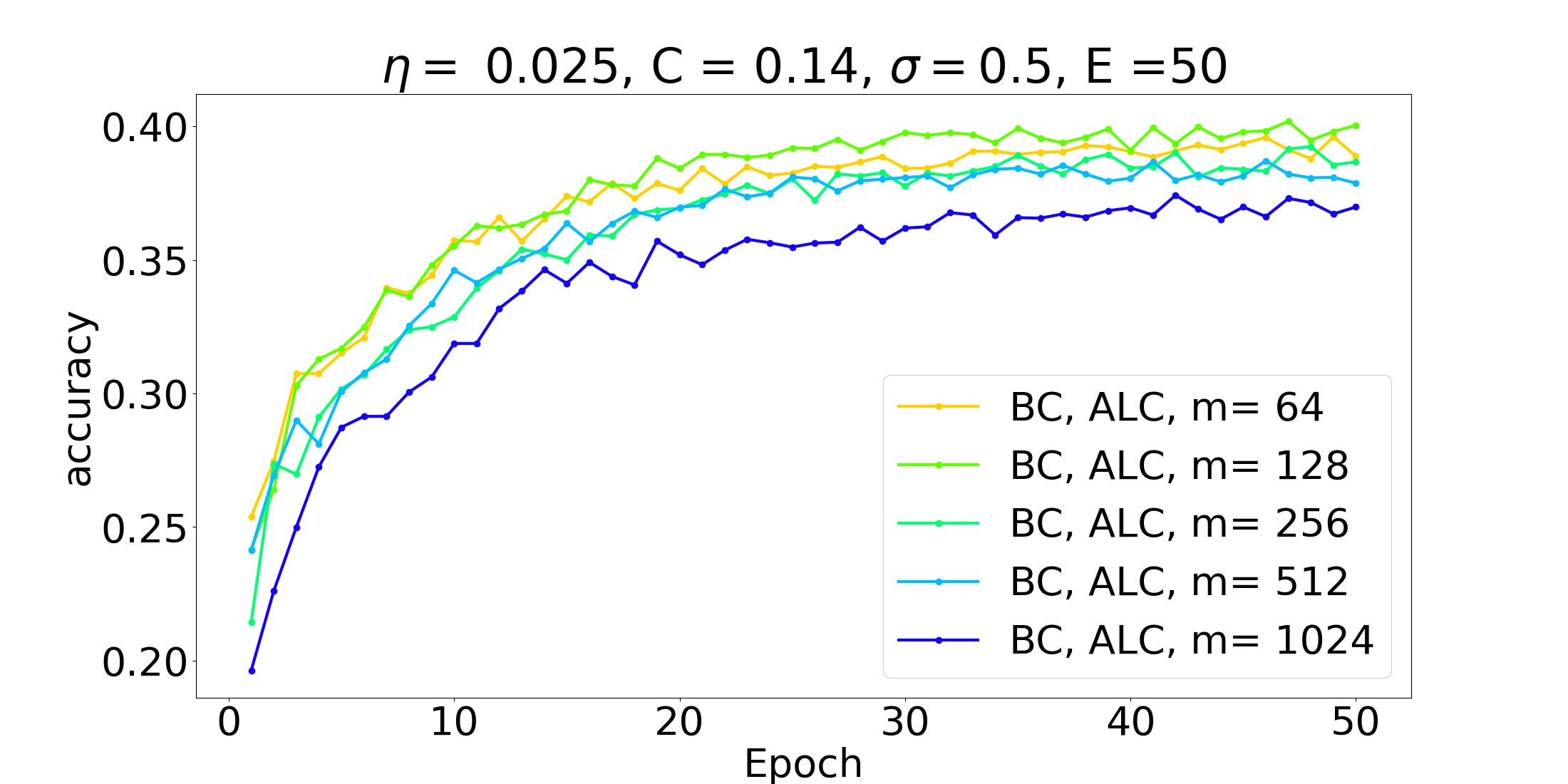}}
    \subfigure[convnet without batch normalization layer]{
    \includegraphics[width=0.49\textwidth]{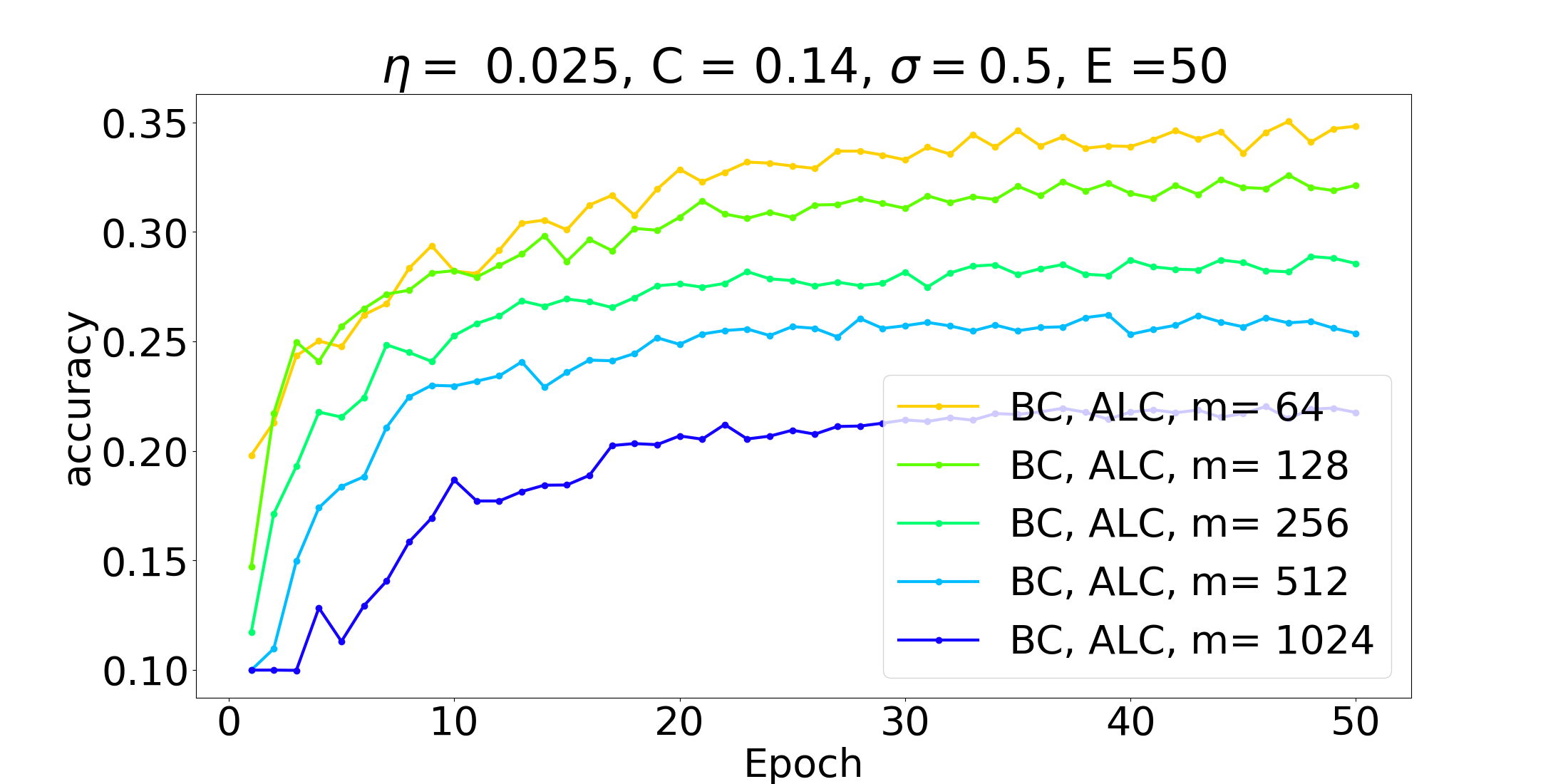}}
    \caption{Comparison of the test accuracy of training CIFAR10 with and without BNLs for convnet using DP-SGD with BC and ALC.
    % Comparison test accuracies of convnet  for CIFAR-10 between having batch normalization layers or not having batch normalization layers under batch clipping and ALC method.
    %Test accuracy for $C = 0.14$, $\sigma = 0.5$ and $m=128,256,512,1024$
    }
    \label{fig:BCandBN}
\end{figure}

We compare training  convnet \ref{tab:convnet}   with and without BNLs for CIFAR10 by using DP-SGD with BC and ALC. 
Figure \ref{fig:BCandBN} shows that we achieve $\approx 5\%$ higher test accuracy for DP-SGD with batch size $m=64$, 
diminishing step size $\eta = 0.025$ with decaying value $\eta_{decay}=0.9$, master clipping constant $C = 0.14$, noise multiplier $\sigma = 0.5$ and total number of epochs $E=50$.

We also run experiments with convnet and resnet18 after removing all BNLs for the set-up in Table \ref{tab:hyperparameter} (as in Section \ref{sec:experiments} in the main body) where we also consider a diminishing master clipping constant with initial value $C=0.095$ decaying with rate $C_{decay}=0.9$ after each epoch.
%our experiments with resnet18 and simple net after removing all the batch normalization layers in these models under both IC and BC modes. We also use the same experiments on the main body which are described in table  \ref{tab:hyperparameter}
%
\begin{table}[ht]
\centering
\scalebox{0.8}{
\begin{tabular}{|c|c|}
\hline
 Learning Rate $\eta$ & $0.025$ \\ \hline 
 Master Clipping value ($C$) & $0.095$ \\ \hline 
 Noise multiplier $\sigma$ & $0.01875$ \\ \hline 
 Learning rate decay ($\eta_{decay})$ & $0.9$ \\ \hline 
 Clipping value decay ($C_{decay}$) & $0.9$ \\ \hline 
 Batch size ($m$) & $[64,128,256,512,1028]$\\ \hline 
 Epochs ($E$) & $50$\\ \hline 
\end{tabular}
}
\caption{Hyperparameter settings}
\label{tab:hyperparameter}
\end{table}

\begin{figure}
    \centering
    \subfigure[css,constant $C$]{\includegraphics[width=0.49\textwidth]{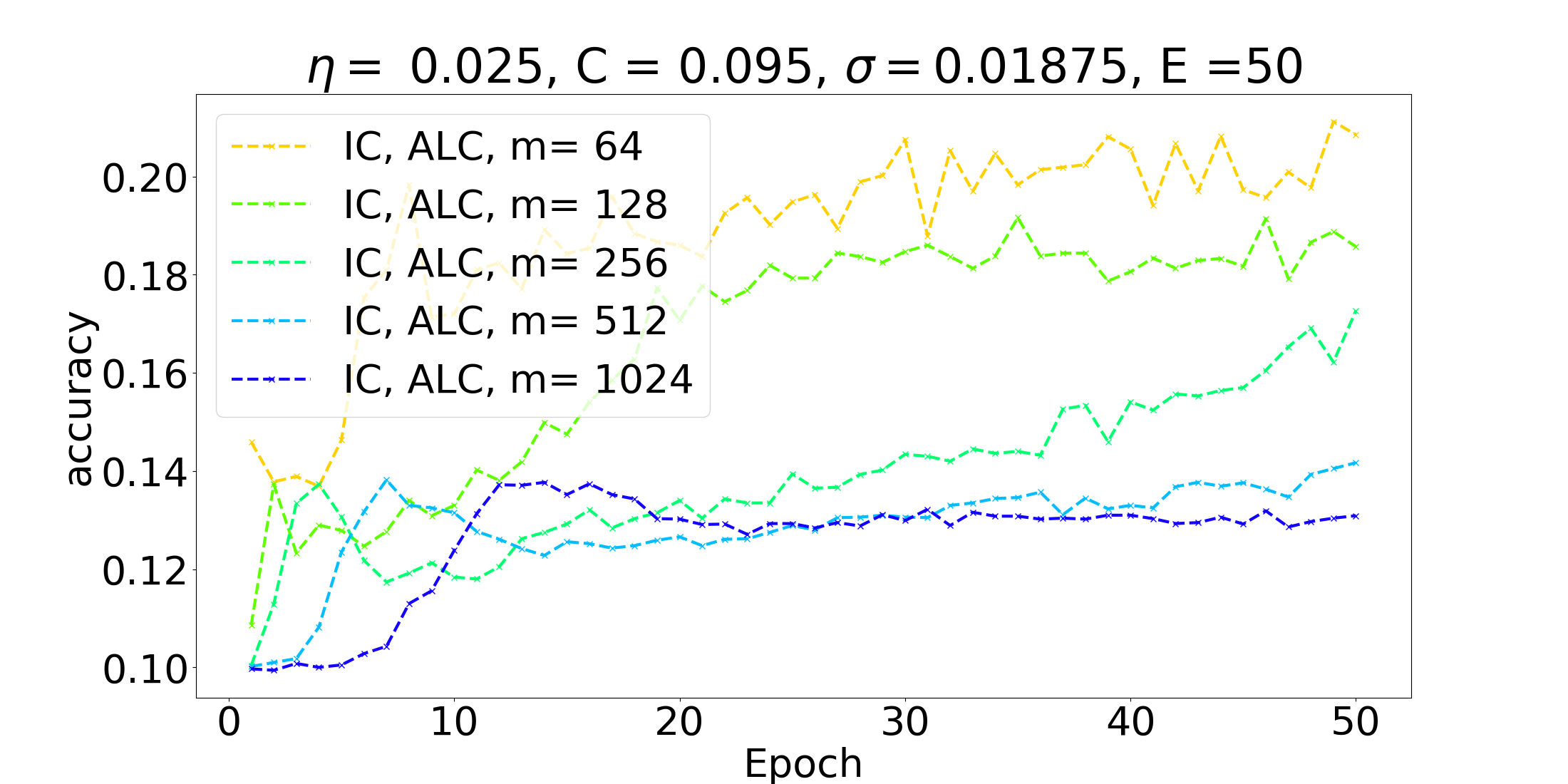}}
    \subfigure[css,diminishing $C$]{\includegraphics[width=0.49\textwidth]{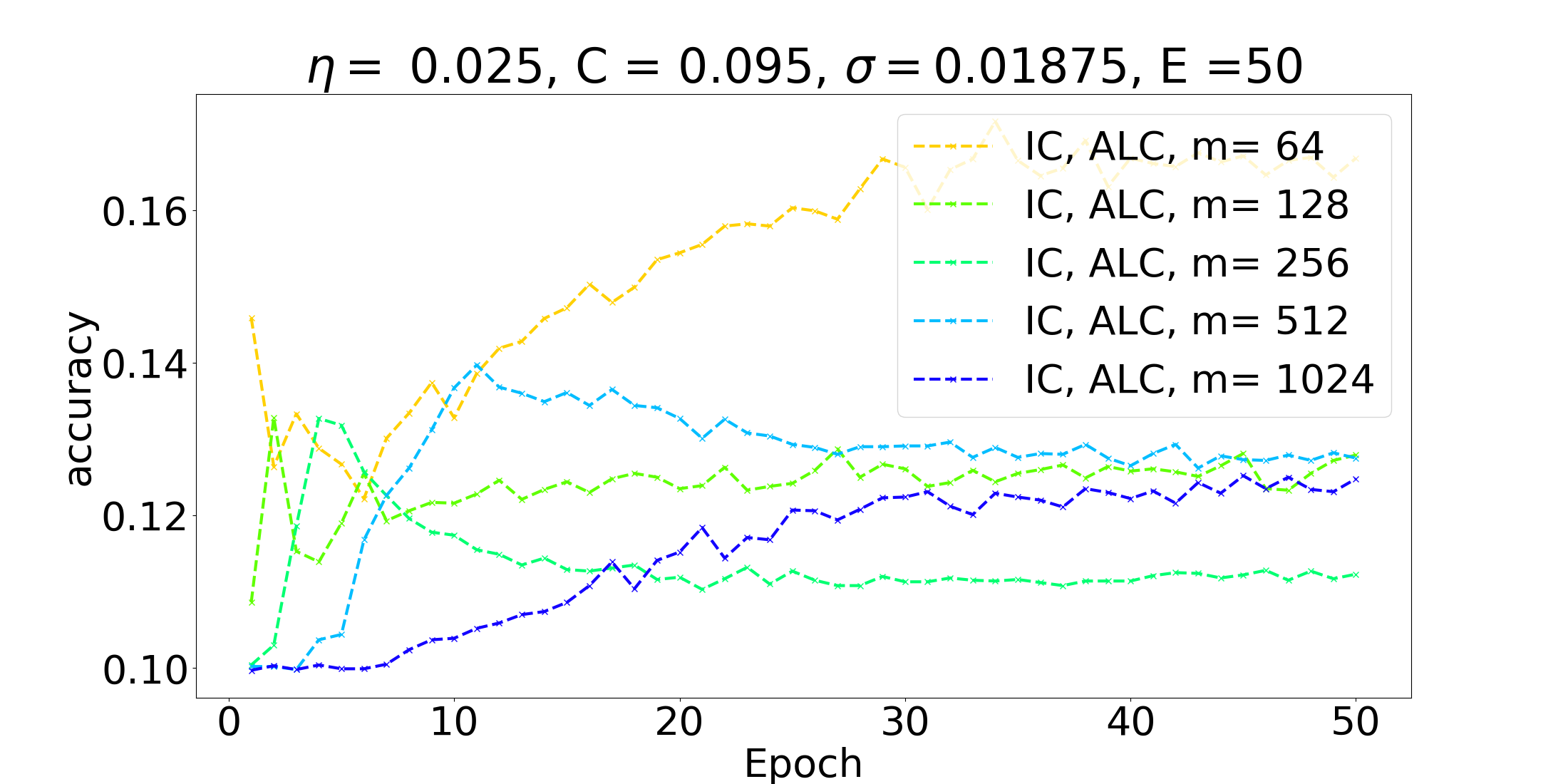}} \\
    \subfigure[dss,constant $C$]{\includegraphics[width=0.49\textwidth]{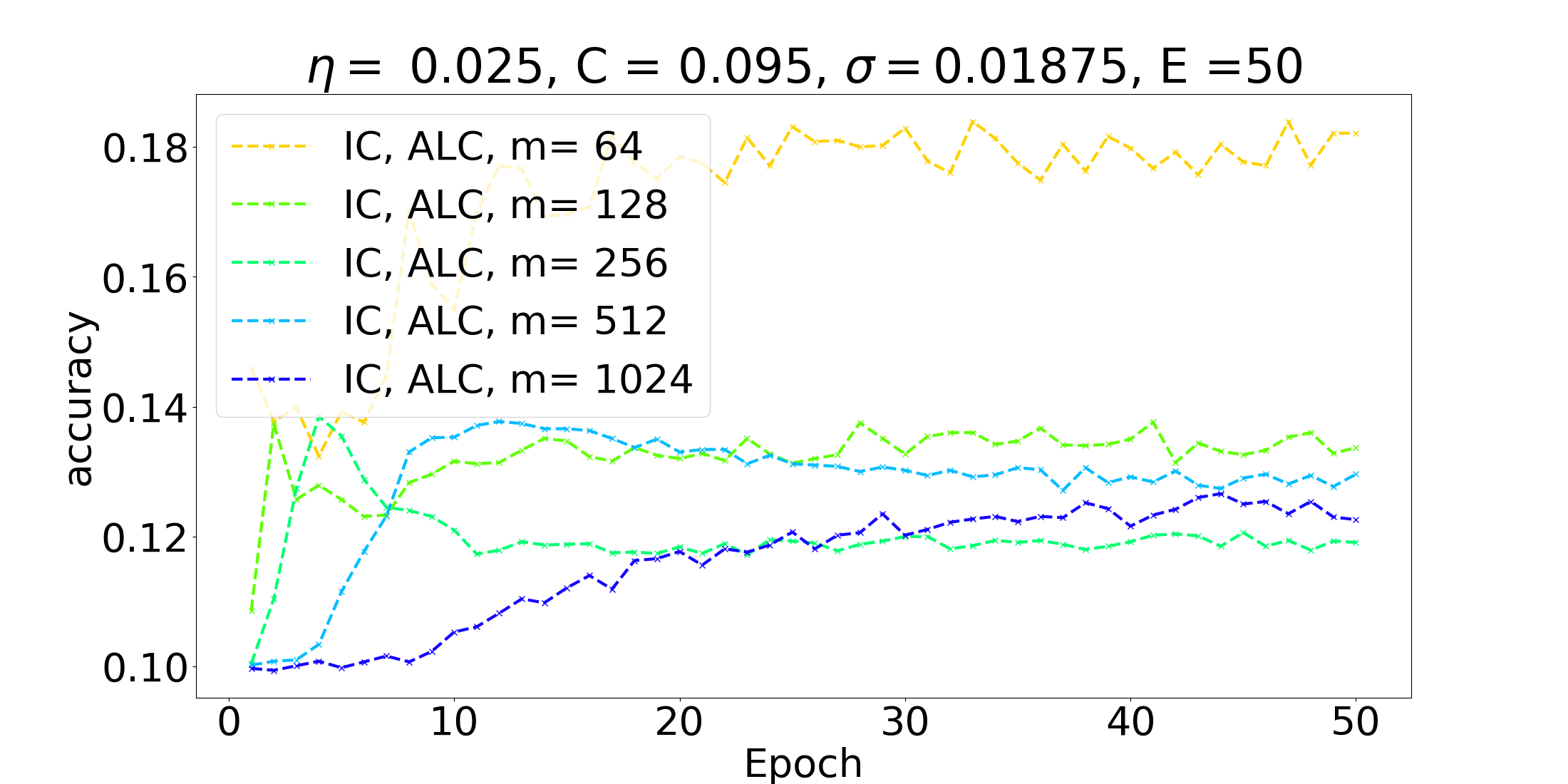}}
    \subfigure[dss,diminishing $C$]{\includegraphics[width=0.49\textwidth]{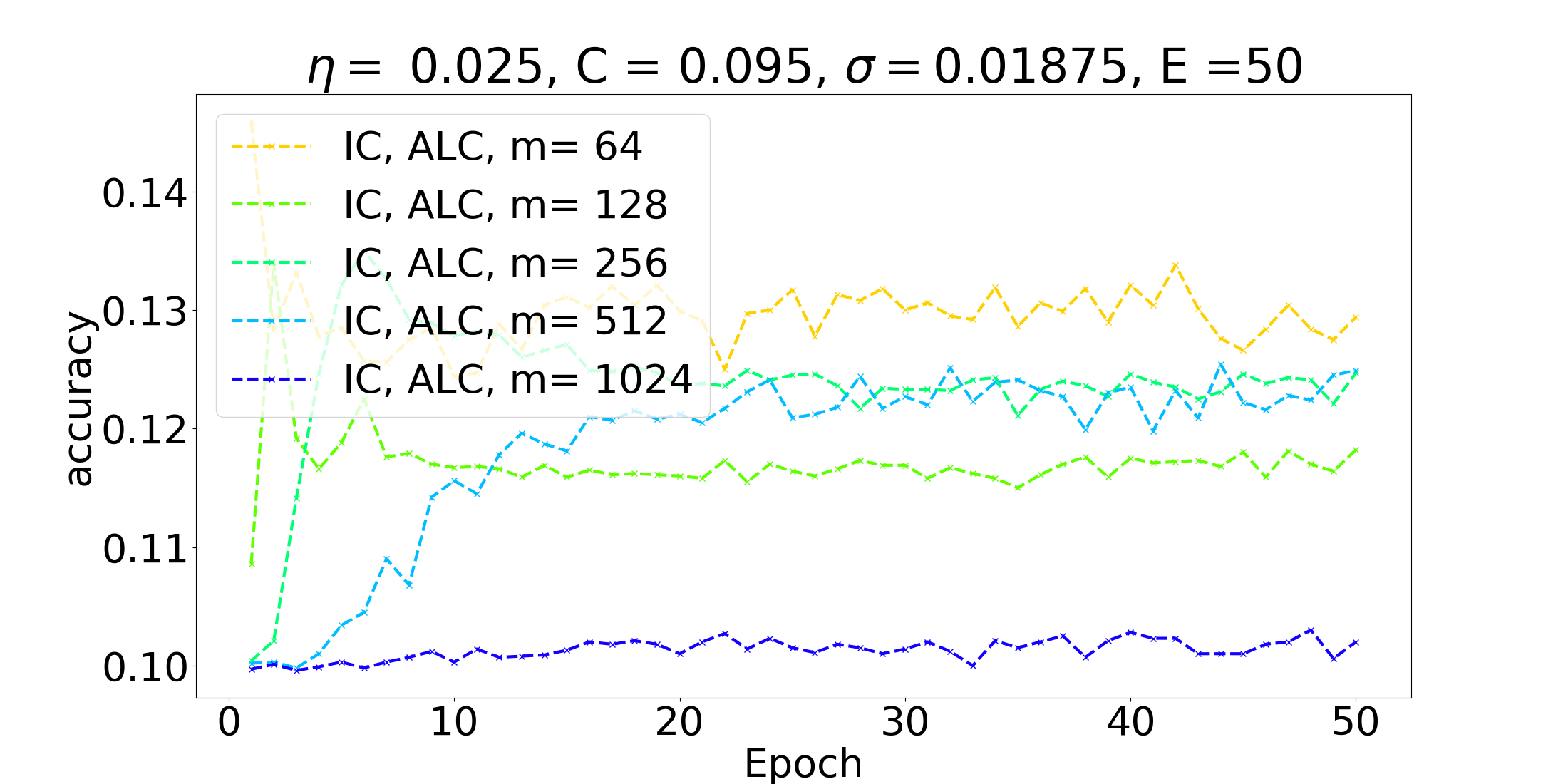}}
    \caption{
    %convnet without batch normalization layer under IC mode, 
    Test accuracy of training CIFAR10  for convnet without BNLs using DP-SGD with IC and ALC. Here, css and dss denote constant step size and diminishing step size, respectively.
    % Comparison test accuracies of convnet  for CIFAR-10 between having batch normalization layers or not having batch normalization layers under batch clipping and ALC method.
    %Test accuracy for $C = 0.14$, $\sigma = 0.5$ and $m=128,256,512,1024$  
    }
    \label{fig:convnetnonBNICresult}
\end{figure}

Figure \ref{fig:convnetnonBNICresult} is for convnet with IC+ALC (as compared to Figure \ref{fig:BCandBN} which is for BC+ALC).
After removing batch normalization layers in the convnet model, the testing accuracy for IC cannot converge to an acceptable value.
%as shown in Figure \ref{fig:convnetnonBNICresult} for individual clipping. 
After 50 epochs, we only achieve $21.12\%$ test accuracy if we train the model with constant step size and constant master clipping value $C$
%with our ALC method 
for mini-batch size $m=64$. This shows that batch clipping outperforms individual clipping  for  convnet without BNLs.

\begin{figure}
    \centering
    \subfigure[css,constant $C$]{\includegraphics[width=0.49\textwidth]{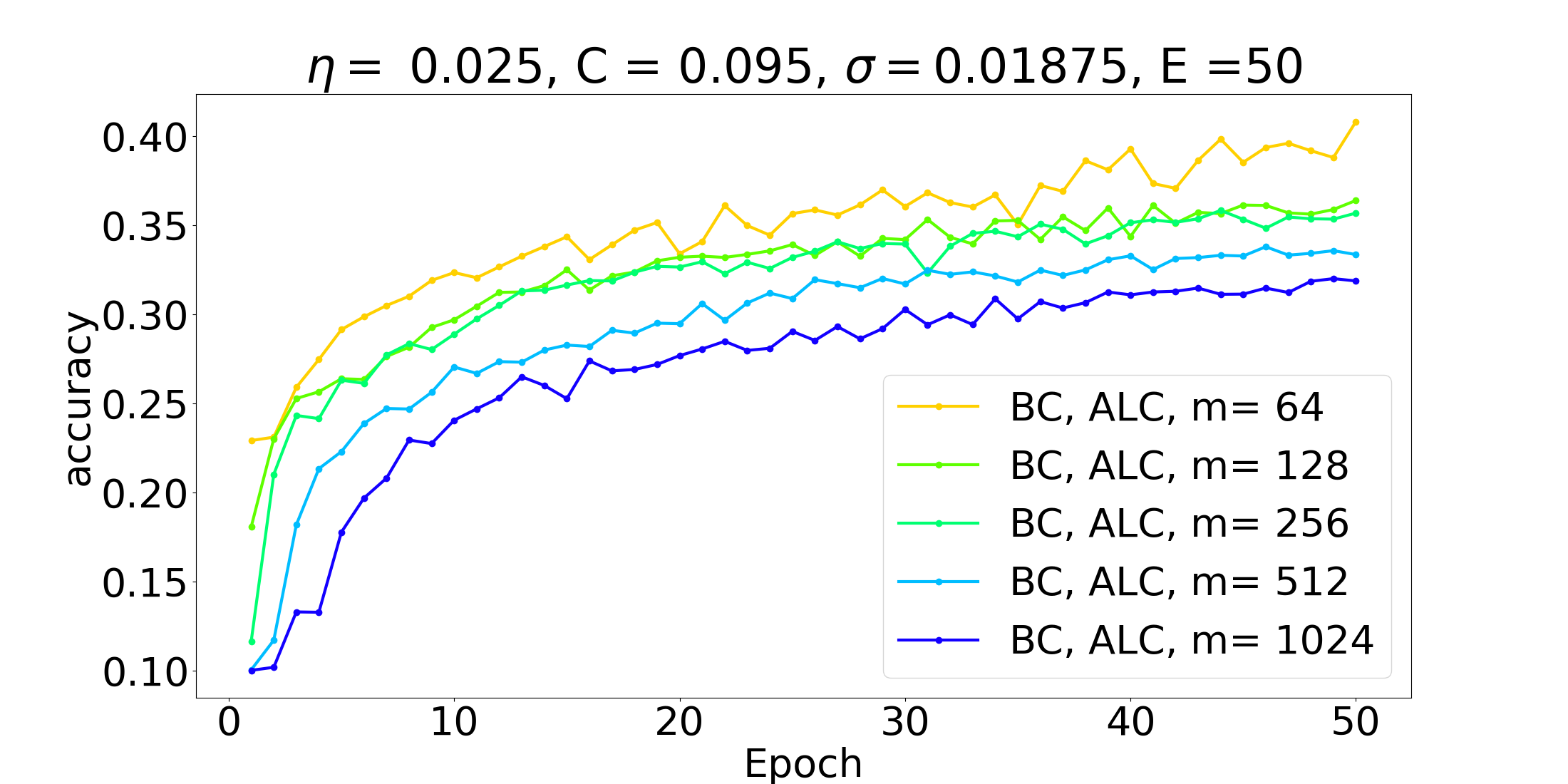}}
    \subfigure[css,diminishing $C$]{\includegraphics[width=0.49\textwidth]{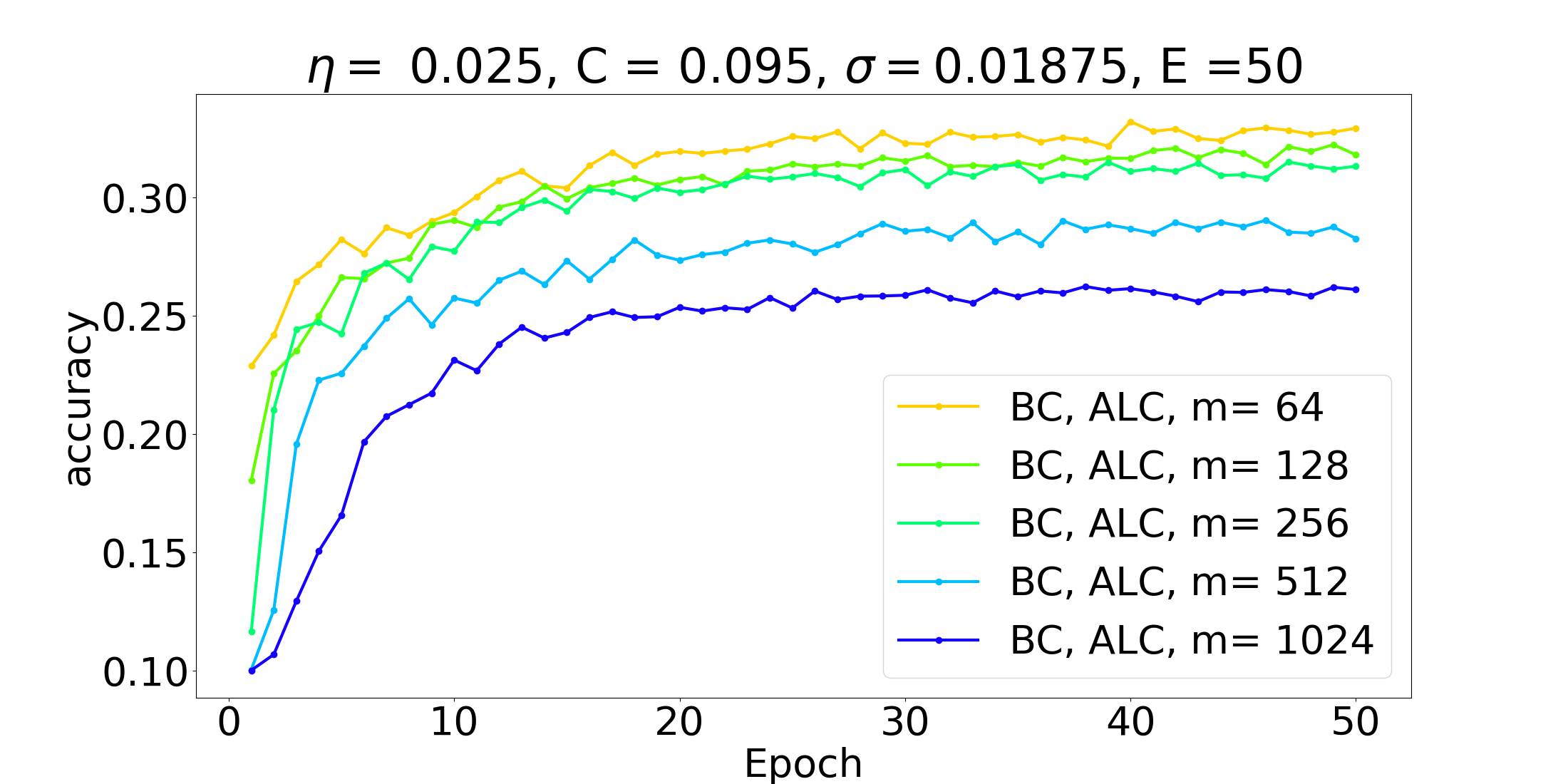}} \\
    \subfigure[dss,constant $C$]{\includegraphics[width=0.49\textwidth]{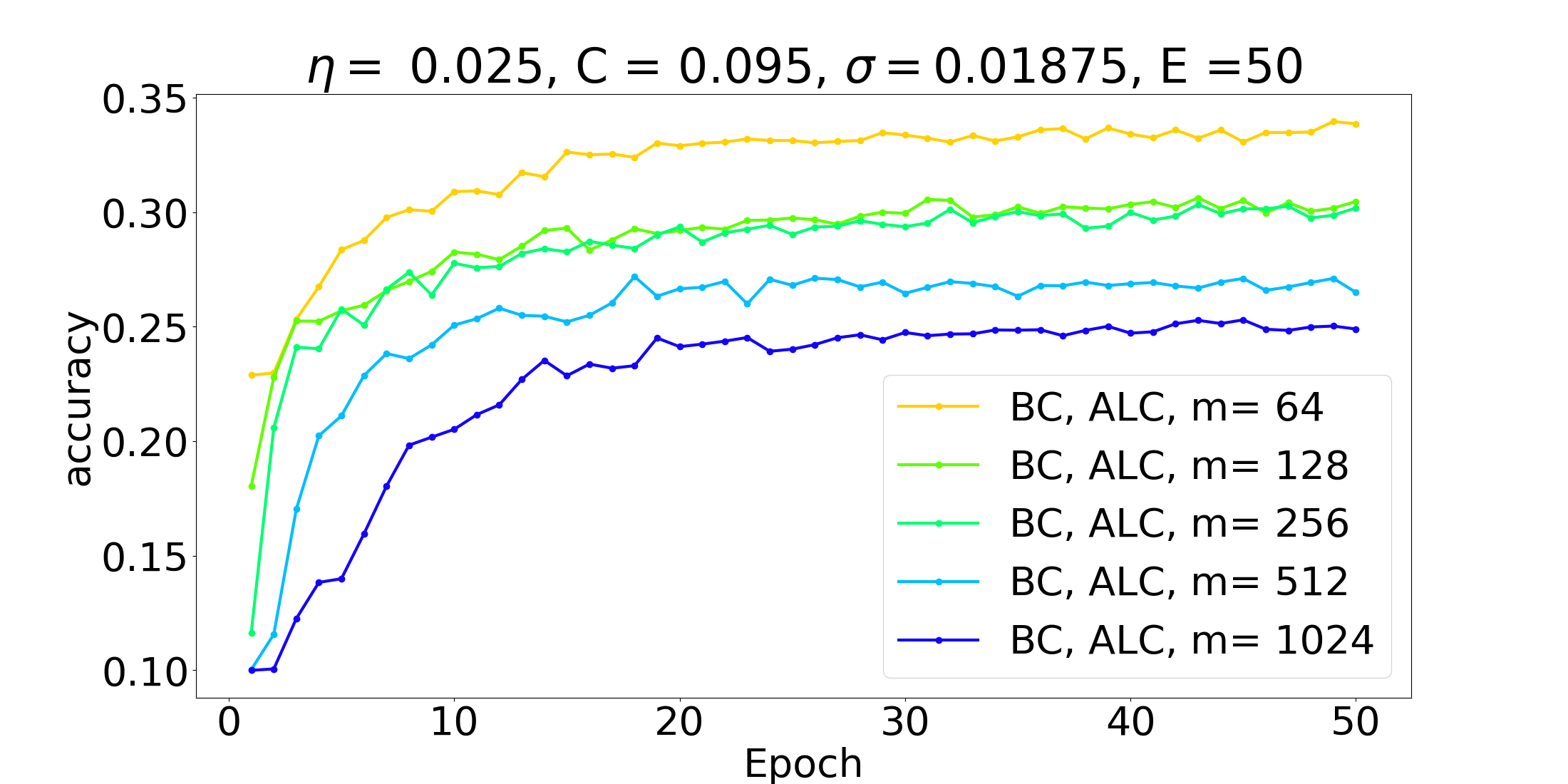}}
    \subfigure[dss,diminishing $C$]{\includegraphics[width=0.49\textwidth]{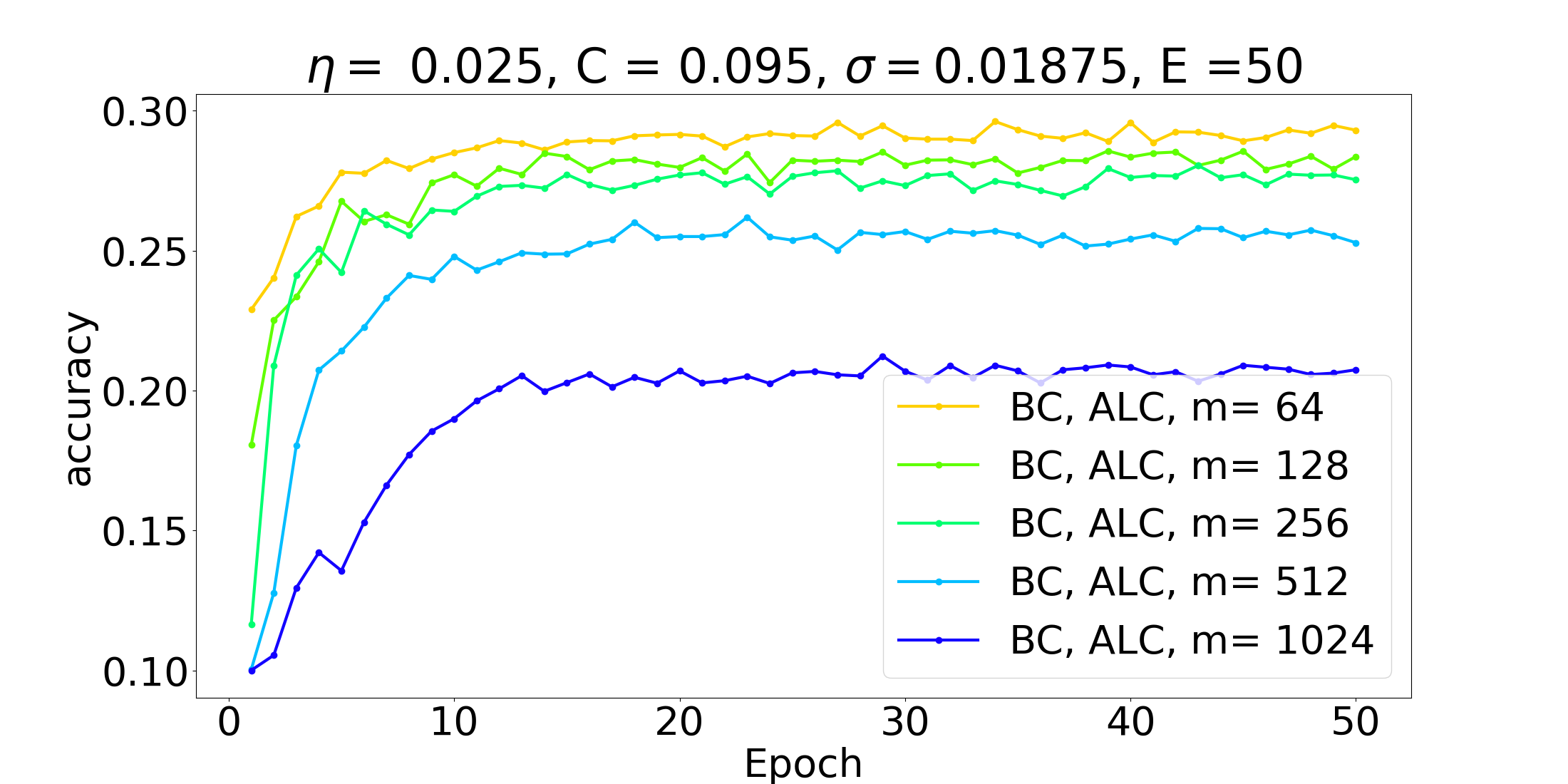}}
    \caption{
    %resnet18 without batch normalization layer under BC mode, css and dss denote constant step size and diminishing step size, respectively.
     Test accuracy of training CIFAR10   for resnet18 without BNLs using DP-SGD with BC and ALC. Here, css and dss denote constant step size and diminishing step size, respectively.}
    \label{fig:resnet18nonBNBCresult}
\end{figure}

\begin{figure}
    \centering
    \subfigure[css,constant $C$]{\includegraphics[width=0.49\textwidth]{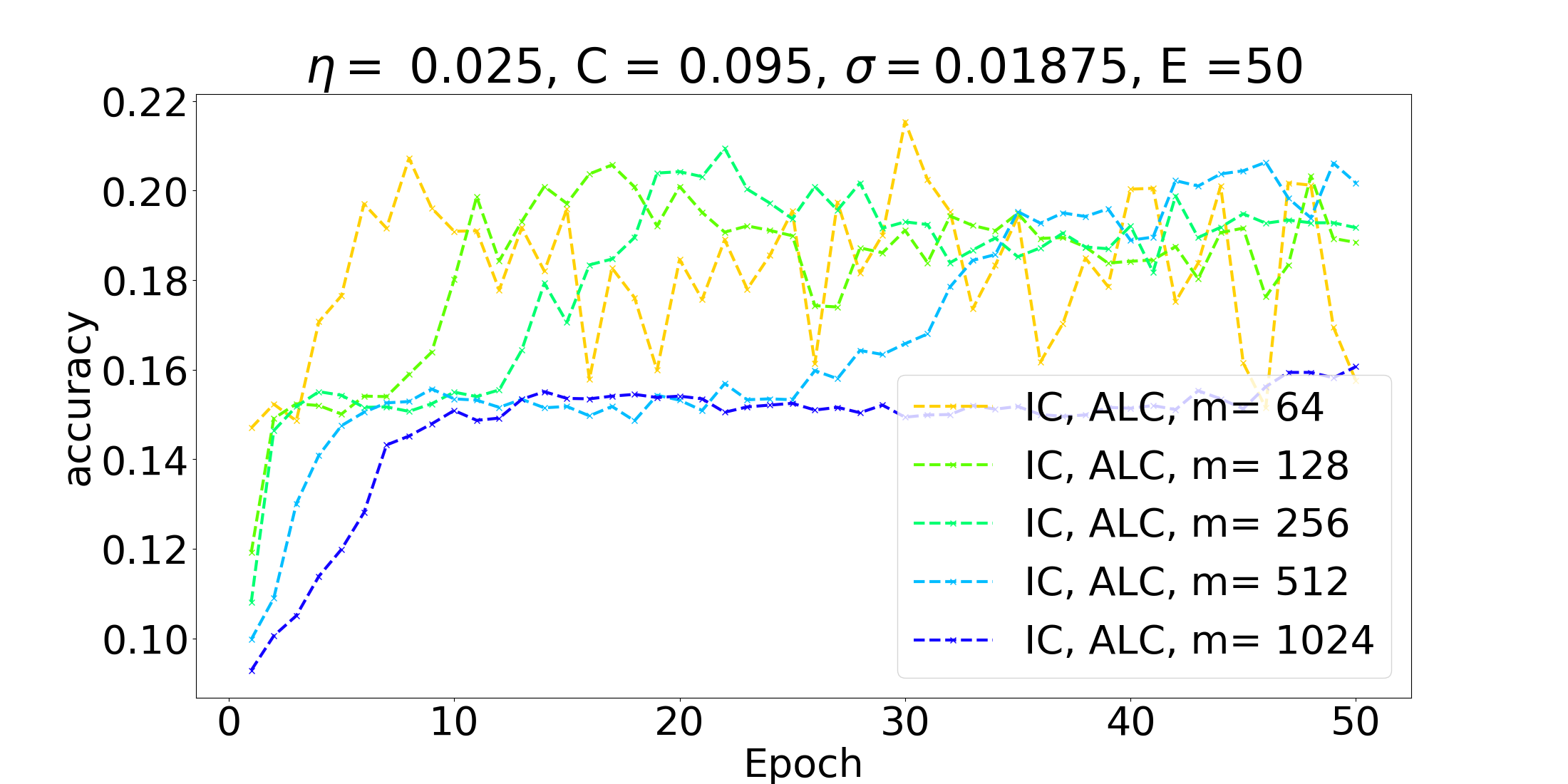}}
    \subfigure[css,diminishing $C$]{\includegraphics[width=0.49\textwidth]{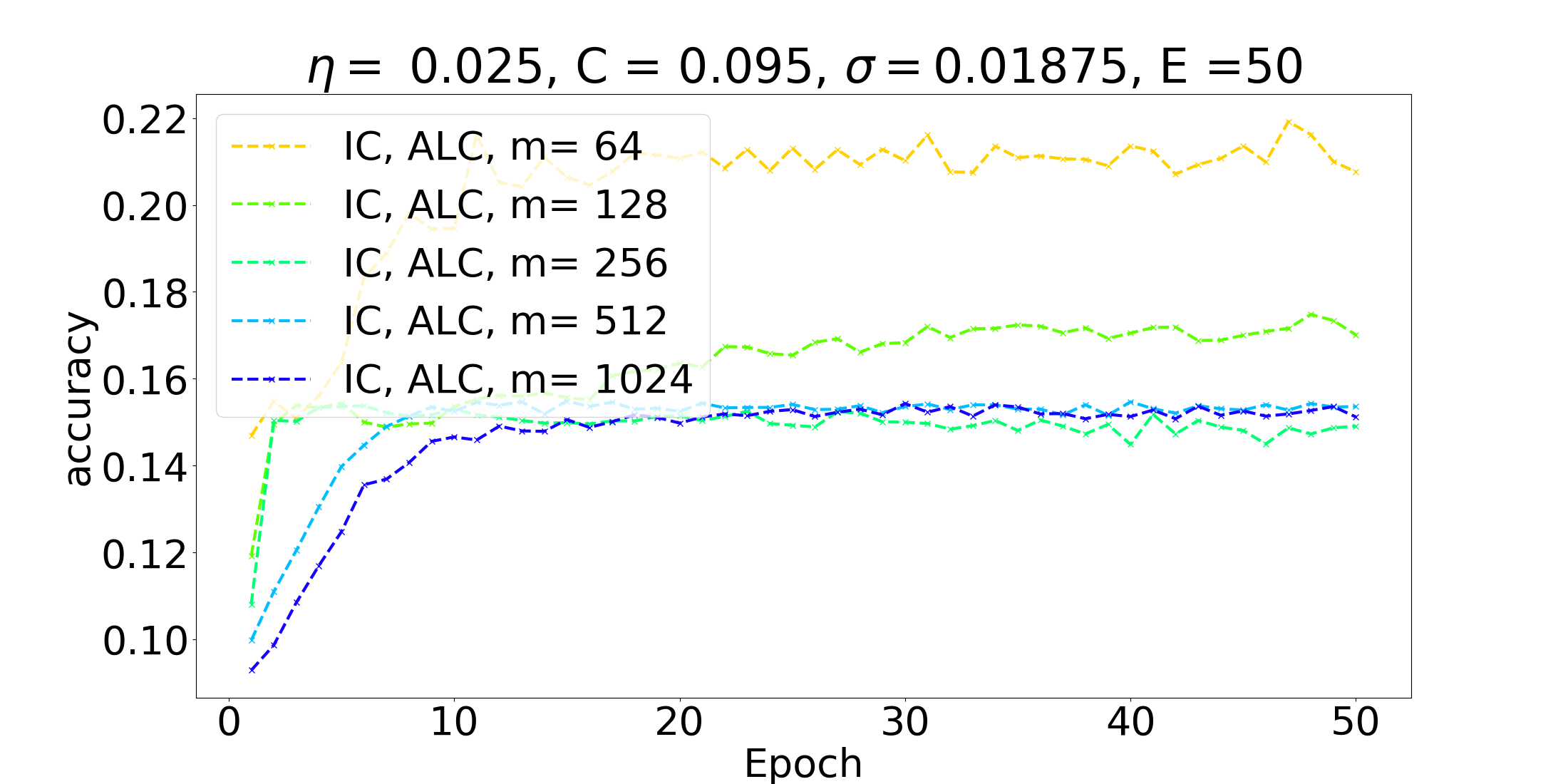}} \\
    \subfigure[dss,constant $C$]{\includegraphics[width=0.49\textwidth]{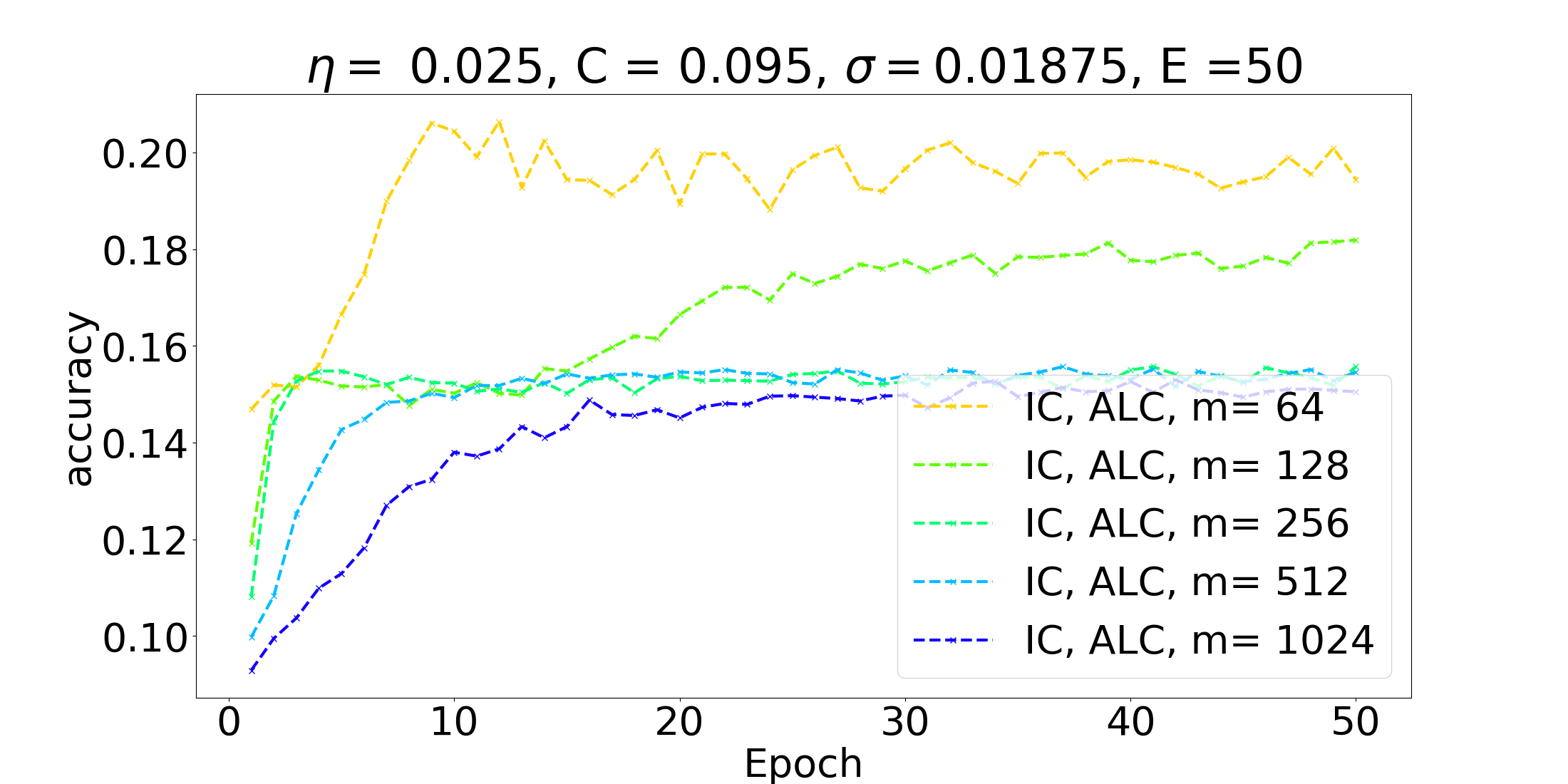}}
    \subfigure[dss,diminishing $C$]{\includegraphics[width=0.49\textwidth]{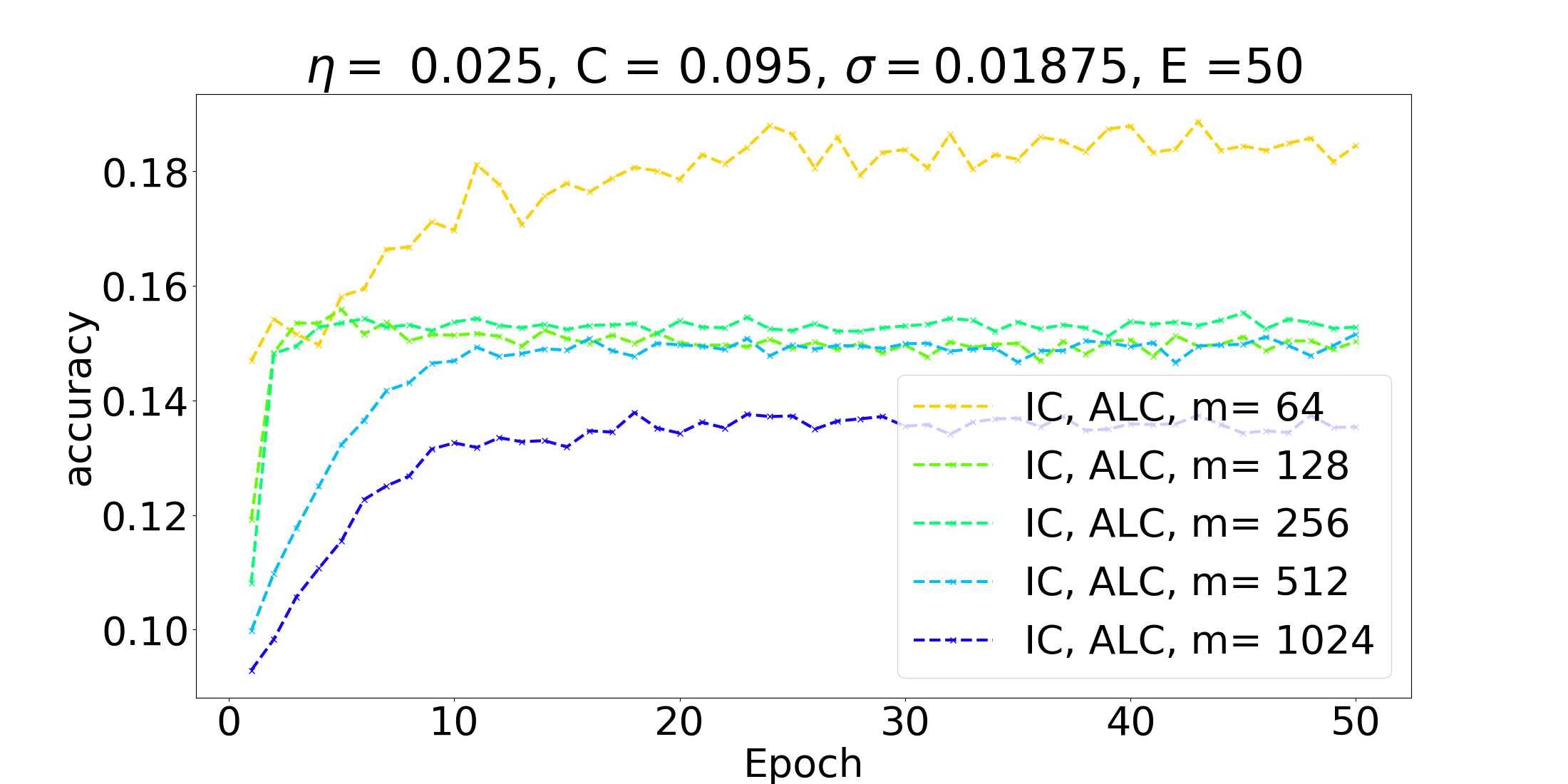}}
    \caption{
    %resnet18 without batch normalization layer under IC mode, css and dss denote constant step size and diminishing step size, respectively.
    Test accuracy of training CIFAR10   for resnet18 without BNLs using DP-SGD with IC and ALC. Here, css and dss denote constant step size and diminishing step size, respectively.}
    \label{fig:resnet18nonBNICresult}
\end{figure}

Next, we ask ourselves whether batch clipping  still outperforms individual clipping for the more complicated model such resnet18 without BNLs. As shown in Figure \ref{fig:resnet18nonBNBCresult}, we achieve $30\% \sim 40\%$ test accuracy for the various combinations of constant and diminishing step size and master constant, respectively (with the best test accuracy close to $40\%$ for constant step size and non-decaying master constant).  
%After switching between constant and diminishing stepsize, master clipping value $C$. 
On the other hand, we only achieve $18\% \sim 22\%$ test accuracy for individual clipping  as shown in Figure \ref{fig:resnet18nonBNICresult}.

\subsection{ALC versus FGC}
\label{subsec:ALCvsFGC}

\begin{figure}
    \centering
    \subfigure[Under the same Gaussian noise]{\includegraphics[width=0.49\textwidth]{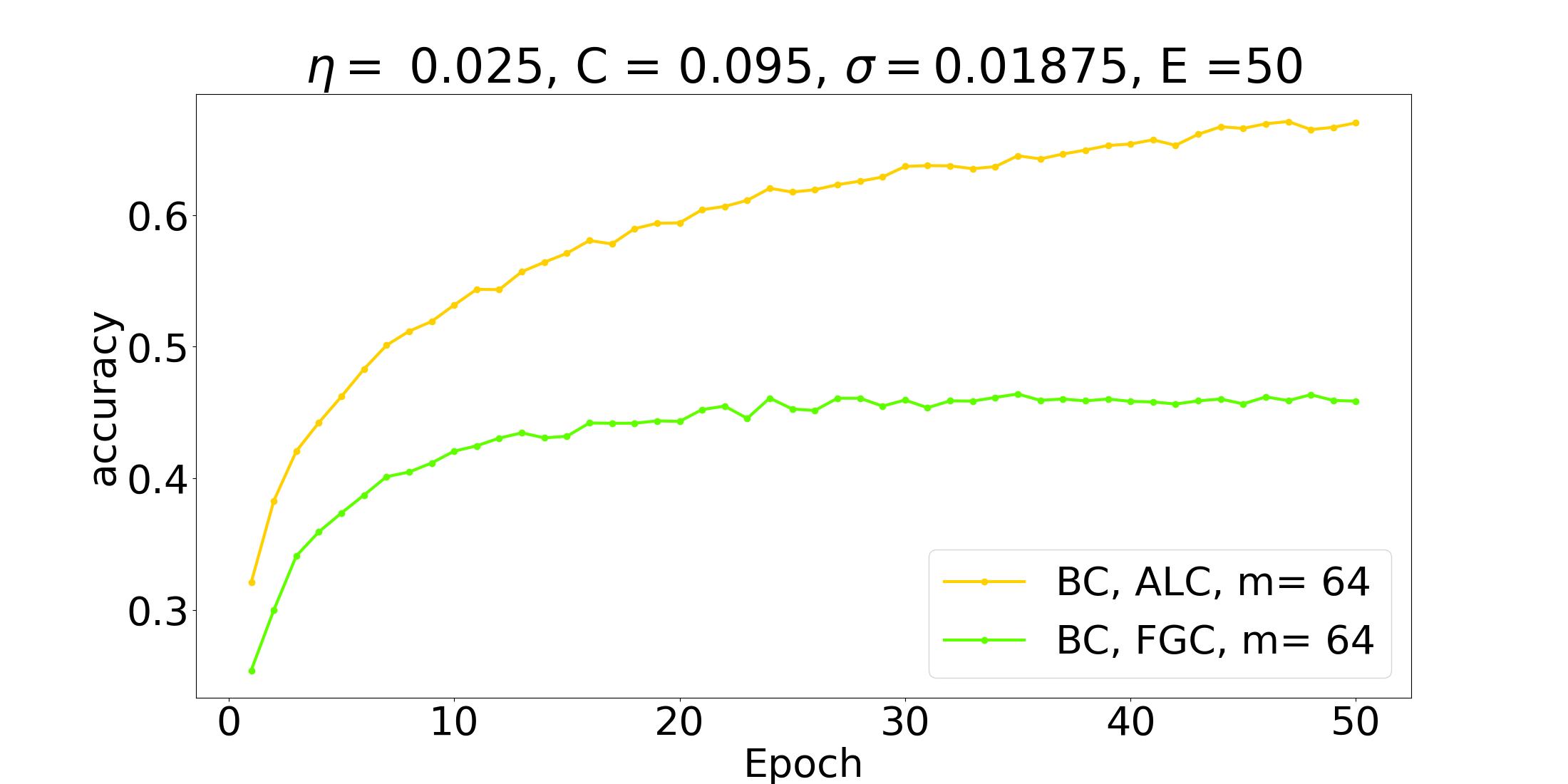}}
    \subfigure[Discounted noise multiplier]{\includegraphics[width=0.49\textwidth]{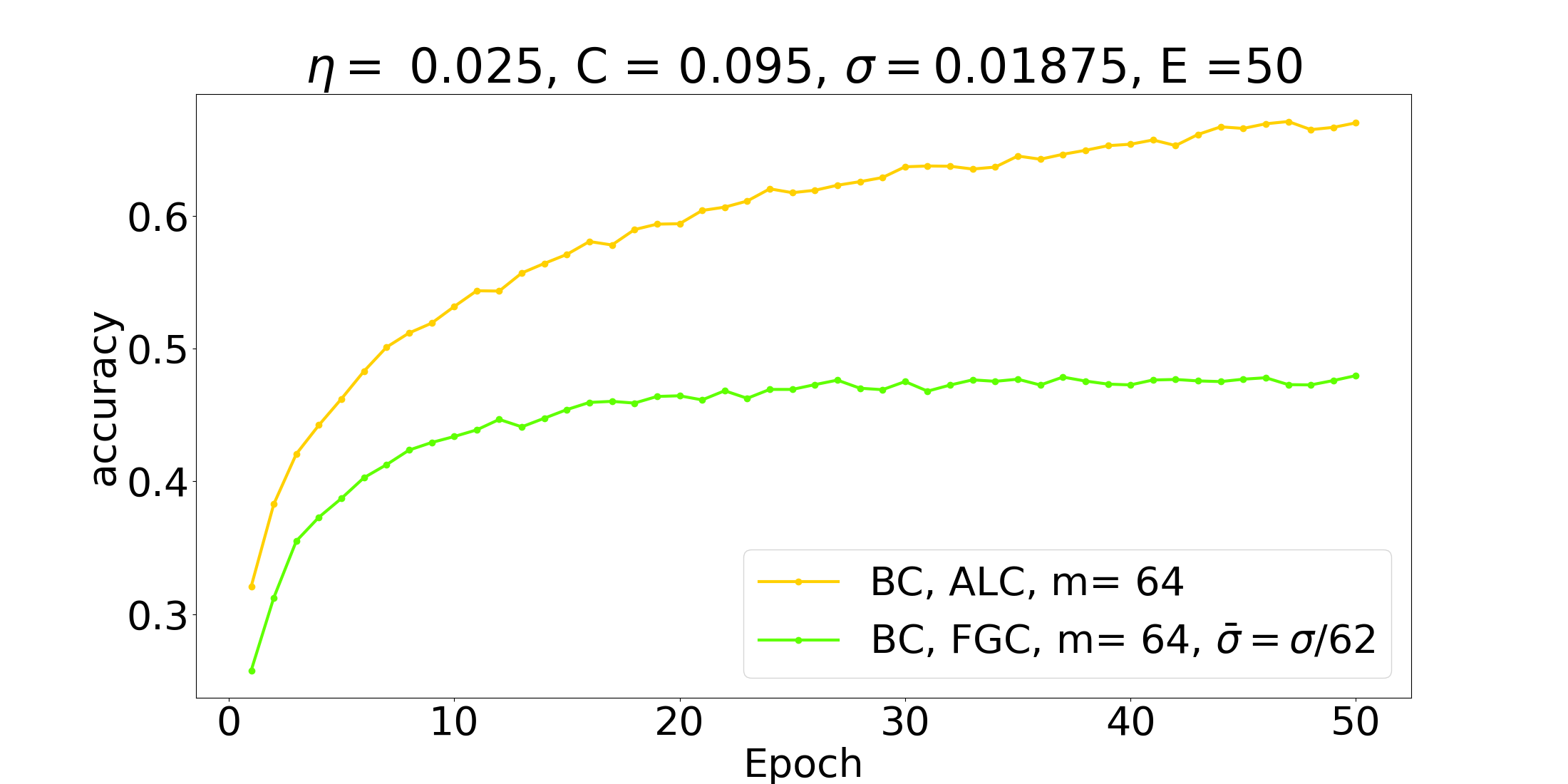}}
    \caption{Test accuracy of DP-SGD of resnet18 with CIFAR10 for  BC+ALC versus BC+FGC.}   
    \label{fig:ALCversusFGCSameAnddiscountednoise}
\end{figure}

We investigate how ALC compares with Full Gradient Clipping (FGC) under the same Gaussian Noise $N(0,(2C\sigma)^2)$. (FGC means that we clip the full gradient and do not separately clip layers as in ALC.) As shown in Figure \ref{fig:ALCversusFGCSameAnddiscountednoise}, we used DP-SGD with BC for 50 epochs in order to train resnet18 with CIFAR10; we use a non-decaying master clipping constant $C= 0.095$, noise multiplier $\sigma = 0.01875$, mini-batch size $m = 64$ and diminishing step size $\eta = 0.025$ with $\eta_{decay}=0.9$.
%under batch clipping mode. 
We achieve $66.66\%$ with ALC and $46.35\%$ with FGC method. This shows evidence that training with ALC leads to better convergence rate as well as better test accuracy. 
% \begin{figure}[ht]
%     \centering
%     \includegraphics[width=0.8\textwidth]{Figs/appendix_fig/resnet18_subsampling_BC_FGC_versus_ALC.png}
%     \caption{Testing accuracy of ALC versus FGC under the same Guassian noise}
%     \label{fig:ALCversusFGCsamenoise}
% \end{figure}

However, in the above experiment the privacy budget is not the same for ALC versus FGC: See Section \ref{sec:layerwiseclipping} in ALC we have an extra factor $\sqrt{L}$ in the DP guarantee, 
%privacy budget for ALC method 
where $L$ is the number of layers in the neural network model. Therefore, we also run the same experiment for FGC  with discounted noise multiplier $\bar{\sigma} = \sigma/\sqrt{L} = \sigma/\sqrt{62}$ so that both ALC and FGC correspond to the same DP guarantee. The result is shown in Figure \ref{fig:ALCversusFGCSameAnddiscountednoise} and shows that ALC outperforms FGC.

% \textcolor{red}{I correct to division by $\sqrt{62}$ and this should also be written in Figure \ref{fig:ALCversusFGCSameAnddiscountednoise}.b. I hope you indeed used division by $\sqrt{62}$ (which leads to worse test accuracy compared to division by 62. Why does FGC not improve from (a) to (b)? Did you plot the wrong curve in (b)?}

% \textcolor{red}{Ha to Marten: I think the results in 18(b) are correct, i.e., (almost) no improvement in accuracy. The dependence between accuracy and $\sigma$ is shown in Figs 9(b) and 12(b). Typically, the accuracies of small $\sigma$s are roughly the same after 50 epoches. I want to say the results in 9,12,18(b) are quite aligned with each other.}

% \begin{figure}[ht]
%     \centering
%     \includegraphics[width=0.8\textwidth]{Figs/appendix_fig/resnet18_subsampling_BC_FGC_versus_ALC_discounted.png}
%     \caption{Testing accuracy of ALC versus FGC with discounted noise multiplier}
%     \label{fig:ALCversusFGCdiscountednoise}
% \end{figure}

% TODO: 
% = convnet vary C (after MNIST)
% = MNIST dataset? (Done)

\section{Towards Balancing DP Guarantees and Accuracy}

In this paper we have discussed how ALC and BC can bring us closer to balancing DP with accuracy. In particular, the focus of this paper is on the ALC and BC techniques for improving robustness against added Gaussian noise for differential privacy.
However, we still need additional techniques in order to achieve a practical balance which allows a reasonable accuracy (say at most 10\% or 20\% drop) together with a DP guarantee which shows a trade-off curve reasonably close to the ideal $1-\alpha$ curve which represents perfect security. 

In Section \ref{sec:balance} we show that ALC+BC as discussed in this paper is not yet sufficient on themselves to find a practical balance: Even though ALC+BC allow us to be significantly more robust against added Gaussian noise for bootstrapping DP, just using these techniques will not yet make training of even lightweight neural network models with less complex training datasets sufficiently robust against the required DP noise for good/solid differential privacy.

One direction of tackling this problem is to enhance and/or optimize the ALC+BC techniques. In Section \ref{sec:optimization} we offer suggestions that focus on improving the presented ALC+BC. We leave it to future work to empirically study these possible improvements and optimizations. And we leave it to future work to find altogether new techniques that are complimentary and go beyond ALC+BL.

\subsection{Lightweight Neural Network Model with Less Complex Training Dataset} \label{sec:balance}

We show that even moving to a lightweight neural network with a less complex training data set still requires additional techniques beyond our BC+ALC and/or improvements  of the BC+ALC techniques. We show that BC+ALC as presented in this paper is on its own not yet sufficient even though they make the gap between DP guarantee and accuracy significantly smaller:
%\textcolor{red}{Ha to Marten: in my personal opinion, we do not want to have this subsection because we never have any fair comparison and good results. Current DPSGD does not give us good DP guarantee at the beginning. As researcher, we have to keep working on it to make it practical.}

We notice that Corollary 5.4 in \cite{dong2021gaussian} shows that, for $c>$, if $E=(N/m)\cdot c^2\rightarrow \infty$, then (\ref{eq:DPIC}) is asymptotically $G_{\mu}$-DP with $\mu=\sqrt{2}\cdot c\cdot h(\sigma)$ for 
$$ h(\sigma) = \sqrt{e^{\sigma^{-2}} \Phi(3\sigma^{-1}/2)+3\Phi(-\sigma^{-1}/2)-2}.$$
By using their interpretation of their result, we may conclude that (\ref{eq:DPIC}) is approximately $G_{\sqrt{2}\cdot c \cdot h(\sigma)}$ for $c=\sqrt{Em/N}$ for concrete large $N$ and relatively small $E$ and $m$. Including ALC means that we need to substitute $\sigma/\sqrt{L}$ for $\sigma$.
%%%%%% CORRECT %%%%%%

For  $\sigma = 0.01875$, we have $h(\sigma/\sqrt{L})\approx \sqrt{e^{(\sigma/\sqrt{L})^{-2}}}=e^{L\sigma^{-2}/2}$, i.e., a very very large number leading to no useful DP guarantee even for large datasets (with large $N$). In other words, $\sigma$ cannot be too small. Even $\sigma=0.5$ with $L=62$ for resnet-18  leads to 
$h(\sigma/\sqrt{L})= \sqrt{e^{62\cdot 4} \Phi(3\cdot \sqrt{62}) + 3\Phi(-\sqrt{62}) -2}$ which is prohibitively large for achieving a good DP guarantee. 
%$=\sqrt{ 54.60\cdot 0.9987 +3\cdot 0.1587 -2}=7.28$. 
%$h(\sigma)= \sqrt{e^4 \Phi(3) + 3\Phi(-1) -2}=\sqrt{ 54.60\cdot 0.9987 +3\cdot 0.1587 -2}=7.28$. 
%For our parameter setting $E=50$, $m=64$, and $N=\frac{9}{10}\cdot 50000$, we obtain $\approx G_{\sqrt{2Em/N}\cdot h(\sigma)}=G_{2.75}$-DP. This is still a rather weak DP guarantee, but for a factor 10 larger dataset we would already see a much more reasonable $G_{0.87}$-DP property. 
Figure \ref{fig:ourDPSGDwithDiffNoises} shows that choosing a larger $\sigma>0.5$ for obtaining a better DP guarantee gives too much noise resulting in a poor test accuracy of at most $20\%$, which is unacceptable.
%The main problem is that Figure \ref{fig:ourDPSGDwithDiffNoises} shows that $\sigma=0.5$ gives too much noise resulting in a poor test accuracy of about $20\%$, which is unacceptable. 
We conclude that resnet-18 and CIFAR10 represent a too deep neural network and complex dataset for a good balance between test accuracy and DP guarantee when using DP-SGD with our BC and ALC improvements. For now, we see that BC and ALC are two steps toward a better balance (after we are able to achieve convergence where this was not possible before for the original DP-SGD with IC) and that more techniques are needed for deep neural network models with complex training datasets. 

Section \ref{app:convnet} shows experiments for the lightweight convnet model with the complex CIFAR10 dataset which achieves a better $\approx 40\%$ test accuracy -- this demonstrates that a lightweight neural network model is more robust against noise and is better suitable for training with DP-SGD with BC and ALC.  Section \ref{app:MNIST} shows experiments for the lightweight BN-LeNet-5 model with the simple MNIST dataset which, if restricted to $50\%$ accuracy (for proper comparison with the $40\%$ test accuracy for convnet with CIFAR10), allows a much larger $\sigma=2.5$ resulting in a much improved $G_{0.52}$-DP guarantee ($h(\sigma/\sqrt{L})=1.513$ with $L=8$; $m=64$, $N=\frac{9}{10}\cdot 60000$, $E=50$). 

A smaller $\sigma=1.5$ (see Figure \ref{fig:LeNet5varySigma}.b in Section \ref{app:MNIST}) achieves a better balance of $\approx 67\%$ test accuracy with $\approx G_{1.99}$-DP ($h(\sigma/\sqrt{L})=5.783$). For a factor 15 larger dataset we would be able to improve $G_{1.99}$-DP to the better $G_{0.51}$-DP guarantee. We conclude that a more lightweight model and/or less complex (and larger) training dataset can potentially lead to a better balance between test accuracy and DP guarantee using the proposed BC and ALC techniques. Nevertheless, we will even want to improve $G_{0.51}$ to some trade-off function more like $G_{0.01}$ such that hypothesis testing indeed resembles a random guess. To date, this remains an open problem -- our ALC+BC techniques provide a step forward, but more complimentary techniques are needed.

\subsection{Towards Improving/Optimizing ALC+BC}
\label{sec:optimization}

From an accuracy perspective we see that  $\sigma=1.5$ (as mentioned above) or larger $\sigma$ may be needed even for a more lightweight network model with a less complex training dataset. For large $\sigma$, we notice that a Taylor series expansion of $h(\sigma)$ shows a linear dependency on $1/\sigma$. This shows that $\mu$ in a $G_{\mu}$-DP guarantee for ALC scales with $\sqrt{L}$ (since, as discussed before, ALC requires a factor $\sqrt{L}$ smaller $\sigma$ if we want to keep the same $G_\mu$-DP guarantee).
%, and this translates in the $\sqrt{L}$ dependency in $\mu$).
%\textcolor{red}{TO DO: we now understand that $\sigma$ is at least in the range $\geq 1$. For large $\sigma$, if this were possible giving good test accuracy, shows a Taylor series approximation with $h(\sigma)= ...$ and we clearly see its linear dependence on $1/\sigma$ implying a DP guarantee that will scale with $\sqrt{L}$ (its effect is not weakened by function $h$}
%Therefore, as a final remark, we observe that ALC effectively lowers $\sigma$ by a factor $\sqrt{L}$ in the DP analysis. 
The linear dependency of $\mu$ on $\sqrt{L}$ is due to the fact that there is no subsampling effect for the separate layers within a gradient computation; the leakage is directly composed over all the layers for a single gradient computation (without using a subsampling operator as is done in the general analysis leading to (\ref{eq:DPIC})). For this reason, it is advantageous to group layers that have similar clipping constant and clip the group rather than the individual layers within the group. E.g., Figure \ref{fig:resnet18gradientnorm} indicates we may use about 4 groups of layers representing a very small norm, to medium and larger norms. This reduces $L=62$ down to $L=4$ for resnet-18. Also, notice that the unexplored sparsification trick mentioned in Section \ref{sec:sparse} may offer another improvement.

Our discussion shows that as future work, we need to further optimize the promising ALC technique. One tempting direction is to not use a single training sample $\xi$ for updating all the layer gradients, but to use $\xi$ for a single layer gradient. So, rather than using a data sample $\xi$ for computing updates for all the layer gradients, we can think of using a data sample $\xi$ for updating just one of the layer gradients. Together with $\xi$ we choose one of the $L$ layers at random. In this way we still train all the layers. We notice that this approach means that we have subsampling for each layer gradient and we do not pay the composition price of $\sqrt{L}$ as explained in \ref{sec:layerwiseclipping}. However, one can think of this as $L$ separate learning tasks, each costing the same amount of training as the original learning task which learn the full weight vector across all layers at once. This means a composition of $L$ leakages and we again pay the price of $\sqrt{L}$ since $G_\mu^{\otimes L}=G_{\sqrt{L}\cdot \mu}$. Or, equivalently, one can argue that we need $L$ times more rounds in order to train the full weight vector, i.e., a factor $L$ more epochs, hence, the $\sqrt{L}$ factor penalty after composing over all the epochs. So,  this idea still does not improve the sought-after balance between DP guarantee and accuracy since we now have proper amplification from subsampling but at the price of $L$ times more rounds, and this cancels out, that is, no improvement.

Based on the above discussion we would like to somehow  only clip the overall full gradient while still keeping the better robustness against DP noise of ALC which requires layerwise clipping. 
Suppose that we associate a multiplication factor $m_j\geq 1$ to each layer $j$. We proceed as follows:
\begin{enumerate}
    \item As before, we first compute the full gradient composed of layer gradients:
$$\nabla_w f(w;\xi)  = ( \nabla_{w_1} f(w;\xi)  ||   .... || \nabla_{w_L} f(w;\xi) ).
$$
\item We use the multiplication factors $m_j$, $1\leq j\leq L$, to compute
$$\{ \nabla_w f(w;\xi)  \}_{m_1,...,m_L} = ( m_1\cdot \nabla_{w_1} f(w;xi)  ||  .... || m_L \cdot \nabla_{w_L} f(w;\xi) ).
$$
\item Now we perform full gradient clipping (FGC) with clipping constant $C$:
$$[ \{ \nabla_w f(w;\xi)  \}_{m_1,...,m_L} ]_C.
$$
We use this in computing formulas (the $a_h$) leading to the noised update $U$. Notice again that we can use the IC or BC approach in these formulas.
%(over a mini-batch for IC or we use BC).
\end{enumerate} 
The differential privacy argument follows the line of thinking of our analysis of BC in Section \ref{sec:BC}. Since we use FGC, we do not pay the $\sqrt{L}$ penalty.  This will significantly improve the trade-off function as discussed in Section \ref{sec:balance}.

What about robustness against the added Gaussian DP noise? The server receives a noised update of the form
$$U = ( U_1 || ... || U_L).$$
The server divides by the multiplication factors and computes
$$( U_1/m_1 || ... || U_L/m_L )$$
with which the global model is updated.
The effect of the proposed trick using multiplication factors $m_j$ is that dividing by the multiplication factors retrieves the original layer gradients -- if there is no clipping noise. 
%(potentially normalized together by the clipping constant). 
In this process we  reduce the added Gaussian noise, since the noises are divided by the factor $m_j\geq 1$.
In our ALC we estimate the expected norm of each layer $j$ denoted by $e_j$. We equate $M$ to the maximum of all $e_j$. We compute layer clipping constants $C\cdot e_j/M$. In the above approach based on multiplication factors we may define $m_j=M/e_j$. Since this will increase each layer norm to $M$ in expectation, we will want to choose a higher overall clipping constant $C$ in FGC (as compared\footnote{We can use the same collecting-layers-into-4-groups argument for resnet18 with which we started this section and conclude that the clipping constant corresponding to the use of multiplication factors is about $\sqrt{4}=2$ larger.} to the master clipping constant $C$ used in ALC). We see that the proposed new trick on one hand remains robust to Gaussian noise added to layers that have a small norm compared to other layers. On the other hand $C$ needs to be fine-tuned and may be larger than the $C$ of ALC, which means that the overall added noise is larger and makes this solution less robust.  
Concluding, we have the original DP guarantee without $\sqrt{L}$ penalty, while we make sure that layer gradients with small norms get multiplied by a large $m_j$ so that the effect of the added noise for that layer is not going to be overpowering. This is also the goal which ALC wants to achieve. We leave it to future work to experiment with multiplication factors and in this sense optimize over ALC and/or find a better balance between DP guarantee and test accuracy.

As a final remark, we notice that we do not need to restrict ourselves to using BC, we may use GBC which allows momentum based update rules. We leave it to future work to find out whether this can lead to more robustness against added Gaussian DP noise.

\end{document}